\documentclass{article} 
\usepackage{iclr2026_conference_arxiv,times}
\usepackage{amsmath,amsfonts,bm}

\def\eqref#1{equation~\ref{#1}}
\def\1{\bm{1}}

\DeclareMathAlphabet{\mathsfit}{\encodingdefault}{\sfdefault}{m}{sl}
\SetMathAlphabet{\mathsfit}{bold}{\encodingdefault}{\sfdefault}{bx}{n}

\usepackage{url}
\usepackage{booktabs}       
\usepackage{amsfonts}       
\usepackage{nicefrac}       
\usepackage{xcolor}         
\usepackage{graphicx}
\usepackage{wrapfig}
\usepackage{multirow}
\usepackage{epsfig}
\usepackage{amsmath}
\usepackage{amssymb}
\usepackage{subfigure}
\usepackage{marvosym}
\usepackage{color}
\usepackage{threeparttable}
\usepackage{algorithm}
\usepackage{algpseudocode}
\usepackage{setspace}
\usepackage{amsthm}
\usepackage{verbatim}
\usepackage{dsfont}
\usepackage{colortbl}
\usepackage{pifont}
\usepackage{transparent}
\usepackage{hyperref}

\title{CoRA: Covariate-Aware Adaptation of \\ Time Series Foundation Models}

\author{Guo Qin\thanks{Equal Contribution}, 
  Zhi Chen\footnotemark[1],
  Yong Liu\footnotemark[1], 
  Zhiyuan Shi, Haixuan Liu,
  Xiangdong Huang, Jianmin Wang, \\ \textbf{Mingsheng Long}\textsuperscript{\Letter} \\
  School of Software, BNRist, Tsinghua University, Beijing 100084, China \\
  \texttt{\small \{qinguo24,chenzhi21,liuyong21\}@mails.tsinghua.edu.cn} \\ 
  \texttt{\small \{huangxdong,jimwang,mingsheng\}@tsinghua.edu.cn}
}

\newcommand{\boldres}[1]{{\textbf{\textcolor{red}{#1}}}}
\newcommand{\secondres}[1]{{\underline{\textcolor{blue}{#1}}}}

\iclrfinalcopy

\begin{document}
\maketitle

\begin{abstract}
Time Series Foundation Models (TSFMs) have shown significant impact through their model capacity, scalability, and zero-shot generalization. However, due to the heterogeneity of inter-variate dependencies and the backbone scalability on large-scale multivariate datasets, most TSFMs are typically pre-trained on univariate time series. This limitation renders them oblivious to crucial information from diverse covariates in real-world forecasting tasks. To further enhance the performance of TSFMs, we propose a general \textbf{Co}variate-awa\textbf{R}e \textbf{A}daptation (\textbf{CoRA}) framework for TSFMs. It leverages pre-trained backbones of foundation models while effectively incorporating exogenous covariates from various modalities, including time series, language, and images, to improve the quality of predictions. Technically, CoRA maintains the equivalence of initialization and parameter consistency during adaptation. With preserved backbones of foundation models as frozen feature extractors, the outcome embeddings from foundation models are empirically demonstrated more informative than raw data. Further, CoRA employs a novel Granger Causality Embedding (GCE) to automatically evaluate covariates regarding their causal predictability with respect to the target variate. We incorporate these weighted embeddings with a zero-initialized condition-injection mechanism, avoiding catastrophic forgetting of pre-trained foundation models and gradually integrates exogenous information. Extensive experiments show that CoRA of TSFMs surpasses state-of-the-art covariate-aware deep forecasters with full or few-shot training samples, achieving $31.1\%$ MSE reduction on covariate-aware forecasting. Compared to other adaptation methods, CoRA exhibits strong compatibility with various advanced TSFMs and extends the scope of covariates to other modalities, presenting a practical paradigm for the application of TSFMs.
\end{abstract}

\begin{figure*}[htbp]
\begin{center}
	\centerline{\includegraphics[width=\columnwidth]{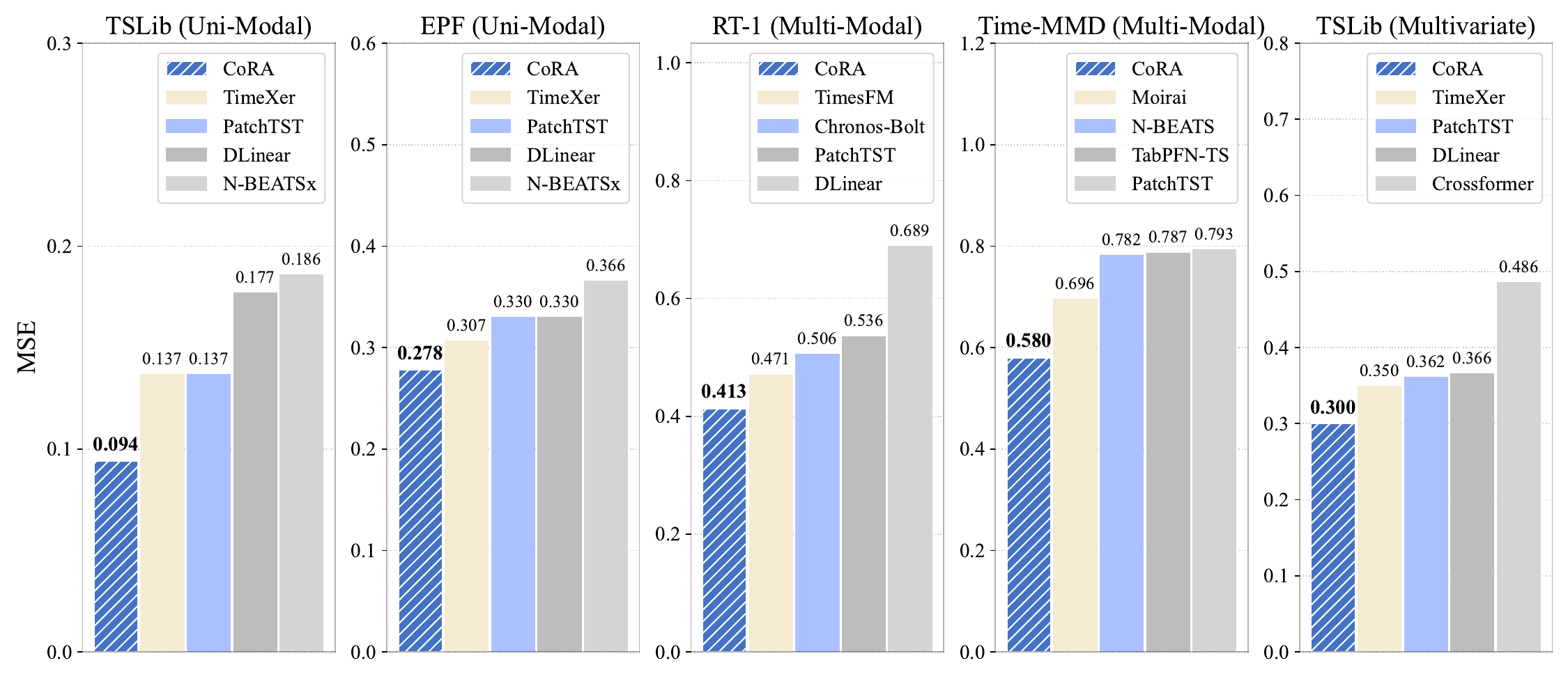}}
        \vspace{-7pt}
	\caption{CoRA performance on different covariate-aware benchmarks.}
	\label{fig:intro_performance}
\end{center}
\vspace{-20pt}
\end{figure*}
\section{Introduction}

Time series forecasting has gained increasing prominence in real-world applications, such as weather forecasting~\citep{hittawe2024time}, supply chain optimization~\citep{panda2023time} and financial market assessment~\citep{cheng2022financial}. With the rapid development of large-scale time-series datasets~\citep{woo2023pushing} and scalable architectures~\citep{vaswani2017attention}, recent research has focused on developing Time Series Foundation Models (TSFMs)~\citep{das2023decoder, liu2024moirai, ansari2024chronos, liu2025sundial}, which exhibit impressive scalability and out-of-box generalization performance across various applications.

\begin{figure*}[b]
\begin{center}
	\centerline{\includegraphics[width=\columnwidth]{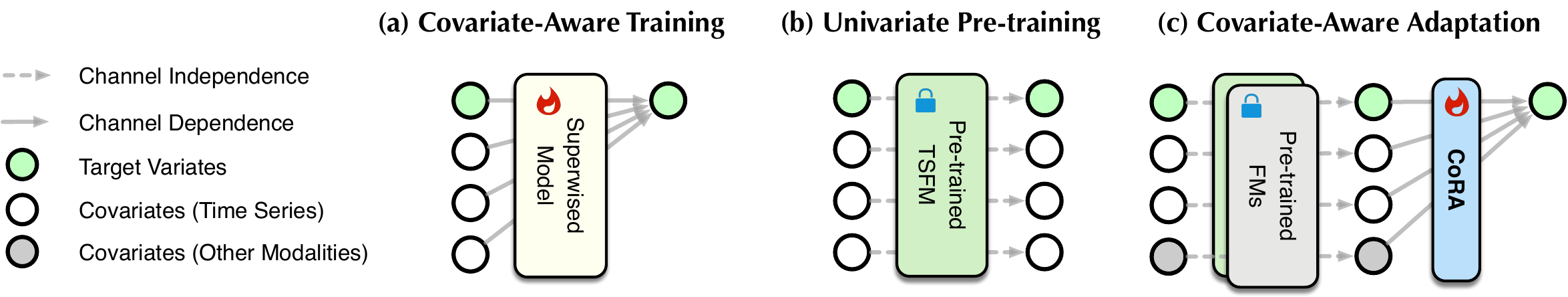}}
        \vspace{-7pt}
	\caption{Several paradigms of time series forecasting: (a) Covariate-aware deep models are supervisedly trained in a channel-dependent way. However, the backbone can be task-specific and challenged to scale up. (b) TSFMs designed to address data heterogeneity are generally pre-trained and predict on univariate time series. which makes them infeasible to utilize inter-variate dependencies explicitly. (c) CoRA leverages various foundation models, incorporates exogenous information to predict the target variate, and rapidly adapts to specific tasks without altering pre-trained models.}
	\label{fig:intro}
\end{center}
\vspace{-20pt}
\end{figure*}

Despite time series are typically multi-dimensional data, most TSFMs are pre-trained on univariate time series~\citep{das2023decoder, liu2024timer, shi2024time}, primarily due to the considerable heterogeneity in dimensionality and inter-variate relationships across datasets. In particular, the dependencies among variates in one dataset often fail to generalize to others. For example, transferring relationships learned from meteorological variates to the financial domain may not be sensible. Besides, covariate-aware deep forecasters, which are trained in a channel-dependence approach~\citep{qiu2025comprehensive}, have not been well-demonstrated to be scalable and versatile. Meanwhile, an important paradigm of foundation models involves large-scale pre-training on general large-scale data and adaptation to task-specific datasets. Therefore, these constraints necessitate the paradigm shift as shown in Figure~\ref{fig:intro}, which adapts TSFMs to covariate-aware forecasting scenarios while revitalizing the pre-trained backbone of foundation models~\citep{arango2025chronosx, benechehab2025adapts}.

Different from adaptation methods for language models such as LoRA~\citep{hu2021lora}, covariate-aware adaptation in time series forecasting faces fundamentally different challenges. The difficulty lies in the multi-dimensionality and the heterogeneity of modalities in covariates. Simply incorporating exogenous information into the target variate is insufficient, because dependencies among variates are often domain-specific, noncausal, and sometimes noisy. Therefore, adaptation of TSFM requires not only the integration of covariate information but also evaluating the causality of different covariates. Guided by the principled criteria, we delve into Granger causality, a foundational concept for identifying causal dependencies in time series forecasting~\citep{granger1969investigating}, and develop a data-dependent approach to ground covariate-aware adaptation with interpretable modular design.

While prior works~\citep{arango2025chronosx, benechehab2025adapts, han2025unica} attempt to incorporate time series covariates into TSFMs, they inject covariate-aware modules that alter the embeddings away from the pre-trained embedding space. Besides, previous adaptation methods introduce trainable modules without zero-initialization, implying that the initial outputs of the adapted model are no longer equivalent to the pre-trained TSFMs. Empirically, adaptation without zero-initialization will cause unstable training, catastrophic forgetting and sometimes even worse performance than just zero-shot evaluation~\citep{hu2021lora, peebles2023scalable}.

In this paper, we introduce \textbf{CoRA}, a general, effective, and interpretable framework to adapt TSFMs on covariate-aware forecasting tasks, where covariates cover time series, language, images, and other structured data. Concretely, CoRA treats pre-trained foundation models of different modalities as frozen embedding extractors. With extracted embeddings from raw covariates, CoRA includes a covariate evaluation and routing module, termed Granger Causality Embedding (GCE), which automatically produces a causally-informed significance score during adaptation. These embeddings are then integrated through a zero-initialized condition-injection mechanism by learning scale and shift parameters. CoRA achieves state-of-the-art performance while requiring fewer samples compared to supervised models and previous adaptation methods. In-depth studies validate the generality and interpretability of the proposed framework. Our main contributions are summarized as follows:

\begin{itemize}
    \item We emphasize that an important paradigm of covariate-aware forecasting on TSFMs, which effectively revitalize pre-trained foundation models and address the unique challenges in utilizing high-dimensional, multi-modal, and causally-dependent covariates.
    \item We propose CoRA, a general and effective covariate-aware adaptation framework that freezes pre-trained models and introduces a Granger Causality Embedding for principled covariate selection, combined with a zero-initialized condition-injection mechanism.
    \item Extensive experiments across diverse benchmarks demonstrate that CoRA achieves state-of-the-art performance, requires fewer training samples, and provides interpretable insights into covariate causality, surpassing both supervised models and other adaptation methods.
\end{itemize}

\section{Related Work}
\subsection{Time Series Foundation Models}
Recent research has explored pre-training Time Series Foundation Models (TSFMs) on large-scale datasets, enabling strong zero-shot generalization to downstream tasks. TimesFM~\citep{das2023decoder} and Timer~\citep{liu2024timer} are the first to adopt a decoder-only Transformer architecture with the next-token prediction objective. Chronos~\citep{ansari2024chronos} introduces a discretization approach for time series and predicts next tokens using LLM backbone and language modeling. Sundial~\citep{liu2025sundial} proposes TimeFlow, incorporating generative modeling to realize the flexibility of probabilistic forecasting. However, these models are limited to univariate pre-training, which restricts their applicability to downstream tasks involving multi-dimensional or multi-modal covariates. One exception is that Moirai~\citep{woo2024unified} adopts multivariate pre-training by flattening variates and appending variate-wise embeddings, but it has to subsample multivariate series with a fixed size for training stability, leading to incomplete perception for high-dimensional time series inputs.

\subsection{Covariate-Aware Deep Forecasters}
In real-world time series forecasting, covariates play a crucial role in improving the predictability of target variate. Classical approaches such as ARIMAX~\citep{williams2001multivariate} and SARIMAX~\citep{vagropoulos2016comparison} model the correlations between covariates and the target variate by linear regression. More recent deep learning methods, such as the Temporal Fusion Transformer~\citep{lim2021temporal}, emphasize variate selection as a key mechanism. Other approaches, including NBEATSx~\citep{olivares2023neural} and TiDE~\citep{das2023long}, argue that forecasting models can directly leverage future covariate information when predicting target values. TimeXer~\citep{wang2024timexer} achieves competent performance by modeling the target variate at the patch level and the covariates at the series level. Time-VLM~\citep{zhong2025time} leverages vision-language backbones to integrate temporal, visual, and textual information for multi-modal forecasting. However, supervised deep models trained from scratch may yield suboptimal performance without substantial task-specific data.

\subsection{Adapation Methods of Foundation Models}
Adaptation of foundation models such as LoRA~\citep{hu2021lora, NEURIPS2023_1feb8787} is typically applied in language and vision models, where the upstream and downstream tasks share the same 1D-sequence structure. In contrast, adapting univariate pre-trained TSFMs to covariate-aware scenarios introduces dimensional changes in the input structure. Prior works such as ChronosX~\citep{arango2025chronosx}, AdaPTS~\citep{benechehab2025adapts}, and UniCA~\citep{han2025unica} modify the input structure of TSFMs by injecting covariate information before passing data into the backbone, which alters the embedding space formulated during pre-training and leads to catastrophic forgetting. Moreover, adaptation of foundation models relies on zero-initialization~\citep{goyal2017accurate} to ensure that the training start-point begins consistently with the pre-trained model. However, such principled strategies have not been properly considered in existing TSFMs adaptation methods.

\section{Approach}
\begin{figure*}[tbp]
\begin{center}
	\centerline{\includegraphics[width=\columnwidth]{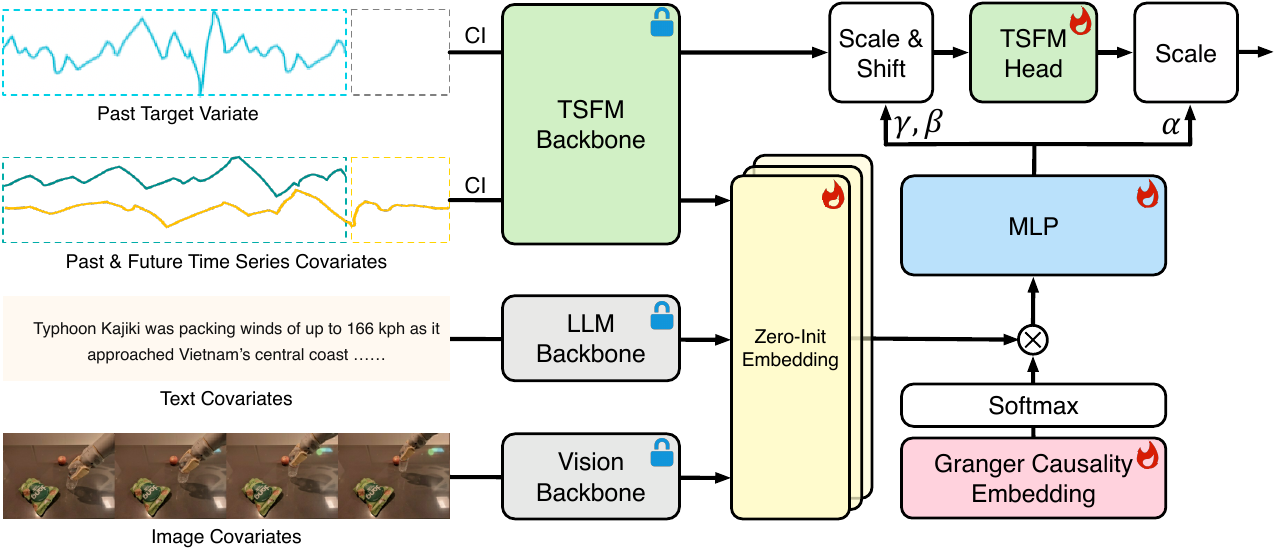}}
        \vspace{-10pt}
	\caption{Overall architecture of CoRA. CoRA freezes the backbone of foundation models as embedding extractors for multi-modal covariates, which are then selected by a trainable Granger Causality Embedding. This refined embedding is injected into the original TSFM head via a zero-initialized module to generate the shifting and scaling factors for final predictions.}
	\label{fig:arch}
\end{center}
\vspace{-20pt}
\end{figure*}
In covariate-aware forecasting, we consider one target variate $\mathbf{x}_{1:T}=\{x_1,\ldots,x_T\} \in \mathbb{R}^{T}$ observed over $T$ time steps along with exogenous covariates $\mathbf{C}_{1:\tau} = \{\mathbf{C}_1,\ldots,\mathbf{C}_{\tau}\}$ \footnote{Covariates may be future-unknown ($\tau=T$), future-known ($\tau=T+H$), or static covariates ($\tau=1$).}. The task is to train a forecaster $f_{\theta}$ parameterized by $\theta$ that can predict the target variate 
$\mathbf{x}_{T+1:T+H} = \{x_{T+1},\ldots,x_{T+H}\}$ for the next $H$ time steps:
\begin{equation}
f_{\theta} : (\mathbf{x}_{1:T}, \mathbf{C}_{1:\tau}) \mapsto \hat{\mathbf{x}}_{T+1:T+H}.
\end{equation}

\subsection{Foundation Models as Forzen Embedding Extractor}

For real-world forecasting, exogenous covariates are always multi-dimensional (e.g., multivariate time series) and multi-modal. In contrast to previous methods that solely adapt the foundation model of time series,  we categorize exogenous covariates into three mainstream modalities. As illustrated in Figure~\ref{fig:arch}, we separate covariates as $N$ one-dimensional sequences, such as univariate time series, text, or image snapshots, and extract per-step embeddings from corresponding frozen models:
\begin{equation}
\mathbf{E}_{1:\tau_i}^{m_i}=\operatorname{FM-Backbone}(\mathbf{C}^{m_i}_{1:\tau_i}),\ i=1,\dots,N,\ m_i \in \{\mathrm{ts}, \mathrm{txt}, \mathrm{img}\}.
\end{equation}

At each time step, the embeddings $\mathbf{E}^{\mathrm{ts}}_{t} \in \mathbb{R}^{N_{\mathrm{ts}} \times D_{\mathrm{ts}}}$, 
$\mathbf{E}^{\mathrm{txt}}_{t} \in \mathbb{R}^{N_{\mathrm{txt}} \times D_{\mathrm{txt}}}$, 
and $\mathbf{E}^{\mathrm{img}}_{t} \in \mathbb{R}^{N_{\mathrm{img}} \times D_{\mathrm{img}}}$ capture the exogenous information of corresponding covariates by leveraging the embeddings generated before the last layer of the foundation models, where $D_{\mathrm{ts}}$, $D_{\mathrm{txt}}$, $D_{\mathrm{img}}$ denote the latent dimensions of the respective foundation models and $N_{\mathrm{ts}}$, $N_{\mathrm{txt}}$, $N_{\mathrm{img}}$ represent the number of covariates categorized into each modality, with the total number of covariates $N = N_{\mathrm{ts}} + N_{\mathrm{txt}} + N_{\mathrm{img}}$.

For dynamic covariates that are recorded at each time step, CoRA regards one covariate as a whole by aggregating the embeddings over all time steps. For typical TSFMs adopting the decoder-only or encoder-decoder architecture, we employ the last-step embedding that corresponds to the latest-known values, which captures all previous context in one single-series covariate. For language and vision foundation models that encode one snapshot, we utilize the averaged embeddings across all snapshots of time steps (for simplicity, we omit the variate index $i$):
\begin{equation}
\tilde{\mathbf{E}}^{\mathrm{ts}} = \mathbf{E}^{\mathrm{ts}}_{\tau},\ \tilde{\mathbf{E}}^{\mathrm{txt}} = \frac{1}{\tau} \sum_{t=1}^{\tau} \mathbf{E}^{\mathrm{txt}}_t,\ \tilde{\mathbf{E}}^{\mathrm{img}} = \frac{1}{\tau} \sum_{t=1}^{\tau} \mathbf{E}^{\mathrm{img}}_t.
\end{equation}
For the target variate, we use the TSFM backbone to extract its embeddings and take the embedding at the last time step $T$ to capture the overall lookback information:
\begin{equation}
\begin{aligned}
\mathbf{E}^{\mathrm{target}}_{1:T} = \operatorname{TSFM-Backbone} (\mathbf{x}_{1:T}) ,\ \tilde{\mathbf{E}}^{\mathrm{target}} = \mathbf{E}^{\mathrm{target}}_{T}.
\end{aligned}
\end{equation}

\subsection{Covariate-Aware Adaptation}
\paragraph{Granger Causality} 

Granger causality test~\citep{granger1969investigating} is a statistical hypothesis test used to determine whether using a covariate $\mathbf{C}$ and $\mathbf{x}_{1:T}$ to predict $\mathbf{x}_{T+1:T+H}$ yields a lower prediction error than using $\mathbf{x}_{1:T}$ alone. If so, $\mathbf{C}$ is said to Granger causes $\mathbf{x}$. Unlike multivariate correlations, Granger causality directly reflects the predictive efficacy of covariates to target variates. For example, the correlation of a sine and cosine wave is zero, but the granger causality test of them is significant.

\paragraph{Covariate Selection} In typical covariate-aware forecasting tasks, multiple covariates are involved, and their significance of Granger causality with respect to the target variate may differ considerably. Therefore, we introduce a trainable Granger Causality Embedding $\mathbf{W}_{\text{GC}} \in \mathbb{R}^{N}$, which learns to quantify the causal influence of each covariate on $\mathbf{x}_{1:T}$. Empirically, we observe that the learned Granger Causality Embedding exhibits highly consistent result with the statistical test of Granger causality in Section~\ref{sec:interp}. Concretely, we first align the embeddings of multi-modal covariates into a unified hidden space since the latent dimensions of foundation models are not necessarily identical:
\begin{equation}
\begin{aligned}
\hat{\mathbf{E}}^{m_i} &= \tilde{\mathbf{E}}^{m_i}\, \mathbf{W}^{m_i} + \mathbf{b}^{m_i},\ i=1,\dots,N,\ 
m_i \in \{\mathrm{ts}, \mathrm{txt}, \mathrm{img}\}, \\
\hat{\mathbf{E}} &= \operatorname{Concat}\Big(\hat{\mathbf{E}}^{\mathrm{ts}}, 
\hat{\mathbf{E}}^{\mathrm{txt}}, \hat{\mathbf{E}}^{\mathrm{img}}\Big).
\end{aligned}
\end{equation}
where $\mathbf{W}^{m_i} \in \mathbb{R}^{D_{m_i} \times D}$, $\mathbf{b}^{m_i} \in \mathbb{R}^{D}$ for $m_i \in \{\mathrm{ts}, \mathrm{txt}, \mathrm{img}\}$, and $\hat{\mathbf{E}} \in \mathbb{R}^{N \times D}$. Afterwards, we use Granger Causality Embedding $W_{\text{GC}} \in \mathbb{R}^{N}$ to evaluate and gate each covariate during the adaptation process, yielding a unified embedding that aligns the latent space of TSFMs:
\begin{equation}
\mathbf{H} = \operatorname{Softmax}(\mathbf{W}_{\text{GC}})\cdot\hat{\mathbf{E}}.
\end{equation}
\paragraph{Covariate Injection} With obtained overall exogenous embeddings of all covariates, we adopt an adaptive layer-normalization (adaLN) layer proposed by DiT~\citep{peebles2023scalable}, which is widely shown to outperform approaches such as concatenation and cross-attention on continuous-valued modality. Specifically, $\mathbf{H}$ is mapped into $\mathbf{\alpha} \in \mathbb{R}^{H}$ and $\mathbf{\beta}$, $\mathbf{\gamma} \in \mathbb{R}^{D}$ via a lightweight $\operatorname{MLP}(\cdot)$. The outcomes are then applied via shift-and-scale operations to modulate the statistics before and after the original head of TSFM, thereby injecting the covariate information into the adaptation process:
\begin{equation}
\begin{aligned}
\mathbf{\gamma}, \mathbf{\beta}, \mathbf{\alpha} &= \operatorname{MLP}\big(\mathbf{H}\big), \\
\hat{\mathbf{x}}_{T+1:T+H} &= (1 + \mathbf{\alpha})\operatorname{TSFM-Head}\big(\mathbf{\gamma} + (1 + \mathbf{\beta})\ \tilde{\mathbf{E}}^{\mathrm{target}}\big).
\end{aligned}
\end{equation}

\paragraph{Zero-Initialization} Similar to LoRA~\citep{hu2021lora}, we zero-initialize the parameters of $\mathbf{W}^{m_i} \in \mathbb{R}^{D_{m_i} \times D}$, $\mathbf{b}^{m_i} \in \mathbb{R}^{D}$ for $m_i \in \{\mathrm{ts}, \mathrm{txt}, \mathrm{img}\}$ and the $\operatorname{MLP}$. Therefore, the overall model is identical to the pre-trained TSFM. This design ensures adaptation begins from the pre-trained state, while progressively integrating additional information in a stable and incremental manner.

\section{Experiments}
We conduct comprehensive experiments to evaluate the effectiveness of CoRA, covering uni-modal and multi-modal covariate-aware forecasting, few-shot forecasting, and extensions to multivariate forecasting. The overall performance is provided in Figure~\ref{fig:intro_performance}. We further provide in-depth analysis, including generality across different TSFMs, ablation studies, and model interpretability.

\subsection{Main Results}
In this section, we conduct extensive experiments to evaluate the performance of CoRA, compared with existing adaptation methods and advanced supervised deep forecasters. For fair comparison, we adopt Sundial~\citep{liu2025sundial} as the backbone model for all adaptation approaches. Moreover, we ensure none of the test sets overlap with Sundial’s training data to avoid potential data leakage.

\subsubsection{Uni-Modal Covariate-Aware Forecasting}

\paragraph{Setups} 
In the uni-modal setting, all covariates are time series. We conduct both long-term and short-term uni-modal covariate-aware forecasting experiments. In the long-term setting, we use seven real-world datasets, including ECL, ETT (4 subsets), Traffic, and Weather, employed in Autoformer~\citep{wu2021autoformer}, where the final dimension serves as the target variate and the remaining dimensions as covariates. In the short-term setting, we adopt the electricity price forecasting (EPF) task~\citep{lago2021forecasting}, with electricity price as the target variate and two correlated covariates. 

\paragraph{Results}
As shown in Table~\ref{tab:tslib_exo_result} and Table~\ref{tab:epf_result}, CoRA delivers state-of-the-art performance across both long- and short-term forecasting. Specifically, in long-term forecasting, CoRA outperforms the strongest supervised model TimeXer~\citep{wang2024timexer}, by 31.1\% in MSE and 19.8\% in MAE, stressing the advantage of building on pre-trained TSFMs rather than training task-specific models from scratch. Compared to other adaptation methods, using the same model Sundial~\citep{liu2025sundial}, CoRA reduces MSE by 18.7\% compared to the second best adaptation method UniCA~\citep{han2025unica}, highlighting the importance of maintaining parameter consistency and equivalent initialization during adaptation. In the EPF task, CoRA reduces MSE by 9.4\% compared to TimeXer and by 6.4\% compared to AdaPTS~\citep{benechehab2025adapts}, further solidifying its position as a superior and generalized approach for uni-modal covariate-aware forecasting.

\begin{table}[tbp]
\vspace{-5pt}
\caption{Averaged results of the long-term covariate-aware forecasting. For all baselines, the look-back length $L$ is fixed at 2880. The reported performance is averaged over prediction horizons $S$ = \{96, 192, 336, 720\} and full results are provided in Table~\ref{tab:tslib_exo_full_result}. Dash (-) denotes out of memory.}
\vspace{-3pt}
\renewcommand{\arraystretch}{0.85} 
\centering
\resizebox{1\columnwidth}{!}{
\begin{threeparttable}
\begin{small}
\renewcommand{\multirowsetup}{\centering}
\setlength{\tabcolsep}{1.45pt}
\label{tab:tslib_exo_result}
\begin{tabular}{c|cc|cc|cc|cc|cc|cc|cc|cc|cc|cc}
    \toprule
    {\multirow{2}{*}{\scalebox{0.8}{Models}}} & 
    \multicolumn{2}{c}{\scalebox{0.75}{\textbf{CoRA}}} &
    \multicolumn{2}{c}{\scalebox{0.8}{AdaPTS}} &
    \multicolumn{2}{c}{\scalebox{0.8}{ChronosX}} &
    \multicolumn{2}{c}{\scalebox{0.8}{UniCA}} &
    \multicolumn{2}{c}{\scalebox{0.8}{TimeXer}} &
    \multicolumn{2}{c}{\scalebox{0.8}{iTransformer}} &
    \multicolumn{2}{c}{\scalebox{0.8}{PatchTST}} &
    \multicolumn{2}{c}{\scalebox{0.8}{NBEATSx}} &
    \multicolumn{2}{c}{\scalebox{0.8}{Crossformer}} &
    \multicolumn{2}{c}{\scalebox{0.8}{DLinear}} \\
    &
    \multicolumn{2}{c}{\scalebox{0.8}{\textbf{(Ours)}}} &
    \multicolumn{2}{c}{\scalebox{0.8}{\citeyearpar{benechehab2025adapts}}} & 
    \multicolumn{2}{c}{\scalebox{0.8}{\citeyearpar{arango2025chronosx}}} & 
    \multicolumn{2}{c}{\scalebox{0.8}{\citeyearpar{han2025unica}}} & 
    \multicolumn{2}{c}{\scalebox{0.8}{\citeyearpar{wang2024timexer}}} & 
    \multicolumn{2}{c}{\scalebox{0.8}{\citeyearpar{liu2023itransformer}}} & 
    \multicolumn{2}{c}{\scalebox{0.8}{\citeyearpar{nie2022time}}} & 
    \multicolumn{2}{c}{\scalebox{0.8}{\citeyearpar{olivares2023neural}}} & 
    \multicolumn{2}{c}{\scalebox{0.8}{\citeyearpar{zhang2023crossformer}}} & 
    \multicolumn{2}{c}{\scalebox{0.8}{\citeyearpar{zeng2023transformers}}} \\
    \cmidrule(lr){1-1} 
    \cmidrule(lr){2-3} \cmidrule(lr){4-5} \cmidrule(lr){6-7} 
    \cmidrule(lr){8-9} \cmidrule(lr){10-11} \cmidrule(lr){12-13} 
    \cmidrule(lr){14-15} \cmidrule(lr){16-17} \cmidrule(lr){18-19} 
    \cmidrule(lr){20-21} 
    \scalebox{0.8}{Metric} & 
    \scalebox{0.8}{MSE} & \scalebox{0.8}{MAE} &
    \scalebox{0.8}{MSE} & \scalebox{0.8}{MAE} &
    \scalebox{0.8}{MSE} & \scalebox{0.8}{MAE} &
    \scalebox{0.8}{MSE} & \scalebox{0.8}{MAE} &
    \scalebox{0.8}{MSE} & \scalebox{0.8}{MAE} &
    \scalebox{0.8}{MSE} & \scalebox{0.8}{MAE} &
    \scalebox{0.8}{MSE} & \scalebox{0.8}{MAE} &
    \scalebox{0.8}{MSE} & \scalebox{0.8}{MAE} &
    \scalebox{0.8}{MSE} & \scalebox{0.8}{MAE} &
    \scalebox{0.8}{MSE} & \scalebox{0.8}{MAE} \\
    \toprule
    
    \scalebox{0.8}{ETTh1} 
    & \boldres{\scalebox{0.8}{0.068}} & \boldres{\scalebox{0.8}{0.203}} 
    & \secondres{\scalebox{0.8}{0.076}} & \secondres{\scalebox{0.8}{0.211}} 
    & \scalebox{0.8}{0.085} & \scalebox{0.8}{0.227} 
    & \scalebox{0.8}{0.085} & \scalebox{0.8}{0.222} 
    & \scalebox{0.8}{0.089} & \scalebox{0.8}{0.240}
    & \scalebox{0.8}{0.160} & \scalebox{0.8}{0.317} 
    & \scalebox{0.8}{0.096} & \scalebox{0.8}{0.249}  
    & \scalebox{0.8}{0.181} & \scalebox{0.8}{0.351}
    & \scalebox{0.8}{0.386} & \scalebox{0.8}{0.501}
    & \scalebox{0.8}{0.263} & \scalebox{0.8}{0.408} \\
    \midrule
    
    \scalebox{0.8}{ETTh2} 
    & \boldres{\scalebox{0.8}{0.141}} & \boldres{\scalebox{0.8}{0.299}} 
    & \secondres{\scalebox{0.8}{0.156}} & \secondres{\scalebox{0.8}{0.311}} 
    & \scalebox{0.8}{0.365} & \scalebox{0.8}{0.466} 
    & \scalebox{0.8}{0.197} & \scalebox{0.8}{0.350} 
    & \scalebox{0.8}{0.194} & \scalebox{0.8}{0.355} 
    & \scalebox{0.8}{0.307} & \scalebox{0.8}{0.445} 
    & \scalebox{0.8}{0.191} & \scalebox{0.8}{0.352} 
    & \scalebox{0.8}{0.181} & \scalebox{0.8}{0.351}
    & \scalebox{0.8}{0.395} & \scalebox{0.8}{0.502} 
    & \scalebox{0.8}{0.320} & \scalebox{0.8}{0.454} \\
    \midrule
    
    \scalebox{0.8}{ETTm1} 
    & \boldres{\scalebox{0.8}{0.043}} & \boldres{\scalebox{0.8}{0.155}} 
    & \secondres{\scalebox{0.8}{0.046}} & \secondres{\scalebox{0.8}{0.165}} 
    & \scalebox{0.8}{0.049} & \secondres{\scalebox{0.8}{0.165}} 
    & \scalebox{0.8}{0.050} & \scalebox{0.8}{0.166} 
    & \scalebox{0.8}{0.062} & \scalebox{0.8}{0.192} 
    & \scalebox{0.8}{0.059} & \scalebox{0.8}{0.186} 
    & \scalebox{0.8}{0.055} & \scalebox{0.8}{0.181}
    & \scalebox{0.8}{0.112} & \scalebox{0.8}{0.268}
    & \scalebox{0.8}{0.068} & \scalebox{0.8}{0.207} 
    & \scalebox{0.8}{0.059} & \scalebox{0.8}{0.184} \\
    \midrule
    
    \scalebox{0.8}{ETTm2} 
    & \boldres{\scalebox{0.8}{0.100}} & \boldres{\scalebox{0.8}{0.237}} 
    & \scalebox{0.8}{0.107} & \secondres{\scalebox{0.8}{0.245}} 
    & \secondres{\scalebox{0.8}{0.106}} & \scalebox{0.8}{0.246} 
    & \scalebox{0.8}{0.122} & \scalebox{0.8}{0.265} 
    & \scalebox{0.8}{0.161} & \scalebox{0.8}{0.304} 
    & \scalebox{0.8}{0.149} & \scalebox{0.8}{0.304} 
    & \scalebox{0.8}{0.131} & \scalebox{0.8}{0.278} 
    & \scalebox{0.8}{0.222} & \scalebox{0.8}{0.384}
    & \scalebox{0.8}{0.208} & \scalebox{0.8}{0.366} 
    & \scalebox{0.8}{0.123} & \scalebox{0.8}{0.266} \\
    \midrule
    
    \scalebox{0.8}{Weather} 
    & \boldres{\scalebox{0.8}{0.001}} & \boldres{\scalebox{0.8}{0.026}} 
    & \secondres{\scalebox{0.8}{0.002}} & \secondres{\scalebox{0.8}{0.027}}  
    & \secondres{\scalebox{0.8}{0.002}} & \scalebox{0.8}{0.033} 
    & \secondres{\scalebox{0.8}{0.002}} & \scalebox{0.8}{0.033} 
    & \secondres{\scalebox{0.8}{0.002}} & \scalebox{0.8}{0.033}
    & \secondres{\scalebox{0.8}{0.002}} & \scalebox{0.8}{0.034} 
    & \secondres{\scalebox{0.8}{0.002}} & \scalebox{0.8}{0.036} 
    & \scalebox{0.8}{0.033} & \scalebox{0.8}{0.086}
    & \scalebox{0.8}{0.004} & \scalebox{0.8}{0.047} 
    & \scalebox{0.8}{0.008} & \scalebox{0.8}{0.076} \\
    \midrule

    \scalebox{0.8}{ECL} 
    & \boldres{\scalebox{0.8}{0.194}} & \boldres{\scalebox{0.8}{0.314}} 
    & \scalebox{0.8}{0.212} & \scalebox{0.8}{0.329} 
    & \secondres{\scalebox{0.8}{0.206}} & \secondres{\scalebox{0.8}{0.323}} 
    & \scalebox{0.8}{0.230} & \scalebox{0.8}{0.347} 
    & \scalebox{0.8}{0.292} & \scalebox{0.8}{0.387} 
    & \scalebox{0.8}{0.293} & \scalebox{0.8}{0.406} 
    & \scalebox{0.8}{0.327} & \scalebox{0.8}{0.431} 
    & \scalebox{0.8}{0.352} & \scalebox{0.8}{0.449}
    & \scalebox{0.8}{0.352} & \scalebox{0.8}{0.446} 
    & \scalebox{0.8}{0.264} & \scalebox{0.8}{0.376} \\
    \midrule
    
    \scalebox{0.8}{Traffic} 
    & \boldres{\scalebox{0.8}{0.112}} & \boldres{\scalebox{0.8}{0.186}} 
    & \scalebox{0.8}{-} & \scalebox{0.8}{-} 
    & \scalebox{0.8}{-} & \scalebox{0.8}{-} 
    & \secondres{\scalebox{0.8}{0.122}} & \secondres{\scalebox{0.8}{0.203}} 
    & \scalebox{0.8}{0.157} & \scalebox{0.8}{0.259} 
    & \scalebox{0.8}{0.139} & \scalebox{0.8}{0.232} 
    & \scalebox{0.8}{0.154} & \scalebox{0.8}{0.255}  
    & \scalebox{0.8}{0.222} & \scalebox{0.8}{0.328}
    & \scalebox{0.8}{0.274} & \scalebox{0.8}{0.332} 
    & \scalebox{0.8}{0.203} & \scalebox{0.8}{0.317} \\
    \bottomrule
\end{tabular}
\end{small}
\end{threeparttable}
}
\vspace{-6pt}
\end{table}

\begin{table}[tbp]
\vspace{-5pt}
\caption{Full results of the short-term covariate-aware forecasting. Following the standard protocol of EPF dataset, with input-output lengths of 168-24. Avg means the average results from all five datasets. Results of end-to-end models are officially reported by TimeXer~\citep{wang2024timexer}.}
\vspace{-3pt}
\renewcommand{\arraystretch}{0.85} 
\centering
\resizebox{1\columnwidth}{!}{
\begin{threeparttable}
\begin{small}
\renewcommand{\multirowsetup}{\centering}
\setlength{\tabcolsep}{1.45pt}
\label{tab:epf_result}
\begin{tabular}{c|cc|cc|cc|cc|cc|cc|cc|cc|cc|cc}
    \toprule
    {\multirow{2}{*}{\scalebox{0.8}{Models}}} & 
    \multicolumn{2}{c}{\scalebox{0.75}{\textbf{CoRA}}} &
    \multicolumn{2}{c}{\scalebox{0.8}{AdaPTS}} &
    \multicolumn{2}{c}{\scalebox{0.8}{UniCA}} &
    \multicolumn{2}{c}{\scalebox{0.8}{ChronosX}} &
    \multicolumn{2}{c}{\scalebox{0.8}{TimeXer}} &
    \multicolumn{2}{c}{\scalebox{0.8}{iTransformer}} &
    \multicolumn{2}{c}{\scalebox{0.8}{PatchTST}} &
    \multicolumn{2}{c}{\scalebox{0.8}{NBEATSx}} &
    \multicolumn{2}{c}{\scalebox{0.8}{Crossformer}} &
    \multicolumn{2}{c}{\scalebox{0.8}{DLinear}} \\
    &
    \multicolumn{2}{c}{\scalebox{0.8}{\textbf{(Ours)}}} &
    \multicolumn{2}{c}{\scalebox{0.8}{\citeyearpar{benechehab2025adapts}}} & 
    \multicolumn{2}{c}{\scalebox{0.8}{\citeyearpar{han2025unica}}} & 
    \multicolumn{2}{c}{\scalebox{0.8}{\citeyearpar{arango2025chronosx}}} & 
    \multicolumn{2}{c}{\scalebox{0.8}{\citeyearpar{wang2024timexer}}} & 
    \multicolumn{2}{c}{\scalebox{0.8}{\citeyearpar{liu2023itransformer}}} & 
    \multicolumn{2}{c}{\scalebox{0.8}{\citeyearpar{nie2022time}}} & 
    \multicolumn{2}{c}{\scalebox{0.8}{\citeyearpar{olivares2023neural}}} & 
    \multicolumn{2}{c}{\scalebox{0.8}{\citeyearpar{zhang2023crossformer}}} & 
    \multicolumn{2}{c}{\scalebox{0.8}{\citeyearpar{zeng2023transformers}}} \\
    \cmidrule(lr){1-1} 
    \cmidrule(lr){2-3} \cmidrule(lr){4-5} \cmidrule(lr){6-7} 
    \cmidrule(lr){8-9} \cmidrule(lr){10-11} \cmidrule(lr){12-13} 
    \cmidrule(lr){14-15} \cmidrule(lr){16-17} \cmidrule(lr){18-19} 
    \cmidrule(lr){20-21} 
    \scalebox{0.8}{Metric} & 
    \scalebox{0.8}{MSE} & \scalebox{0.8}{MAE} &
    \scalebox{0.8}{MSE} & \scalebox{0.8}{MAE} &
    \scalebox{0.8}{MSE} & \scalebox{0.8}{MAE} &
    \scalebox{0.8}{MSE} & \scalebox{0.8}{MAE} &
    \scalebox{0.8}{MSE} & \scalebox{0.8}{MAE} &
    \scalebox{0.8}{MSE} & \scalebox{0.8}{MAE} &
    \scalebox{0.8}{MSE} & \scalebox{0.8}{MAE} &
    \scalebox{0.8}{MSE} & \scalebox{0.8}{MAE} &
    \scalebox{0.8}{MSE} & \scalebox{0.8}{MAE} &
    \scalebox{0.8}{MSE} & \scalebox{0.8}{MAE} \\
    \toprule
    
    \scalebox{0.8}{NP} 
    & \boldres{\scalebox{0.8}{0.222}} & \boldres{\scalebox{0.8}{0.246}}
    & \secondres{\scalebox{0.8}{0.231}} & \secondres{\scalebox{0.8}{0.259}} 
    & \scalebox{0.8}{0.265} & \scalebox{0.8}{0.289} 
    & \scalebox{0.8}{0.254} & \scalebox{0.8}{0.278} 
    & \scalebox{0.8}{0.236} & \scalebox{0.8}{0.268}  
    & \scalebox{0.8}{0.265} & \scalebox{0.8}{0.300} 
    & \scalebox{0.8}{0.267} & \scalebox{0.8}{0.284} 
    & \scalebox{0.8}{0.272} & \scalebox{0.8}{0.301} 
    & \scalebox{0.8}{0.240} & \scalebox{0.8}{0.285} 
    & \scalebox{0.8}{0.309} & \scalebox{0.8}{0.321} \\
    \midrule
    
    \scalebox{0.8}{PJM} 
    & \boldres{\scalebox{0.8}{0.073}} & \boldres{\scalebox{0.8}{0.165}} 
    & \secondres{\scalebox{0.8}{0.080}} & \secondres{\scalebox{0.8}{0.173}}
    & \scalebox{0.8}{0.090} & \scalebox{0.8}{0.187} 
    & \scalebox{0.8}{0.089} & \scalebox{0.8}{0.189} 
    & \scalebox{0.8}{0.093} & \scalebox{0.8}{0.192} 
    & \scalebox{0.8}{0.097} & \scalebox{0.8}{0.197}  
    & \scalebox{0.8}{0.106} & \scalebox{0.8}{0.209} 
    & \scalebox{0.8}{0.097} & \scalebox{0.8}{0.189}
    & \scalebox{0.8}{0.101} & \scalebox{0.8}{0.199} 
    & \scalebox{0.8}{0.108} & \scalebox{0.8}{0.215} \\
    \midrule
    
    \scalebox{0.8}{BE} 
    & \boldres{\scalebox{0.8}{0.339}} & \boldres{\scalebox{0.8}{0.236}}
    & \secondres{\scalebox{0.8}{0.355}} & \scalebox{0.8}{0.261}
    & \scalebox{0.8}{0.368} & \scalebox{0.8}{0.273}
    & \scalebox{0.8}{0.371} & \scalebox{0.8}{0.274}
    & \scalebox{0.8}{0.379} & \secondres{\scalebox{0.8}{0.243}}  
    & \scalebox{0.8}{0.394} & \scalebox{0.8}{0.270} 
    & \scalebox{0.8}{0.400} & \scalebox{0.8}{0.262} 
    & \scalebox{0.8}{0.389} & \scalebox{0.8}{0.265} 
    & \scalebox{0.8}{0.420} & \scalebox{0.8}{0.290} 
    & \scalebox{0.8}{0.463} & \scalebox{0.8}{0.313} \\
    \midrule
    
    \scalebox{0.8}{FR} 
    & \boldres{\scalebox{0.8}{0.357}} & \boldres{\scalebox{0.8}{0.206}}
    & \scalebox{0.8}{0.363} & \scalebox{0.8}{0.218}
    & \scalebox{0.8}{0.365} & \scalebox{0.8}{0.218}
    & \secondres{\scalebox{0.8}{0.361}} & \scalebox{0.8}{0.217}
    & \scalebox{0.8}{0.385} & \secondres{\scalebox{0.8}{0.208}} 
    & \scalebox{0.8}{0.439} & \scalebox{0.8}{0.233} 
    & \scalebox{0.8}{0.411} & \scalebox{0.8}{0.220} 
    & \scalebox{0.8}{0.393} & \scalebox{0.8}{0.211}
    & \scalebox{0.8}{0.434} & \secondres{\scalebox{0.8}{0.208}} 
    & \scalebox{0.8}{0.429} & \scalebox{0.8}{0.260} \\
    \midrule
    
    \scalebox{0.8}{DE} 
    & \boldres{\scalebox{0.8}{0.401}} & \boldres{\scalebox{0.8}{0.388}}
    & \scalebox{0.8}{0.455} & \scalebox{0.8}{0.424}
    & \scalebox{0.8}{0.553} & \scalebox{0.8}{0.466}
    & \scalebox{0.8}{0.453} & \scalebox{0.8}{0.426}
    & \secondres{\scalebox{0.8}{0.440}} & \secondres{\scalebox{0.8}{0.415}}  
    & \scalebox{0.8}{0.479} & \scalebox{0.8}{0.443} 
    & \scalebox{0.8}{0.461} & \scalebox{0.8}{0.432} 
    & \scalebox{0.8}{0.499} & \scalebox{0.8}{0.447} 
    & \scalebox{0.8}{0.574} & \scalebox{0.8}{0.430} 
    & \scalebox{0.8}{0.520} & \scalebox{0.8}{0.463} \\
    \midrule
    
    \scalebox{0.8}{AVG} 
    & \boldres{\scalebox{0.8}{0.278}} & \boldres{\scalebox{0.8}{0.248}}
    & \secondres{\scalebox{0.8}{0.297}} & \scalebox{0.8}{0.267}
    & \scalebox{0.8}{0.328} & \scalebox{0.8}{0.287}
    & \scalebox{0.8}{0.306} & \scalebox{0.8}{0.277}
    & \scalebox{0.8}{0.307} & \secondres{\scalebox{0.8}{0.265}}  
    & \scalebox{0.8}{0.335} & \scalebox{0.8}{0.289} 
    & \scalebox{0.8}{0.330} & \scalebox{0.8}{0.282} 
    & \scalebox{0.8}{0.330} & \scalebox{0.8}{0.283} 
    & \scalebox{0.8}{0.354} & \scalebox{0.8}{0.284} 
    & \scalebox{0.8}{0.366} & \scalebox{0.8}{0.314} \\
    \bottomrule
\end{tabular}
\end{small}
\end{threeparttable}
}
\vspace{-6pt}
\end{table}

\subsubsection{Multi-Modal Covariate-Aware Forecasting}

\paragraph{Setups} 
We evaluate CoRA on tasks involving multi-modal covariates, specifically images and text. For image-based covariates, we construct a subset from the RT-1~\citep{brohan2022rt} dataset, which contains a target time series with image covariates at each timestamp. For text-based covariates, we choose the Time-MMD~\citep{liu2024time} dataset, which includes a target time series with a corresponding text covariate. Moreover, CoRA adopts ViT\footnote{\href{https://huggingface.co/google/vit-base-patch16-224-in21k}{https://huggingface.co/google/vit-base-patch16-224-in21k}.}~\citep{wu2020visual} and Qwen3-Embedding\footnote{\href{https://huggingface.co/Qwen/Qwen3-Embedding-0.6B}{https://huggingface.co/Qwen/Qwen3-Embedding-0.6B}.}~\citep{qwen3embedding} as backbone to extract features from image and text respectively.

\begin{figure*}[tbp]
\begin{center}
	\centerline{\includegraphics[width=\columnwidth]{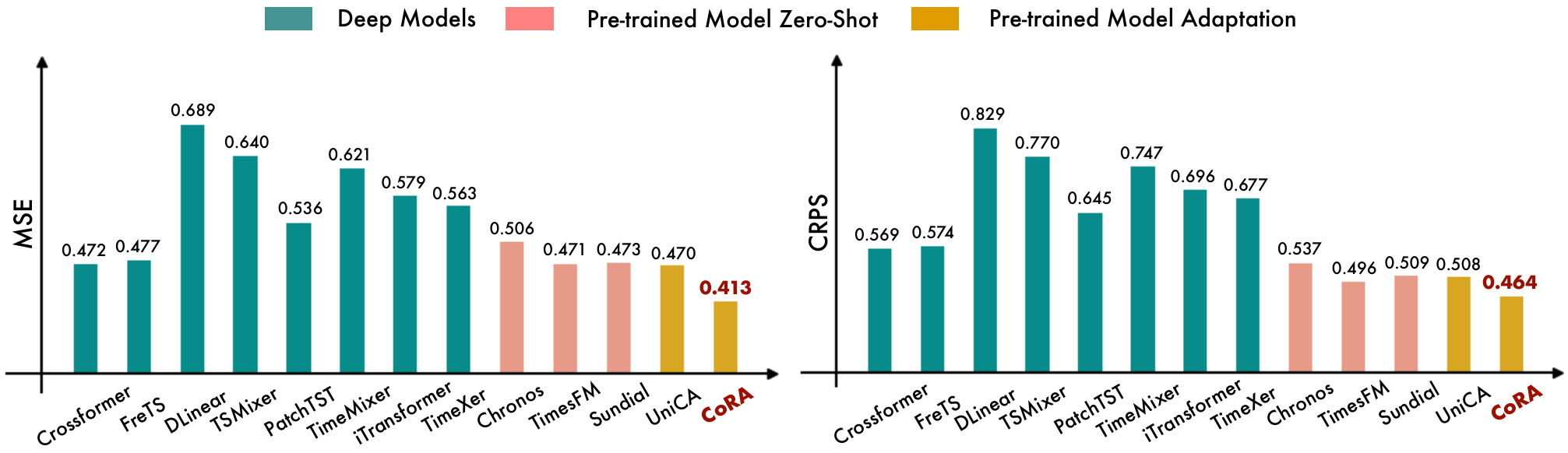}}
        \vspace{-7pt}
	\caption{Multi-modal covariate-aware forecasting on a subset of RT-1~\citep{brohan2022rt} with a time series target variate and an image covariate. Input length is set to 32 and prediction length is 4.}
	\label{fig:rt_1_result}
\end{center}
\vspace{-20pt}
\end{figure*}

\paragraph{Results}
As shown in Figure~\ref{fig:rt_1_result} and Table~\ref{tab:average_result_timemmd}, CoRA achieves state-of-the-art performance across all metrics. On the RT-1~\citep{brohan2022rt} dataset, CoRA outperforms the best end-to-end supervised model and TSFM zero-shot by 12.7\% in MSE and 8.8\% in CRPS. While on the Time-MMD benchmark~\citep{liu2024time}, the improvements are 1.9\% in MSE and 3.7\% in CRPS. These results demonstrate that properly modeling auxiliary modalities provides substantial benefits for forecasting. Compared with UniCA~\citep{han2025unica}, which does not maintain backbone consistency or use proper zero-initialization, CoRA consistently achieves superior performance on both benchmarks.

\begin{table}[tbp]
\vspace{-5pt}
  \caption{Multi-modal covariate-aware forecasting on Time-MMD~\citep{liu2024time} with textual covariates. Baseline results are reported by UniCA~\citep{han2025unica}, with full results in Table~\ref{tab:full_result_timemmd}.}
  \label{tab:average_result_timemmd}
  \centering
  \renewcommand{\arraystretch}{1.3}
  \resizebox{\columnwidth}{!}{
  \small 

  \begin{tabular}{@{}ll| *{11}{c}@{}} 
    \toprule
    & \multirow{2}{*}{Models} & 
    \multicolumn{1}{c}{\textbf{CoRA}} &
    \multicolumn{1}{c}{UniCA} &
    \multicolumn{1}{c}{Sundial} &
    \multicolumn{1}{c}{Moirai} &
    \multicolumn{1}{c}{TabPFN-TS} &
    \multicolumn{1}{c}{PatchTST} &
    \multicolumn{1}{c}{TTM} &
    \multicolumn{1}{c}{TiDE} &
    \multicolumn{1}{c}{N-BEATS} &
    \multicolumn{1}{c}{TFT} &
    \multicolumn{1}{c}{DeepAR}  \\
    & & 
    \textbf{(Ours)} &
    \citeyearpar{han2025unica} &
    \citeyearpar{liu2025sundial} &
    \citeyearpar{woo2024unified} &
    \citeyearpar{hoo2025tables} &
    \citeyearpar{nie2022time} &
    \citeyearpar{ekambaram2024tiny} &
    \citeyearpar{das2023long} &
    \citeyearpar{olivares2023neural} &
    \citeyearpar{lim2021temporal} &
    \citeyearpar{salinas2020deepar} \\
    \toprule 

    & Average
    & \boldres{0.641} & \secondres{0.661} & 0.662 & 0.751 
    & 0.795 & 0.933 & 0.820 & 0.927 
    & 0.882 & 0.947 & 1.361 \\
    & MSE
    & \boldres{0.580} & \secondres{0.591} & \secondres{0.591} & 0.696 
    & 0.787 & 0.793 & 0.685 & 0.869 
    & 0.782 & 0.992 & 1.605 \\
    & MAE
    & \boldres{0.690} & \secondres{0.716} & \secondres{0.716} & 0.821 
    & 0.837 & 1.009 & 0.866 & 0.976 
    & 0.884 & 0.958 & 1.219 \\
    & CRPS
    & \boldres{0.653} & \secondres{0.677} & 0.678 & 0.735 
    & 0.762 & 0.996 & 0.909 & 0.937 
    & 0.980 & 0.891 & 1.260 \\
    \bottomrule

    \end{tabular}
    }
\vspace{-6pt}
\end{table}

\subsubsection{Few-Shot Forecasting}

\paragraph{Setups} In real-world applications, the available training data is often highly limited, making few-shot forecasting a critical challenge for robust deployment. We evaluate CoRA on the well-established electricity price forecasting (EPF) task~\citep{lago2021forecasting}, comparing it with alternative adaptation methods and end-to-end models across a range of data scarcity levels.

\begin{figure*}[htbp]
\begin{center}
	\centerline{\includegraphics[width=\columnwidth]{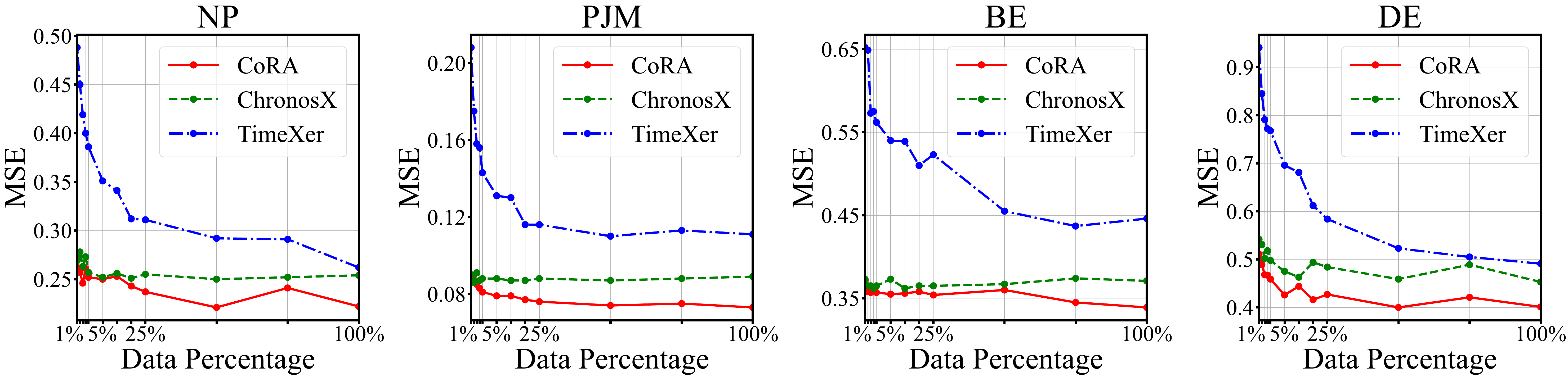}}
        \vspace{-7pt}
	\caption{Few-shot forecasting on the EPF dataset, comparing CoRA with TimeXer~\citep{wang2024timexer} and ChronosX~\citep{arango2025chronosx} across different levels of data availability.}
	\label{fig:few_shot_forecasting}
\end{center}
\vspace{-20pt}
\end{figure*}

\paragraph{Results}
As shown in Figure~\ref{fig:few_shot_forecasting}, CoRA consistently outperforms TimeXer~\citep{wang2024timexer} and ChronosX~\citep{arango2025chronosx} under different data availability levels. When the number of samples is particularly small (1\% to 25\%), the end-to-end model TimeXer performs significantly worse than adaptation methods based on pre-trained TSFMs, highlighting that pre-trained models can adapt to downstream tasks more quickly and effectively with limited data. Even with sufficient data, TimeXer still underperforms compared with adaptation methods, due to its relatively smaller model capacity. Moreover, thanks to principled designs that preserve the pre-trained backbone and employ proper zero-initialization, CoRA consistently outperforms ChronosX across different data percentage.

\subsubsection{Multivariate Time Series Forecasting}

\paragraph{Setups}
CoRA naturally extends to the multivariate time series forecasting scenarios via the channel-independence mechanism, enabling joint prediction of multiple target variates. We evaluate this on seven real-world datasets introduced in Autoformer~\citep{wu2021autoformer}.

\paragraph{Results}
As shown in Table~\ref{tab:tslib_multivar_result}, CoRA outperforms all other supervised forecasters, achieving average MSE and MAE reductions of 14.5\% and 12.2\% compared to TimeXer~\citep{wang2024timexer}. CoRA's superior performance stems from its use of pre-trained TSFMs that have already internalized universal temporal patterns from large-scale datasets. This enables CoRA to more accurately capture inter-variate dependencies and generalize effectively across diverse datasets.

\begin{table}[htbp]
\vspace{-5pt}
  \caption{Averaged results of the multivariate forecasting task on well-acknowledged benchmarks. For all baselines, the look-back length $L$ is fixed at 2880. The reported performance is averaged over prediction horizons $S$ = \{96, 192, 336, 720\} and full results are provided in Table~\ref{tab:tslib_multivar_full_result}.}
  \vspace{-3pt}
  \renewcommand{\arraystretch}{0.85} 
  \centering
  \resizebox{1\columnwidth}{!}{
  \begin{threeparttable}
  \begin{small}
  \renewcommand{\multirowsetup}{\centering}
  \setlength{\tabcolsep}{1.45pt}
  \label{tab:tslib_multivar_result}
  \begin{tabular}{c|cc|cc|cc|cc|cc|cc|cc|cc|cc|cc}
    \toprule
    {\multirow{2}{*}{\scalebox{0.8}{Models}}} & 
    \multicolumn{2}{c}{\rotatebox{0}{\scalebox{0.75}{\textbf{CoRA}}}} &
    \multicolumn{2}{c}{\rotatebox{0}{\scalebox{0.8}{Timer-XL}}} &
    \multicolumn{2}{c}{\rotatebox{0}{\scalebox{0.8}{TimeXer}}} &
    \multicolumn{2}{c}{\rotatebox{0}{\scalebox{0.8}{iTransformer}}} &
    \multicolumn{2}{c}{\rotatebox{0}{\scalebox{0.8}{{PatchTST}}}} &
    \multicolumn{2}{c}{\rotatebox{0}{\scalebox{0.8}{Crossformer}}} &
    \multicolumn{2}{c}{\rotatebox{0}{\scalebox{0.8}{TiDE}}} &
    \multicolumn{2}{c}{\rotatebox{0}{\scalebox{0.8}{DLinear}}} &
    \multicolumn{2}{c}{\rotatebox{0}{\scalebox{0.8}{SCINet}}} &
    \multicolumn{2}{c}{\rotatebox{0}{\scalebox{0.8}{Autoformer}}}  \\
    &
    \multicolumn{2}{c}{\scalebox{0.8}{\textbf{(Ours)}}} &
    \multicolumn{2}{c}{\scalebox{0.8}{\citeyearpar{liu2024timer}}} & 
    \multicolumn{2}{c}{\scalebox{0.8}{\citeyearpar{wang2024timexer}}} & 
    \multicolumn{2}{c}{\scalebox{0.8}{\citeyearpar{liu2023itransformer}}} & 
    \multicolumn{2}{c}{\scalebox{0.8}{\citeyearpar{nie2022time}}} & 
    \multicolumn{2}{c}{\scalebox{0.8}{\citeyearpar{zhang2023crossformer}}} & 
    \multicolumn{2}{c}{\scalebox{0.8}{\citeyearpar{das2023long}}} &
    \multicolumn{2}{c}{\scalebox{0.8}{\citeyearpar{zeng2023transformers}}} &
    \multicolumn{2}{c}{\scalebox{0.8}{\citeyearpar{liu2022scinet}}} &
    \multicolumn{2}{c}{\scalebox{0.8}{\citeyearpar{wu2021autoformer}}}  \\
    \cmidrule(lr){1-1} \cmidrule(lr){2-3} \cmidrule(lr){4-5}\cmidrule(lr){6-7} \cmidrule(lr){8-9}\cmidrule(lr){10-11}\cmidrule(lr){12-13} \cmidrule(lr){14-15} \cmidrule(lr){16-17} \cmidrule(lr){18-19} \cmidrule(lr){20-21} 
    \scalebox{0.8}{Metric}
    & \scalebox{0.8}{MSE} & \scalebox{0.8}{MAE} 
    & \scalebox{0.8}{MSE} & \scalebox{0.8}{MAE}  
    & \scalebox{0.8}{MSE} & \scalebox{0.8}{MAE}  
    & \scalebox{0.8}{MSE} & \scalebox{0.8}{MAE}  
    & \scalebox{0.8}{MSE} & \scalebox{0.8}{MAE}  
    & \scalebox{0.8}{MSE} & \scalebox{0.8}{MAE} 
    & \scalebox{0.8}{MSE} & \scalebox{0.8}{MAE} 
    & \scalebox{0.8}{MSE} & \scalebox{0.8}{MAE} 
    & \scalebox{0.8}{MSE} & \scalebox{0.8}{MAE} 
    & \scalebox{0.8}{MSE} & \scalebox{0.8}{MAE} \\
    \toprule
    
    \scalebox{0.8}{ETTh1} 
    & \boldres{\scalebox{0.8}{0.404}} & \boldres{\scalebox{0.8}{0.422}} 
    & \scalebox{0.8}{0.548} & \scalebox{0.8}{0.547} 
    & \secondres{\scalebox{0.8}{0.492}} & \secondres{\scalebox{0.8}{0.488}} 
    & \scalebox{0.8}{0.508} & \scalebox{0.8}{0.515} 
    & \scalebox{0.8}{0.516} & \scalebox{0.8}{0.504} 
    & \scalebox{0.8}{0.643} & \scalebox{0.8}{0.594} 
    & \scalebox{0.8}{0.656} & \scalebox{0.8}{0.587}
    & \scalebox{0.8}{0.519} & \scalebox{0.8}{0.512} 
    & \scalebox{0.8}{0.780} & \scalebox{0.8}{0.660}
    & \scalebox{0.8}{0.812} & \scalebox{0.8}{0.661} \\ 
    \midrule
    
    \scalebox{0.8}{ETTh2}
    & \boldres{\scalebox{0.8}{0.331}} & \boldres{\scalebox{0.8}{0.381}} 
    & \secondres{\scalebox{0.8}{0.422}} & \secondres{\scalebox{0.8}{0.454}} 
    & \scalebox{0.8}{0.454} & \scalebox{0.8}{0.476} 
    & \scalebox{0.8}{0.440} & \scalebox{0.8}{0.476} 
    & \scalebox{0.8}{0.490} & \scalebox{0.8}{0.503} 
    & \scalebox{0.8}{0.810} & \scalebox{0.8}{0.691} 
    & \scalebox{0.8}{0.555} & \scalebox{0.8}{0.532}
    & \scalebox{0.8}{0.620} & \scalebox{0.8}{0.589} 
    & \scalebox{0.8}{0.667} & \scalebox{0.8}{0.592}
    & \scalebox{0.8}{0.840} & \scalebox{0.8}{0.707} \\ 
    \midrule
    
    \scalebox{0.8}{ETTm1}
    & \boldres{\scalebox{0.8}{0.337}} & \boldres{\scalebox{0.8}{0.371}} 
    & \scalebox{0.8}{0.381} & \scalebox{0.8}{0.419} 
    & \scalebox{0.8}{0.398} & \scalebox{0.8}{0.424} 
    & \scalebox{0.8}{0.379} & \scalebox{0.8}{0.413} 
    & \scalebox{0.8}{0.400} & \scalebox{0.8}{0.424} 
    & \scalebox{0.8}{0.436} & \scalebox{0.8}{0.457} 
    & \scalebox{0.8}{0.363} & \scalebox{0.8}{0.393}
    & \secondres{\scalebox{0.8}{0.357}} & \secondres{\scalebox{0.8}{0.387} }
    & \scalebox{0.8}{0.425} & \scalebox{0.8}{0.447}
    & \scalebox{0.8}{0.857} & \scalebox{0.8}{0.682} \\ 
    \midrule
    
    \scalebox{0.8}{ETTm2}
    & \boldres{\scalebox{0.8}{0.256}} & \boldres{\scalebox{0.8}{0.317}}
    & \scalebox{0.8}{0.318} & \scalebox{0.8}{0.383} 
    & \scalebox{0.8}{0.274} & \scalebox{0.8}{0.343} 
    & \scalebox{0.8}{0.276} & \scalebox{0.8}{0.342} 
    & \scalebox{0.8}{0.292} & \scalebox{0.8}{0.355} 
    & \scalebox{0.8}{0.569} & \scalebox{0.8}{0.593} 
    & \scalebox{0.8}{0.306} & \scalebox{0.8}{0.370}
    & \secondres{\scalebox{0.8}{0.266}} & \secondres{\scalebox{0.8}{0.335}} 
    & \scalebox{0.8}{0.308} & \scalebox{0.8}{0.378}
    & \scalebox{0.8}{0.457} & \scalebox{0.8}{0.495} \\ 
    \midrule
    
    \scalebox{0.8}{Weather}
    & \boldres{\scalebox{0.8}{0.230}} & \boldres{\scalebox{0.8}{0.269}} 
    & \scalebox{0.8}{0.316} & \scalebox{0.8}{0.348} 
    & \scalebox{0.8}{0.262} & \scalebox{0.8}{0.303} 
    & \scalebox{0.8}{0.251} & \scalebox{0.8}{0.305} 
    & \scalebox{0.8}{0.251} & \scalebox{0.8}{0.290} 
    & \scalebox{0.8}{0.235} & \scalebox{0.8}{0.285} 
    & \secondres{\scalebox{0.8}{0.234}} & \secondres{\scalebox{0.8}{0.281}}
    & \scalebox{0.8}{0.237} & \scalebox{0.8}{0.291} 
    & \scalebox{0.8}{0.249} & \scalebox{0.8}{0.296}
    & \scalebox{0.8}{0.500} & \scalebox{0.8}{0.487} \\ 
    \midrule

    \scalebox{0.8}{ECL} 
    & \boldres{\scalebox{0.8}{0.155}} & \boldres{\scalebox{0.8}{0.250}} 
    & \boldres{\scalebox{0.8}{0.155}} & \secondres{\scalebox{0.8}{0.252}} 
    & \scalebox{0.8}{0.172} & \scalebox{0.8}{0.275} 
    & \scalebox{0.8}{0.194} & \scalebox{0.8}{0.299} 
    & \scalebox{0.8}{0.163} & \scalebox{0.8}{0.265} 
    & \scalebox{0.8}{0.184} & \scalebox{0.8}{0.281} 
    & \scalebox{0.8}{0.160} & \scalebox{0.8}{0.254}
    & \secondres{\scalebox{0.8}{0.156}} & \scalebox{0.8}{0.255} 
    & \scalebox{0.8}{0.181} & \scalebox{0.8}{0.285}
    & \scalebox{0.8}{0.292} & \scalebox{0.8}{0.390} \\ 
    \midrule
    
    \scalebox{0.8}{Traffic}
    & \boldres{\scalebox{0.8}{0.384}} & \boldres{\scalebox{0.8}{0.265}} 
    & \scalebox{0.8}{0.597} & \scalebox{0.8}{0.510} 
    & \secondres{\scalebox{0.8}{0.401}} & \scalebox{0.8}{0.281}
    & \scalebox{0.8}{0.407} & \scalebox{0.8}{0.291} 
    & \scalebox{0.8}{0.422} & \scalebox{0.8}{0.298} 
    & \scalebox{0.8}{0.522} & \scalebox{0.8}{0.285} 
    & \scalebox{0.8}{0.402} & \secondres{\scalebox{0.8}{0.276}}
    & \scalebox{0.8}{0.406} & \scalebox{0.8}{0.284} 
    & \scalebox{0.8}{0.478} & \scalebox{0.8}{0.352}
    & \scalebox{0.8}{0.742} & \scalebox{0.8}{0.464} \\ 
    \bottomrule
  \end{tabular}
    \end{small}
  \end{threeparttable}
}
\vspace{-6pt}
\end{table}

\subsection{Model Analysis}
In this section, we perform thorough experiments to analyze several properties of CoRA, including its generalization to other TSFMs such as TimesFM~\citep{das2023decoder}, Chronos-bolt~\citep{ansari2024chronos}, and FlowState~\citep{graf2025flowstate}, ablation studies on the method’s key components, and the interpretability of learned Granger Causality Embedding.

\begin{figure*}[htbp]
\begin{center}
	\centerline{\includegraphics[width=\columnwidth]{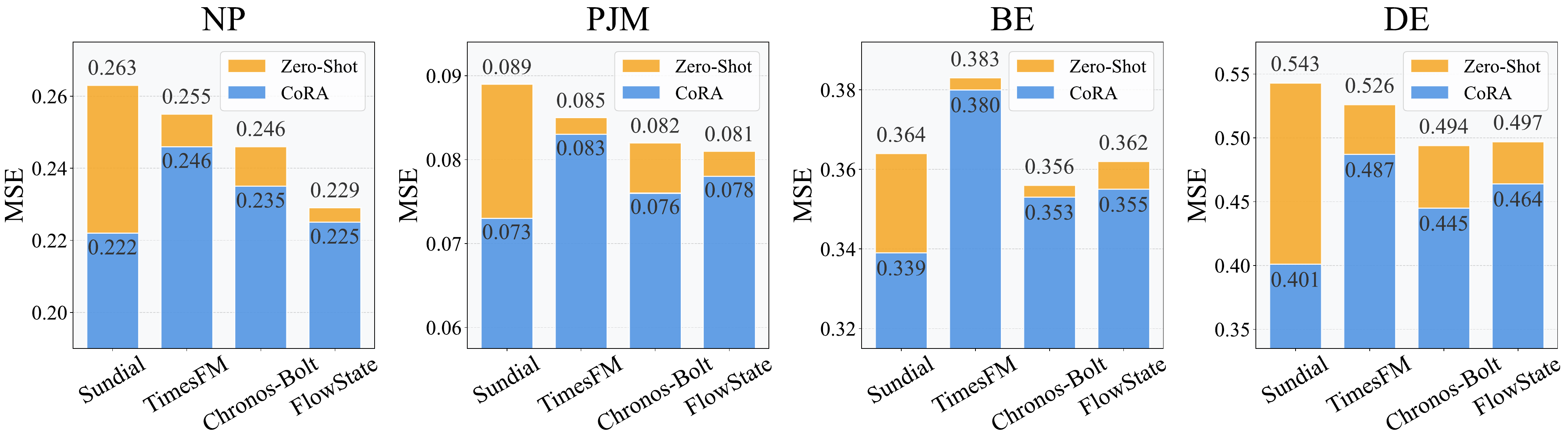}}
        \vspace{-7pt}
	\caption{Performance gains of CoRA across diverse TSFMs. Full results are provided in Table~\ref{tab:generality}.}
	\label{fig:generality}
\end{center}
\vspace{-20pt}
\end{figure*}

\paragraph{Generality}
Figure~\ref{fig:generality} shows that CoRA further boosts the performance of various TSFMs on top of their zero-shot results. Average MSE reductions are 14.2\% on Sundial~\citep{liu2025sundial}, 3.3\% on TimesFM~\citep{das2024decoder}, 4.9\% on Chronos-Bolt~\citep{ansari2024chronos}, and 3.3\% on FlowState~\citep{graf2025flowstate}. These results demonstrate that CoRA offers an effective and flexible adaptation strategy, seamlessly integrating with diverse backbone architectures.

\paragraph{Ablation Study}
We provide a thorough ablation study to examine our proposed CoRA in Table~\ref{tab:ablation}. Our results show that each component is crucial for CoRA's performance by addressing specific challenges in covariate-aware time series forecasting. Without the covariates' information, forecasting performance degrades, underscoring the necessity of incorporating external signals to enhance the predictability of the target. Without the adaLN module, we find that simply adding the condition to the TSFM head input is insufficient. Instead, our condition-injection mechanism is highly effective by influencing the statistics of the TSFM head to fuse information. Similarly, when we removed the Granger Causality Embedding, replacing it with mean aggregation, the model's performance dropped. This demonstrates the importance of our selection and routing mechanism, which automatically assigns appropriate weights to different covariates based on their inherent causality. Finally, we observed that replacing zero-initialization with Xavier initialization resulted in worse performance. This confirms that zero-initialization is vital for preserving the valuable knowledge learned during pre-training and ensuring a stable adaptation process. 

\begin{table}[htbp]
\vspace{-5pt}
  \caption{Ablation study of CoRA. (1) \textit{w/o} covariate denotes  Supervised Fine-Tuning (SFT), trained without using covariates. (2) \textit{w/o} adaLN replaces the adaLN module by directly adding the condition to the input of the TSFM head. (3) \textit{w/o} selection replaces the Granger Causality Embedding with mean aggregation. (4) \textit{w/o} zero-init replaces zero-initialization with Xavier initialization.}
  \label{tab:ablation}
  \centering
  \begin{threeparttable}
  \begin{small}
  \renewcommand{\multirowsetup}{\centering}
  \setlength{\tabcolsep}{4.2pt}
  \resizebox{1\columnwidth}{!}{
  \begin{tabular}{c|cc|cc|cc|cc|cc|cc}
    \toprule
    {Datasets}
    & \multicolumn{2}{c}{NP} & \multicolumn{2}{c}{PJM} & \multicolumn{2}{c}{BE}  
    & \multicolumn{2}{c}{FR} & \multicolumn{2}{c}{DE} & \multicolumn{2}{c}{Avg} \\
    \cmidrule(lr){2-3} \cmidrule(lr){4-5} \cmidrule(lr){6-7} 
    \cmidrule(lr){8-9}\cmidrule(lr){10-11} \cmidrule(lr){12-13} 
    {Models}
    & MSE & MAE & MSE & MAE & MSE & MAE 
    & MSE & MAE & MSE & MAE & MSE & MAE \\
    \toprule
    \textbf{CoRA}
    & \textbf{0.222} & \textbf{0.246} & \textbf{0.073}
    & \textbf{0.165} & \textbf{0.339} & \textbf{0.236} 
    & \textbf{0.357} & \textbf{0.206} & \textbf{0.401} 
    & \textbf{0.388} & \textbf{0.278} & \textbf{0.248} \\
    \midrule
    w/o covariate
    & 0.231 & 0.256 & 0.078 & 0.172
    & 0.352 & 0.262 & 0.360 & 0.214
    & 0.458 & 0.426 & 0.296 & 0.266\\
    \midrule
    w/o adaLN
    & 0.260 & 0.288 & 0.085 & 0.180
    & 0.351 & 0.238 & 0.368 & 0.210
    & 0.506 & 0.451 & 0.314 & 0.273 \\
    \midrule
    w/o selection
    & 0.273 & 0.266 & 0.080 & 0.177
    & 0.356 & 0.262 & 0.360 & 0.215
    & 0.472 & 0.423 & 0.301 & 0.269\\
    \midrule
    w/o zero-init
    & 0.234 & 0.262 & 0.078 & 0.173
    & 0.350 & 0.257 & 0.360 & 0.208
    & 0.430 & 0.415 & 0.290 & 0.263\\
    \bottomrule
  \end{tabular}}
  \end{small}
  \end{threeparttable}
  \vspace{-6pt}
\end{table}

\begin{figure*}[htbp]
\begin{center}
	\centerline{\includegraphics[width=\columnwidth]{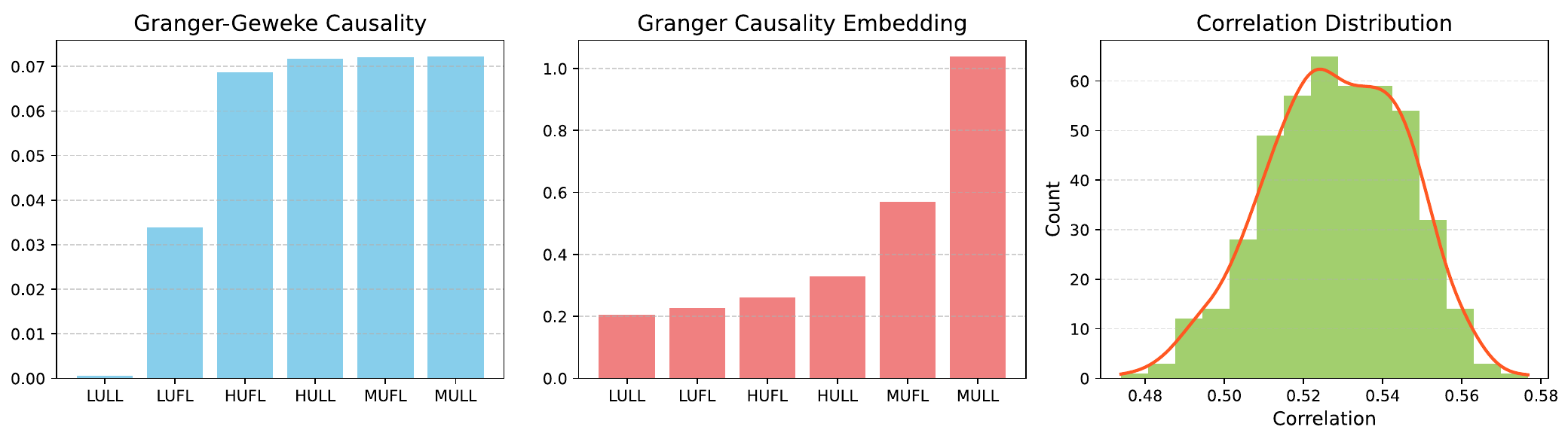}}
        \vspace{-7pt}
	\caption{Correlation between traditional statistic Granger-Geweke Causality~\citep{dhamala2018granger} and the Granger Causality Embedding learned by CoRA on ETTh1 Dataset.}
	\label{fig:granger}
\end{center}
\vspace{-20pt}
\end{figure*}

\paragraph{Interpretability}\label{sec:interp}
To study the interpretability of CoRA, we compare the learned Granger Causality Embedding with the traditional Granger-Geweke Causality~\citep{dhamala2018granger}. We select 1000 windows from the ETTh1 dataset and compute the Granger-Geweke Causality for each window (detailed description in the Algorithm~\ref{alg:Granger}) as well as the Granger Causality Embedding learned by CoRA. Figure~\ref{fig:granger} demonstrates a strong correlation between the Granger–Geweke Causality and the Granger Causality Embedding. Furthermore, we plot a histogram of the Pearson correlation coefficient~\citep{pearson1895vii} across the 1000 windows, which clearly demonstrates their consistency.

\section{Conclusion}
In this paper, we introduce CoRA, a general, flexible, and interpretable framework for adapting pre-trained foundation models to covariate-aware forecasting tasks. An important paradigm of foundation models involves large-scale pre-training on general datasets followed by adaptation to task-specific datasets. CoRA leverages this paradigm by using the powerful backbones of diverse foundation models as frozen embedding extractors. It then employs a Granger Causality Embedding to weight and select covariates based on their causal relationship to the target variate, and a zero-initialized adaLN module for stable and progressive fusion of this information. Our extensive experiments consistently show that CoRA outperforms both advanced supervised models and other adaptation methods while requiring fewer training samples, bridging the gap between powerful pre-trained models and the complex multi-modal and multivariate challenges of real-world scenarios.

\bibliography{iclr2026_conference}

\begin{thebibliography}{50}
\providecommand{\natexlab}[1]{#1}
\providecommand{\url}[1]{\texttt{#1}}
\expandafter\ifx\csname urlstyle\endcsname\relax
  \providecommand{\doi}[1]{doi: #1}\else
  \providecommand{\doi}{doi: \begingroup \urlstyle{rm}\Url}\fi

\bibitem[Ansari et~al.(2024)Ansari, Stella, Turkmen, Zhang, Mercado, Shen, Shchur, Rangapuram, Arango, Kapoor, et~al.]{ansari2024chronos}
Abdul~Fatir Ansari, Lorenzo Stella, Caner Turkmen, Xiyuan Zhang, Pedro Mercado, Huibin Shen, Oleksandr Shchur, Syama~Sundar Rangapuram, Sebastian~Pineda Arango, Shubham Kapoor, et~al.
\newblock Chronos: Learning the language of time series.
\newblock \emph{arXiv preprint arXiv:2403.07815}, 2024.

\bibitem[Arango et~al.(2025)Arango, Mercado, Kapoor, Ansari, Stella, Shen, Senetaire, Turkmen, Shchur, Maddix, et~al.]{arango2025chronosx}
Sebastian~Pineda Arango, Pedro Mercado, Shubham Kapoor, Abdul~Fatir Ansari, Lorenzo Stella, Huibin Shen, Hugo Senetaire, Caner Turkmen, Oleksandr Shchur, Danielle~C Maddix, et~al.
\newblock Chronosx: Adapting pretrained time series models with exogenous variables.
\newblock \emph{arXiv preprint arXiv:2503.12107}, 2025.

\bibitem[Benechehab et~al.(2025)Benechehab, Feofanov, Paolo, Thomas, Filippone, and K{\'e}gl]{benechehab2025adapts}
Abdelhakim Benechehab, Vasilii Feofanov, Giuseppe Paolo, Albert Thomas, Maurizio Filippone, and Bal{\'a}zs K{\'e}gl.
\newblock Adapts: Adapting univariate foundation models to probabilistic multivariate time series forecasting.
\newblock \emph{arXiv preprint arXiv:2502.10235}, 2025.

\bibitem[Brohan et~al.(2022)Brohan, Brown, Carbajal, Chebotar, Dabis, Finn, Gopalakrishnan, Hausman, Herzog, Hsu, et~al.]{brohan2022rt}
Anthony Brohan, Noah Brown, Justice Carbajal, Yevgen Chebotar, Joseph Dabis, Chelsea Finn, Keerthana Gopalakrishnan, Karol Hausman, Alex Herzog, Jasmine Hsu, et~al.
\newblock Rt-1: Robotics transformer for real-world control at scale.
\newblock \emph{arXiv preprint arXiv:2212.06817}, 2022.

\bibitem[Cheng et~al.(2022)Cheng, Yang, Xiang, and Liu]{cheng2022financial}
Dawei Cheng, Fangzhou Yang, Sheng Xiang, and Jin Liu.
\newblock Financial time series forecasting with multi-modality graph neural network.
\newblock \emph{Pattern Recognition}, 121:\penalty0 108218, 2022.

\bibitem[Das et~al.(2023{\natexlab{a}})Das, Kong, Leach, Sen, and Yu]{das2023long}
Abhimanyu Das, Weihao Kong, Andrew Leach, Rajat Sen, and Rose Yu.
\newblock Long-term forecasting with tide: Time-series dense encoder.
\newblock \emph{arXiv preprint arXiv:2304.08424}, 2023{\natexlab{a}}.

\bibitem[Das et~al.(2023{\natexlab{b}})Das, Kong, Sen, and Zhou]{das2023decoder}
Abhimanyu Das, Weihao Kong, Rajat Sen, and Yichen Zhou.
\newblock A decoder-only foundation model for time-series forecasting.
\newblock \emph{arXiv preprint arXiv:2310.10688}, 2023{\natexlab{b}}.

\bibitem[Das et~al.(2024)Das, Kong, Sen, and Zhou]{das2024decoder}
Abhimanyu Das, Weihao Kong, Rajat Sen, and Yichen Zhou.
\newblock A decoder-only foundation model for time-series forecasting.
\newblock In \emph{Forty-first International Conference on Machine Learning}, 2024.

\bibitem[Dettmers et~al.(2023)Dettmers, Pagnoni, Holtzman, and Zettlemoyer]{NEURIPS2023_1feb8787}
Tim Dettmers, Artidoro Pagnoni, Ari Holtzman, and Luke Zettlemoyer.
\newblock Qlora: Efficient finetuning of quantized llms.
\newblock In A.~Oh, T.~Naumann, A.~Globerson, K.~Saenko, M.~Hardt, and S.~Levine (eds.), \emph{Advances in Neural Information Processing Systems}, volume~36, pp.\  10088--10115. Curran Associates, Inc., 2023.
\newblock URL \url{https://proceedings.neurips.cc/paper_files/paper/2023/file/1feb87871436031bdc0f2beaa62a049b-Paper-Conference.pdf}.

\bibitem[Dhamala et~al.(2018)Dhamala, Liang, Bressler, and Ding]{dhamala2018granger}
Mukesh Dhamala, Hualou Liang, Steven~L Bressler, and Mingzhou Ding.
\newblock Granger-geweke causality: Estimation and interpretation.
\newblock \emph{NeuroImage}, 175:\penalty0 460--463, 2018.

\bibitem[Ekambaram et~al.(2024)Ekambaram, Jati, Dayama, Mukherjee, Nguyen, Gifford, Reddy, and Kalagnanam]{ekambaram2024tiny}
Vijay Ekambaram, Arindam Jati, Pankaj Dayama, Sumanta Mukherjee, Nam Nguyen, Wesley~M Gifford, Chandra Reddy, and Jayant Kalagnanam.
\newblock Tiny time mixers (ttms): Fast pre-trained models for enhanced zero/few-shot forecasting of multivariate time series.
\newblock \emph{Advances in Neural Information Processing Systems}, 37:\penalty0 74147--74181, 2024.

\bibitem[Goyal et~al.(2017)Goyal, Doll{\'a}r, Girshick, Noordhuis, Wesolowski, Kyrola, Tulloch, Jia, and He]{goyal2017accurate}
Priya Goyal, Piotr Doll{\'a}r, Ross Girshick, Pieter Noordhuis, Lukasz Wesolowski, Aapo Kyrola, Andrew Tulloch, Yangqing Jia, and Kaiming He.
\newblock Accurate, large minibatch sgd: Training imagenet in 1 hour.
\newblock \emph{arXiv preprint arXiv:1706.02677}, 2017.

\bibitem[Graf et~al.(2025)Graf, Ortner, Wo{\'L}{\c{s}}niak, Pantazi, et~al.]{graf2025flowstate}
Lars Graf, Thomas Ortner, Stanis{\'L} Wo{\'L}{\c{s}}niak, Angeliki Pantazi, et~al.
\newblock Flowstate: Sampling rate invariant time series forecasting.
\newblock \emph{arXiv preprint arXiv:2508.05287}, 2025.

\bibitem[Granger(1969)]{granger1969investigating}
Clive~WJ Granger.
\newblock Investigating causal relations by econometric models and cross-spectral methods.
\newblock \emph{Econometrica: journal of the Econometric Society}, pp.\  424--438, 1969.

\bibitem[Han et~al.(2025)Han, Liu, Deng, Jiang, Sun, Yu, Wang, Lu, Ma, Ye, et~al.]{han2025unica}
Lu~Han, Yu~Liu, Qiwen Deng, Jian Jiang, Yinbo Sun, Zhe Yu, Binfeng Wang, Xingyu Lu, Lintao Ma, Han-Jia Ye, et~al.
\newblock Unica: Adapting time series foundation model to general covariate-aware forecasting.
\newblock \emph{arXiv preprint arXiv:2506.22039}, 2025.

\bibitem[Hittawe et~al.(2024)Hittawe, Harrou, Togou, Sun, and Knio]{hittawe2024time}
Mohamad~Mazen Hittawe, Fouzi Harrou, Mohammed~Amine Togou, Ying Sun, and Omar Knio.
\newblock Time-series weather prediction in the red sea using ensemble transformers.
\newblock \emph{Applied Soft Computing}, 164:\penalty0 111926, 2024.

\bibitem[Hoo et~al.(2025)Hoo, M{\"u}ller, Salinas, and Hutter]{hoo2025tables}
Shi~Bin Hoo, Samuel M{\"u}ller, David Salinas, and Frank Hutter.
\newblock From tables to time: How tabpfn-v2 outperforms specialized time series forecasting models.
\newblock \emph{arXiv preprint arXiv:2501.02945}, 2025.

\bibitem[Hu et~al.(2021)Hu, Shen, Wallis, Allen-Zhu, Li, Wang, Wang, and Chen]{hu2021lora}
Edward~J Hu, Yelong Shen, Phillip Wallis, Zeyuan Allen-Zhu, Yuanzhi Li, Shean Wang, Lu~Wang, and Weizhu Chen.
\newblock Lora: Low-rank adaptation of large language models.
\newblock \emph{arXiv preprint arXiv:2106.09685}, 2021.

\bibitem[Jin et~al.(2023)Jin, Wang, Ma, Chu, Zhang, Shi, Chen, Liang, Li, Pan, et~al.]{jin2023time}
Ming Jin, Shiyu Wang, Lintao Ma, Zhixuan Chu, James~Y Zhang, Xiaoming Shi, Pin-Yu Chen, Yuxuan Liang, Yuan-Fang Li, Shirui Pan, et~al.
\newblock Time-llm: Time series forecasting by reprogramming large language models.
\newblock \emph{arXiv preprint arXiv:2310.01728}, 2023.

\bibitem[Lago et~al.(2021)Lago, Marcjasz, De~Schutter, and Weron]{lago2021forecasting}
Jesus Lago, Grzegorz Marcjasz, Bart De~Schutter, and Rafa{\l} Weron.
\newblock Forecasting day-ahead electricity prices: A review of state-of-the-art algorithms, best practices and an open-access benchmark.
\newblock \emph{Applied Energy}, 293:\penalty0 116983, 2021.

\bibitem[Lim et~al.(2021)Lim, Ar{\i}k, Loeff, and Pfister]{lim2021temporal}
Bryan Lim, Sercan~{\"O} Ar{\i}k, Nicolas Loeff, and Tomas Pfister.
\newblock Temporal fusion transformers for interpretable multi-horizon time series forecasting.
\newblock \emph{International Journal of Forecasting}, 37\penalty0 (4):\penalty0 1748--1764, 2021.

\bibitem[Liu et~al.(2024{\natexlab{a}})Liu, Xu, Zhao, Kong, Prabhakar~Kamarthi, Sasanur, Sharma, Cui, Wen, Zhang, et~al.]{liu2024time}
Haoxin Liu, Shangqing Xu, Zhiyuan Zhao, Lingkai Kong, Harshavardhan Prabhakar~Kamarthi, Aditya Sasanur, Megha Sharma, Jiaming Cui, Qingsong Wen, Chao Zhang, et~al.
\newblock Time-mmd: Multi-domain multimodal dataset for time series analysis.
\newblock \emph{Advances in Neural Information Processing Systems}, 37:\penalty0 77888--77933, 2024{\natexlab{a}}.

\bibitem[Liu et~al.(2022)Liu, Zeng, Chen, Xu, Lai, Ma, and Xu]{liu2022scinet}
Minhao Liu, Ailing Zeng, Muxi Chen, Zhijian Xu, Qiuxia Lai, Lingna Ma, and Qiang Xu.
\newblock Scinet: Time series modeling and forecasting with sample convolution and interaction.
\newblock \emph{Advances in Neural Information Processing Systems}, 35:\penalty0 5816--5828, 2022.

\bibitem[Liu et~al.(2024{\natexlab{b}})Liu, Liu, Woo, Aksu, Liang, Zimmermann, Liu, Savarese, Xiong, and Sahoo]{liu2024moirai}
Xu~Liu, Juncheng Liu, Gerald Woo, Taha Aksu, Yuxuan Liang, Roger Zimmermann, Chenghao Liu, Silvio Savarese, Caiming Xiong, and Doyen Sahoo.
\newblock Moirai-moe: Empowering time series foundation models with sparse mixture of experts.
\newblock \emph{arXiv preprint arXiv:2410.10469}, 2024{\natexlab{b}}.

\bibitem[Liu et~al.(2023)Liu, Hu, Zhang, Wu, Wang, Ma, and Long]{liu2023itransformer}
Yong Liu, Tengge Hu, Haoran Zhang, Haixu Wu, Shiyu Wang, Lintao Ma, and Mingsheng Long.
\newblock itransformer: Inverted transformers are effective for time series forecasting.
\newblock \emph{arXiv preprint arXiv:2310.06625}, 2023.

\bibitem[Liu et~al.(2024{\natexlab{c}})Liu, Qin, Huang, Wang, and Long]{liu2024timer}
Yong Liu, Guo Qin, Xiangdong Huang, Jianmin Wang, and Mingsheng Long.
\newblock Timer-xl: Long-context transformers for unified time series forecasting.
\newblock \emph{arXiv preprint arXiv:2410.04803}, 2024{\natexlab{c}}.

\bibitem[Liu et~al.(2025)Liu, Qin, Shi, Chen, Yang, Huang, Wang, and Long]{liu2025sundial}
Yong Liu, Guo Qin, Zhiyuan Shi, Zhi Chen, Caiyin Yang, Xiangdong Huang, Jianmin Wang, and Mingsheng Long.
\newblock Sundial: A family of highly capable time series foundation models.
\newblock \emph{arXiv preprint arXiv:2502.00816}, 2025.

\bibitem[Nie et~al.(2022)Nie, Nguyen, Sinthong, and Kalagnanam]{nie2022time}
Yuqi Nie, Nam~H Nguyen, Phanwadee Sinthong, and Jayant Kalagnanam.
\newblock A time series is worth 64 words: Long-term forecasting with transformers.
\newblock \emph{arXiv preprint arXiv:2211.14730}, 2022.

\bibitem[Olivares et~al.(2023)Olivares, Challu, Marcjasz, Weron, and Dubrawski]{olivares2023neural}
Kin~G Olivares, Cristian Challu, Grzegorz Marcjasz, Rafa{\l} Weron, and Artur Dubrawski.
\newblock Neural basis expansion analysis with exogenous variables: Forecasting electricity prices with nbeatsx.
\newblock \emph{International Journal of Forecasting}, 39\penalty0 (2):\penalty0 884--900, 2023.

\bibitem[Oreshkin et~al.(2019)Oreshkin, Carpov, Chapados, and Bengio]{oreshkin2019n}
Boris~N Oreshkin, Dmitri Carpov, Nicolas Chapados, and Yoshua Bengio.
\newblock N-beats: Neural basis expansion analysis for interpretable time series forecasting.
\newblock \emph{arXiv preprint arXiv:1905.10437}, 2019.

\bibitem[Panda \& Mohanty(2023)Panda and Mohanty]{panda2023time}
Sandeep~Kumar Panda and Sachi~Nandan Mohanty.
\newblock Time series forecasting and modeling of food demand supply chain based on regressors analysis.
\newblock \emph{Ieee Access}, 11:\penalty0 42679--42700, 2023.

\bibitem[Pearson(1895)]{pearson1895vii}
Karl Pearson.
\newblock Vii. note on regression and inheritance in the case of two parents.
\newblock \emph{proceedings of the royal society of London}, 58\penalty0 (347-352):\penalty0 240--242, 1895.

\bibitem[Peebles \& Xie(2023)Peebles and Xie]{peebles2023scalable}
William Peebles and Saining Xie.
\newblock Scalable diffusion models with transformers.
\newblock In \emph{Proceedings of the IEEE/CVF international conference on computer vision}, pp.\  4195--4205, 2023.

\bibitem[Qiu et~al.(2025)Qiu, Cheng, Wu, Hu, Guo, and Yang]{qiu2025comprehensive}
Xiangfei Qiu, Hanyin Cheng, Xingjian Wu, Jilin Hu, Chenjuan Guo, and Bin Yang.
\newblock A comprehensive survey of deep learning for multivariate time series forecasting: A channel strategy perspective.
\newblock \emph{arXiv preprint arXiv:2502.10721}, 2025.

\bibitem[Salinas et~al.(2020)Salinas, Flunkert, Gasthaus, and Januschowski]{salinas2020deepar}
David Salinas, Valentin Flunkert, Jan Gasthaus, and Tim Januschowski.
\newblock Deepar: Probabilistic forecasting with autoregressive recurrent networks.
\newblock \emph{International journal of forecasting}, 36\penalty0 (3):\penalty0 1181--1191, 2020.

\bibitem[Shi et~al.(2024)Shi, Wang, Nie, Li, Ye, Wen, and Jin]{shi2024time}
Xiaoming Shi, Shiyu Wang, Yuqi Nie, Dianqi Li, Zhou Ye, Qingsong Wen, and Ming Jin.
\newblock Time-moe: Billion-scale time series foundation models with mixture of experts.
\newblock \emph{arXiv preprint arXiv:2409.16040}, 2024.

\bibitem[Vagropoulos et~al.(2016)Vagropoulos, Chouliaras, Kardakos, Simoglou, and Bakirtzis]{vagropoulos2016comparison}
Stylianos~I Vagropoulos, GI~Chouliaras, Evaggelos~G Kardakos, Christos~K Simoglou, and Anastasios~G Bakirtzis.
\newblock Comparison of sarimax, sarima, modified sarima and ann-based models for short-term pv generation forecasting.
\newblock In \emph{2016 IEEE international energy conference (ENERGYCON)}, pp.\  1--6. IEEE, 2016.

\bibitem[Vaswani et~al.(2017)Vaswani, Shazeer, Parmar, Uszkoreit, Jones, Gomez, Kaiser, and Polosukhin]{vaswani2017attention}
Ashish Vaswani, Noam Shazeer, Niki Parmar, Jakob Uszkoreit, Llion Jones, Aidan~N Gomez, {\L}ukasz Kaiser, and Illia Polosukhin.
\newblock Attention is all you need.
\newblock \emph{Advances in neural information processing systems}, 30, 2017.

\bibitem[Wang et~al.(2024)Wang, Wu, Dong, Liu, Qiu, Zhang, Wang, and Long]{wang2024timexer}
Yuxuan Wang, Haixu Wu, Jiaxiang Dong, Yong Liu, Yunzhong Qiu, Haoran Zhang, Jianmin Wang, and Mingsheng Long.
\newblock Timexer: Empowering transformers for time series forecasting with exogenous variables.
\newblock \emph{arXiv preprint arXiv:2402.19072}, 2024.

\bibitem[Williams(2001)]{williams2001multivariate}
Billy~M Williams.
\newblock Multivariate vehicular traffic flow prediction: Evaluation of arimax modeling.
\newblock \emph{Transportation Research Record}, 1776\penalty0 (1):\penalty0 194--200, 2001.

\bibitem[Woo et~al.(2023)Woo, Liu, Kumar, and Sahoo]{woo2023pushing}
Gerald Woo, Chenghao Liu, Akshat Kumar, and Doyen Sahoo.
\newblock Pushing the limits of pre-training for time series forecasting in the cloudops domain.
\newblock \emph{arXiv preprint arXiv:2310.05063}, 2023.

\bibitem[Woo et~al.(2024)Woo, Liu, Kumar, Xiong, Savarese, and Sahoo]{woo2024unified}
Gerald Woo, Chenghao Liu, Akshat Kumar, Caiming Xiong, Silvio Savarese, and Doyen Sahoo.
\newblock Unified training of universal time series forecasting transformers.
\newblock \emph{arXiv preprint arXiv:2402.02592}, 2024.

\bibitem[Wu et~al.(2020)Wu, Xu, Dai, Wan, Zhang, Yan, Tomizuka, Gonzalez, Keutzer, and Vajda]{wu2020visual}
Bichen Wu, Chenfeng Xu, Xiaoliang Dai, Alvin Wan, Peizhao Zhang, Zhicheng Yan, Masayoshi Tomizuka, Joseph Gonzalez, Kurt Keutzer, and Peter Vajda.
\newblock Visual transformers: Token-based image representation and processing for computer vision, 2020.

\bibitem[Wu et~al.(2021)Wu, Xu, Wang, and Long]{wu2021autoformer}
Haixu Wu, Jiehui Xu, Jianmin Wang, and Mingsheng Long.
\newblock Autoformer: Decomposition transformers with auto-correlation for long-term series forecasting.
\newblock \emph{Advances in Neural Information Processing Systems}, 34:\penalty0 22419--22430, 2021.

\bibitem[Zeng et~al.(2023)Zeng, Chen, Zhang, and Xu]{zeng2023transformers}
Ailing Zeng, Muxi Chen, Lei Zhang, and Qiang Xu.
\newblock Are transformers effective for time series forecasting?
\newblock In \emph{Proceedings of the AAAI conference on artificial intelligence}, volume~37, pp.\  11121--11128, 2023.

\bibitem[Zhang et~al.(2025)Zhang, Li, Long, Zhang, Lin, Yang, Xie, Yang, Liu, Lin, Huang, and Zhou]{qwen3embedding}
Yanzhao Zhang, Mingxin Li, Dingkun Long, Xin Zhang, Huan Lin, Baosong Yang, Pengjun Xie, An~Yang, Dayiheng Liu, Junyang Lin, Fei Huang, and Jingren Zhou.
\newblock Qwen3 embedding: Advancing text embedding and reranking through foundation models.
\newblock \emph{arXiv preprint arXiv:2506.05176}, 2025.

\bibitem[Zhang \& Yan(2022)Zhang and Yan]{zhang2022crossformer}
Yunhao Zhang and Junchi Yan.
\newblock Crossformer: Transformer utilizing cross-dimension dependency for multivariate time series forecasting.
\newblock In \emph{The Eleventh International Conference on Learning Representations}, 2022.

\bibitem[Zhang \& Yan(2023)Zhang and Yan]{zhang2023crossformer}
Yunhao Zhang and Junchi Yan.
\newblock Crossformer: Transformer utilizing cross-dimension dependency for multivariate time series forecasting.
\newblock In \emph{The eleventh international conference on learning representations}, 2023.

\bibitem[Zhong et~al.(2025)Zhong, Ruan, Jin, Li, Wen, and Liang]{zhong2025time}
Siru Zhong, Weilin Ruan, Ming Jin, Huan Li, Qingsong Wen, and Yuxuan Liang.
\newblock Time-vlm: Exploring multimodal vision-language models for augmented time series forecasting.
\newblock \emph{arXiv preprint arXiv:2502.04395}, 2025.

\bibitem[Zhou et~al.(2021)Zhou, Zhang, Peng, Zhang, Li, Xiong, and Zhang]{zhou2021informer}
Haoyi Zhou, Shanghang Zhang, Jieqi Peng, Shuai Zhang, Jianxin Li, Hui Xiong, and Wancai Zhang.
\newblock Informer: Beyond efficient transformer for long sequence time-series forecasting.
\newblock In \emph{Proceedings of the AAAI conference on artificial intelligence}, volume~35, pp.\  11106--11115, 2021.

\end{thebibliography}
\bibliographystyle{iclr2026_conference}

\clearpage
\appendix
\section{Experimental Details}
\subsection{Datasets}
To comprehensively evaluate the performance of CoRA, we conduct extensive experiments on several well-established benchmarks. The evaluation covers uni-modal, multi-modal covariate-aware forecasting and multivariate forecasting tasks. The datasets we used are described below:

For uni-modal long-term covariate-aware forecasting task, we include the following benchmark datasets: ETT (Electricity Transforming Temperature)~\citep{zhou2021informer} contains seven power transformer load factors from July 2016 to July 2018. According to sampling frequency and location, the dataset is partitioned into four subsets: ETTh1 and ETTh2 contain hourly measurements, whereas ETTm1 and ETTm2 provide observations at 15-minute intervals. Weather~\citep{wu2021autoformer} comprises 21 meteorological variates collected at 10-minute intervals throughout 2020 from the Max Planck Institute for Biogeochemistry. ECL (Electricity Consuming Load)~\citep{wu2021autoformer} records hourly electricity consumption of 321 residential and commercial clients, offering diverse patterns of consumption behavior. Traffic~\citep{wu2021autoformer} consists of hourly road occupancy data from 862 sensors installed on highways in the San Francisco Bay Area, covering the period from January 2015 to December 2016. Further statistics are reported in Table~\ref{tab:dataset_descriptions}.

For uni-modal short-term covariate-aware forecasting task, we include the following benchmark datasets: EPF (Electricity Price Forecasting)~\citep{lago2021forecasting} contains 6 years of hourly day-ahead electricity prices, complemented by two exogenous forecast series (load and renewable generation). The dataset spans five major European electricity markets, facilitating robust cross-market performance analysis under diverse price dynamics and market conditions. (1) NP (Nord Pool) covers the Nord Pool electricity market, containing hourly electricity prices together with grid load and wind power forecasts from 2013-01-01 to 2018-12-24. (2) PJM corresponds to the Pennsylvania–New Jersey–Maryland market, including the zonal electricity price in the Commonwealth Edison (COMED) area, system load, and COMED load forecasts from 2013-01-01 to 2018-12-24. (3) BE denotes Belgium's electricity market, recording hourly electricity prices, load forecasts in Belgium, and generation forecasts in France from 2011-01-09 to 2016-12-31. (4) FR corresponds to the French electricity market, containing hourly prices with associated load and generation forecasts from 2012-01-09 to 2017-12-31. (5) DE represents the German electricity market, providing hourly prices, zonal load forecasts in the TSO Amprion zone, and wind and solar generation forecasts from 2012-01-09 to 2017-12-31. Further statistics are reported in Table~\ref{tab:dataset_descriptions}.

To assess CoRA's capability in multi-modal covariate-aware forecasting, we employ RT-1~\citep{brohan2022rt}, a large-scale robotic dataset with about 130k demonstrations collected over 17 months using 13 robots in office kitchen environments. It covers 744 skills, ranging from basic object manipulation to long-horizon instructions, each paired with natural language commands and visual observations. The dataset provides rich multi-modal supervision, supporting studies on instruction-conditioned and multi-modal forecasting. The RT-1 dataset is particularly valuable for studying multi-modal and instruction-conditioned forecasting, as it provides paired visual observations and natural language descriptions aligned with robotic trajectories. In our experiments, we use a subset of RT-1, specifically the 'Move Object Near Object' skill, and further restrict it to series with lengths no shorter than 45. Each sequence is partitioned into training, validation, and test sets by assigning the last four points as test targets and the preceding four points as validation targets, with the remaining points used for training. This protocol guarantees at least one validation and one test instance per series, under a setup with an input length of 32 and a prediction horizon of 4. Time-MMD~\citep{liu2024time} is a large-scale multi-modal dataset encompassing nine diverse domains, including agriculture, climate, healthcare, and transportation. Each time series is paired with corresponding textual information sourced from curated domain reports and structured web search results, enabling evaluation of text-enhanced forecasting performance. For consistency with prior work~\citep{han2025unica}, we exclude the Agriculture and Economy subsets, and keep all other experimental settings identical to the official configuration. Details of these datasets are provided in Table~\ref{tab:dataset_statistics}.

\clearpage
\begin{table}[ht]
  \caption{Detailed dataset descriptions. \textit{Nums} denotes the number of covariates. \textit{Freq} denotes the sampling interval of time points. The dataset size is given as (Train, Validation, Test).}
  \label{tab:dataset_descriptions}
  
  \renewcommand{\arraystretch}{1.5}
  \setlength{\tabcolsep}{4pt}
  
  \resizebox{\textwidth}{!}{  
  \footnotesize
  \begin{tabular}{@{}l l c l
      >{\raggedright\arraybackslash}p{3cm}
      >{\raggedright\arraybackslash}p{3cm}
      c c@{}}
    \toprule
    Dataset & Domain & Nums & Freq & Target Variate & Covariate & Dataset Size & Prediction Horizon \\
    \toprule
    Electricity & Energy & 320 & 1H & Electricity Consumption & Electricity Consumption & (18317, 2633, 5261) & (96, 192, 336, 720) \\
    Weather & Weather & 20 & 10M & CO\textsubscript{2} Concentration & Climate Feature & (36792, 5271, 10540) & (96, 192, 336, 720) \\
    ETTh & Energy & 6 & 1H & Oil Temperature & Power Load Feature & (8545, 2881, 2881) & (96, 192, 336, 720) \\
    ETTm & Energy & 6 & 15M & Oil Temperature & Power Load Feature & (34465, 11521, 11521) & (96, 192, 336, 720) \\
    Traffic & Traffic & 861 & 1H & Road Occupancy Rates & Road Occupancy Rates & (12185, 1757, 3509) & (96, 192, 336, 720) \\
    \midrule
    NP & Electricity & 2 & 1H & Nord Pool Electricity Price & Grid Load, Wind Power & (36500, 5219, 10460) & 24\\
    PJM & Electricity & 2 & 1H & PJM Electricity Price & System Load, Zonal COMED Load & (36500, 5219, 10460) & 24\\
    BE & Electricity & 2 & 1H & Belgium Electricity Price & Generation, System Load & (36500, 5219, 10460) & 24\\
    FR & Electricity & 2 & 1H & France Electricity Price & Generation, System Load & (36500, 5219, 10460) & 24 \\
    DE & Electricity & 2 & 1H & German Electricity Price & Wind Power, Amprion Zonal Load & (36500, 5219, 10460) & 24 \\
    \bottomrule
  \end{tabular}
  }
\end{table}

\begin{table}[ht]
  \centering
  \caption{Detailed descriptions of RT-1~\citep{brohan2022rt} and TimeMMD~\citep{liu2024time}.}
  \label{tab:dataset_statistics}
  
  \renewcommand{\arraystretch}{2}
  \setlength{\tabcolsep}{4pt}
  
  \resizebox{\textwidth}{!}{
  \begin{tabular}{@{}llcccccc@{}}
    \toprule
    Dataset & Domain & Num. Obs. & Num. Series & Freq & Target Variate & Covariate Type & Prediction Horizon \\
    \toprule
    RT-1 & Solar Power & 33,420 & 2871 & $\frac{1}{3}$S & height to bottom & Image & 4 \\
    \midrule
    \multirow{9}{*}{TimeMMD} 
      & Agriculture   & 486 & 1 & 1M & Retail Broiler Composite    & Text & 12 \\
      & Climate       & 496 & 1 & 1M & Drought Level               & Text & 12 \\
      & Economy       & 423 & 1 & 1M & International Trade Balance & Text & 12 \\
      & Energy        & 1479 & 1 & 1M & Gasoline Prices             & Text & 12 \\
      & Environment   & 11102 & 1 & 1M & Air Quality Index           & Text & 12 \\
      & Health        & 1389 & 1 & 1W & Influenza Patients Proportion & Text & 12 \\
      & Security      & 297 & 1 & 1D & Disaster and Emergency Grants & Text & 12 \\
      & Social Good   & 900 & 1 & 1M & Unemployment Rate           & Text & 12 \\
      & Traffic       & 531 & 1 & 1M & Travel Volume               & Text & 12 \\
    \bottomrule
  \end{tabular}
  }
\end{table}

\subsection{Baseline Models}
We compared our method to multiple advanced baselines across various forecasting tasks.
\paragraph{Time Series Foundation Models}
We evaluate CoRA across multiple Time Series Foundation Models, including Sundial~\citep{liu2025sundial}, TimesFM~\citep{das2023decoder}, Chronos-Bolt~\citep{ansari2024chronos}, and FlowState~\citep{graf2025flowstate}. Specifically, on the Time-MMD dataset~\citep{liu2024time}, we further include Moirai~\citep{liu2024moirai} and TabPFN-TS~\citep{hoo2025tables} as baselines.

\paragraph{Covariate-Aware Deep models}
We compare CoRA with diverse advanced supervised deep forecasters. 
These include Transformer-based architectures such as TimeXer~\citep{wang2024timexer}, iTransformer~\citep{liu2023itransformer}, PatchTST~\citep{nie2022time}, Crossformer~\citep{zhang2022crossformer}, Autoformer~\citep{wu2021autoformer}, TiDE~\citep{das2023long}, Time-LLM~\citep{jin2023time}, TTM~\citep{ekambaram2024tiny} and TFT~\citep{lim2021temporal}; classical sequence models such as N-BEATS~\citep{oreshkin2019n}, NBEATSx~\citep{olivares2023neural} and DeepAR~\citep{salinas2020deepar};
and other strong baselines including DLinear~\citep{zeng2023transformers} and SCINet~\citep{liu2022scinet}.

\paragraph{Adaptation Method}
We evaluate CoRA against other covariate adaptation methods, including UniCA~\citep{han2025unica}, ChronosX~\citep{arango2025chronosx}, and AdaPTS~\citep{benechehab2025adapts}. 
In addition, to assess the role of covariates explicitly, we also compare with Supervised Fine-Tuning (SFT), which adapts model parameters without leveraging covariate signals.

\subsection{Implementation Details}

All experiments are conducted using PyTorch on NVIDIA A100 Tensor Core GPUs. We employ the Adam optimizer, along with the respective loss function of each foundation model, for optimization; unless otherwise specified, the default loss function is mean squared error (MSE).

The training process is limited to a maximum of 50 epochs with early stopping, and patience is set to 3. The learning rate is selected from the set \{5e-6, 1e-5, 2e-5\}, and the batch size is fixed at 128.

For EPF, we follow the benchmark results reported in ~\citep{wang2024timexer}. For Time-MMD~\citep{liu2024time}, we use the results reported in ~\citep{han2025unica}, both of which are strictly based on the configurations in original papers. For all other results, we reproduce both the adaptation methods and the deep forecasting models from their official repositories, keeping hyperparameters and training configurations unchanged to ensure a fair evaluation of each base model.

In addition, we provide an algorithmic description to illustrate the core component of our framework. 
Algorithm~\ref{alg:CoRA} presents the workflow of the proposed CoRA method, where multi-modal covariates are encoded through 
their respective foundation model backbones, aligned and reweighted by a Granger Causality embedding, and then integrated with 
the target series representations for final prediction. 
For completeness, Algorithm~\ref{alg:Granger} further outlines the procedure for estimating Granger-Geweke Causality~\citep{dhamala2018granger} between 
covariates and the target variate, which serves as the theoretical grounding for our covariate selection and weighting mechanism.

\begin{algorithm}[htbp]
\caption{CoRA Algorithm} 
\label{alg:CoRA}
\begin{algorithmic}[1]
    \Require 
        Past target series $\mathbf{x}_{1:T}=\{x_1,\ldots,x_T\}$; 
        Covariates $\mathbf{C}_{1:\tau} = \{\mathbf{C}_1,\ldots,\mathbf{C}_{\tau}\}$ (time series, text, image); 
        Prediction horizon $H$
    \State $\mathbf{E}_{1:\tau_i}^{m_i}=\operatorname{FM-Backbone}(\mathbf{C}^{m_i}_{1:\tau_i}),\ i=1,\dots,N,\ m_i \in \{\mathrm{ts}, \mathrm{txt}, \mathrm{img}\}$
    \State $\tilde{\mathbf{E}}^{\mathrm{ts}} = \mathbf{E}^{\mathrm{ts}}_{\tau}$
    \State $\tilde{\mathbf{E}}^{\mathrm{txt}} = \frac{1}{\tau} \sum_{t=1}^{\tau} \mathbf{E}^{\mathrm{txt}}_t,\ \tilde{\mathbf{E}}^{\mathrm{img}} = \frac{1}{\tau} \sum_{t=1}^{\tau} \mathbf{E}^{\mathrm{img}}_t$
    \State $\mathbf{E}^{\mathrm{target}}_{1:T} = \operatorname{TSFM-Backbone} (\mathbf{x}_{1:T})$
    \State $\tilde{\mathbf{E}}^{\mathrm{target}} = \mathbf{E}^{\mathrm{target}}_{T}$
    \State $\hat{\mathbf{E}}^{m_i} = \tilde{\mathbf{E}}^{m_i}\, \mathbf{W}^{m_i} + \mathbf{b}^{m_i},\ i=1,\dots,N,\ 
m_i \in \{\mathrm{ts}, \mathrm{txt}, \mathrm{img}\}$
    \State $\hat{\mathbf{E}} = \operatorname{Concat}\Big(\hat{\mathbf{E}}^{\mathrm{ts}}, \hat{\mathbf{E}}^{\mathrm{txt}}, \hat{\mathbf{E}}^{\mathrm{img}}\Big)$
    \State $\mathbf{H} = \operatorname{Softmax}(\mathbf{W}_{\text{GC}})\cdot\hat{\mathbf{E}}$
    \State $\mathbf{\gamma}, \mathbf{\beta}, \mathbf{\alpha} = \operatorname{MLP}\big(\mathbf{H}\big)$
    \State $\hat{\mathbf{x}}_{T+1:T+H} = (1 + \mathbf{\alpha})\operatorname{TSFM-Head}\big(\mathbf{\gamma} + (1 + \mathbf{\beta})\ \tilde{\mathbf{E}}^{\mathrm{target}}\big)$
    \State \Return $\hat{\mathbf{x}}_{T+1:T+H}$
\end{algorithmic}
\end{algorithm}

\begin{algorithm}[htbp]
\caption{Granger Causality Algorithm}
\label{alg:Granger}
\begin{algorithmic}[1]
    \Require covariate series $A$, target series $B$, maximum lag $L_{\max}$, criterion
    \Ensure Granger causality strength $GC$, selected lag $l$
    \State Select lag $l$ by minimizing criterion over $1,\dots,L_{\max}$
    \State Fit restricted model on $B_t$ \Comment{use $\{B_{t-1},\dots,B_{t-l}\}$, residual variance $\sigma_r^2$}
    \State Fit unrestricted model on $B_t$ \Comment{use $\{B_{t-1},\dots,B_{t-l},A_{t-1},\dots,A_{t-l}\}$, residual variance $\sigma_u^2$}
    \State Compute Granger causality strength: $GC \leftarrow \log \tfrac{\sigma_r^2}{\sigma_u^2}$
    \State \Return GC
\end{algorithmic}
\end{algorithm}

\section{Full Results}

\subsection{Full Results of Uni-Modal Covariate-Aware Forecasting}
Table~\ref{tab:tslib_exo_full_result} reports the complete results of the uni-modal covariate-aware forecasting task across widely used datasets. All adaptation methods built on Sundial are fine-tuned only for the output horizon of 720, consistent with the available pre-trained Sundial weights. For shorter horizons, the outputs are obtained by truncating the 720-length predictions. In contrast, the baseline deep models are individually trained for each prediction length. Overall, adaptation methods on top of TSFMs consistently outperform conventional deep models, and our proposed CoRA achieves state-of-the-art results, demonstrating its effectiveness as a general approach for covariate-aware adaptation.

\begin{table}[htbp]
  \caption{Full results of the long-term covariate-aware forecasting task. For all baselines, the look-back length $L$ is fixed at 2880 and dash (-) denotes out of memory (OOM) problem.}
  \vspace{-3pt}
  \renewcommand{\arraystretch}{0.85} 
  \centering
  \resizebox{1\columnwidth}{!}{
  \begin{threeparttable}
  \begin{small}
  \renewcommand{\multirowsetup}{\centering}
  \setlength{\tabcolsep}{1.45pt}
  \label{tab:tslib_exo_full_result}
  \begin{tabular}{c|c|cc|cc|cc|cc|cc|cc|cc|cc|cc|cc}
    \toprule
    \multicolumn{2}{c|}{\multirow{2}{*}{Models}} & 
    \multicolumn{2}{c}{\rotatebox{0}{\scalebox{0.75}{\textbf{CoRA}}}} &
    \multicolumn{2}{c}{\rotatebox{0}{\scalebox{0.8}{AdaPTS}}} &
    \multicolumn{2}{c}{\rotatebox{0}{\scalebox{0.8}{ChronosX}}} &
    \multicolumn{2}{c}{\rotatebox{0}{\scalebox{0.8}{UniCA}}} &
    \multicolumn{2}{c}{\rotatebox{0}{\scalebox{0.8}{TimeXer}}} &
    \multicolumn{2}{c}{\rotatebox{0}{\scalebox{0.8}{iTransformer}}} &
    \multicolumn{2}{c}{\rotatebox{0}{\scalebox{0.8}{{PatchTST}}}} &
    \multicolumn{2}{c}{\rotatebox{0}{\scalebox{0.8}{NBEATSx}}} &
    \multicolumn{2}{c}{\rotatebox{0}{\scalebox{0.8}{Crossformer}}} &
    \multicolumn{2}{c}{\rotatebox{0}{\scalebox{0.8}{DLinear}}} \\
    \multicolumn{2}{c|}{} &
    \multicolumn{2}{c}{\scalebox{0.8}{\textbf{(Ours)}}} &
    \multicolumn{2}{c}{\scalebox{0.8}{\citeyearpar{benechehab2025adapts}}} & 
    \multicolumn{2}{c}{\scalebox{0.8}{\citeyearpar{arango2025chronosx}}} & 
    \multicolumn{2}{c}{\scalebox{0.8}{\citeyearpar{han2025unica}}} & 
    \multicolumn{2}{c}{\scalebox{0.8}{\citeyearpar{wang2024timexer}}} & 
    \multicolumn{2}{c}{\scalebox{0.8}{\citeyearpar{liu2023itransformer}}} & 
    \multicolumn{2}{c}{\scalebox{0.8}{\citeyearpar{nie2022time}}} & 
    \multicolumn{2}{c}{\scalebox{0.8}{\citeyearpar{olivares2023neural}}} & 
    \multicolumn{2}{c}{\scalebox{0.8}{\citeyearpar{zhang2023crossformer}}} & 
    \multicolumn{2}{c}{\scalebox{0.8}{\citeyearpar{zeng2023transformers}}}  \\
    \cmidrule(lr){3-4}\cmidrule(lr){5-6} \cmidrule(lr){7-8}\cmidrule(lr){9-10}\cmidrule(lr){11-12} \cmidrule(lr){13-14} \cmidrule(lr){15-16} \cmidrule(lr){17-18} \cmidrule(lr){19-20} \cmidrule(lr){21-22} 
    \multicolumn{2}{c|}{Metric}
    & \scalebox{0.8}{MSE} & \scalebox{0.8}{MAE}  
    & \scalebox{0.8}{MSE} & \scalebox{0.8}{MAE}  
    & \scalebox{0.8}{MSE} & \scalebox{0.8}{MAE}  
    & \scalebox{0.8}{MSE} & \scalebox{0.8}{MAE}  
    & \scalebox{0.8}{MSE} & \scalebox{0.8}{MAE}  
    & \scalebox{0.8}{MSE} & \scalebox{0.8}{MAE} 
    & \scalebox{0.8}{MSE} & \scalebox{0.8}{MAE} 
    & \scalebox{0.8}{MSE} & \scalebox{0.8}{MAE} 
    & \scalebox{0.8}{MSE} & \scalebox{0.8}{MAE} 
    & \scalebox{0.8}{MSE} & \scalebox{0.8}{MAE} \\
    \toprule
    
    \multirow{5}{*}{\rotatebox{90}{ETTh1}}
    & \scalebox{0.8}{96} 
    & \boldres{\scalebox{0.8}{0.051}} & \boldres{\scalebox{0.8}{0.171}} 
    & \secondres{\scalebox{0.8}{0.054}} & \secondres{\scalebox{0.8}{0.174}} 
    & \scalebox{0.8}{0.066} & \scalebox{0.8}{0.195} 
    & \scalebox{0.8}{0.055} & \secondres{\scalebox{0.8}{0.174}} 
    & \scalebox{0.8}{0.078} & \scalebox{0.8}{0.227} 
    & \scalebox{0.8}{0.075} & \scalebox{0.8}{0.219}
    & \scalebox{0.8}{0.080} & \scalebox{0.8}{0.229} 
    & \scalebox{0.8}{0.153} & \scalebox{0.8}{0.326}
    & \scalebox{0.8}{0.167} & \scalebox{0.8}{0.340} 
    & \scalebox{0.8}{0.156} & \scalebox{0.8}{0.316} \\
    
    & \scalebox{0.8}{192} 
    & \boldres{\scalebox{0.8}{0.064}} & \boldres{\scalebox{0.8}{0.197}} 
    & \secondres{\scalebox{0.8}{0.068}} & \secondres{\scalebox{0.8}{0.199}} 
    & \scalebox{0.8}{0.075} & \scalebox{0.8}{0.213} 
    & \scalebox{0.8}{0.070} & \scalebox{0.8}{0.201} 
    & \scalebox{0.8}{0.084} & \scalebox{0.8}{0.235}
    & \scalebox{0.8}{0.114} & \scalebox{0.8}{0.270} 
    & \scalebox{0.8}{0.084} & \scalebox{0.8}{0.235}
    & \scalebox{0.8}{0.176} & \scalebox{0.8}{0.349}
    & \scalebox{0.8}{0.299} & \scalebox{0.8}{0.463} 
    & \scalebox{0.8}{0.176} & \scalebox{0.8}{0.338} \\
    
    & \scalebox{0.8}{336} 
    & \boldres{\scalebox{0.8}{0.071}} & \boldres{\scalebox{0.8}{0.210}} 
    & \secondres{\scalebox{0.8}{0.079}} & \secondres{\scalebox{0.8}{0.220}} 
    & \scalebox{0.8}{0.086} & \scalebox{0.8}{0.232} 
    & \scalebox{0.8}{0.085} & \scalebox{0.8}{0.227} 
    & \scalebox{0.8}{0.090} & \scalebox{0.8}{0.244} 
    & \scalebox{0.8}{0.160} & \scalebox{0.8}{0.324} 
    & \scalebox{0.8}{0.088} & \scalebox{0.8}{0.239}
    & \scalebox{0.8}{0.206} & \scalebox{0.8}{0.383}
    & \scalebox{0.8}{0.500} & \scalebox{0.8}{0.565} 
    & \scalebox{0.8}{0.211} & \scalebox{0.8}{0.378} \\
    
    & \scalebox{0.8}{720} 
    & \boldres{\scalebox{0.8}{0.086}} & \boldres{\scalebox{0.8}{0.233}} 
    & \secondres{\scalebox{0.8}{0.101}} & \secondres{\scalebox{0.8}{0.252}} 
    & \scalebox{0.8}{0.113} & \scalebox{0.8}{0.268} 
    & \scalebox{0.8}{0.128} & \scalebox{0.8}{0.286} 
    & \scalebox{0.8}{0.102} & \scalebox{0.8}{0.255}
    & \scalebox{0.8}{0.292} & \scalebox{0.8}{0.455} 
    & \scalebox{0.8}{0.130} & \scalebox{0.8}{0.291}
    & \scalebox{0.8}{0.190} & \scalebox{0.8}{0.347}
    & \scalebox{0.8}{0.576} & \scalebox{0.8}{0.635} 
    & \scalebox{0.8}{0.507} & \scalebox{0.8}{0.598} \\
    \cmidrule(lr){2-22}
    
    & \scalebox{0.8}{Avg} 
    & \boldres{\scalebox{0.8}{0.068}} & \boldres{\scalebox{0.8}{0.203}} 
    & \secondres{\scalebox{0.8}{0.076}} & \secondres{\scalebox{0.8}{0.211}} 
    & \scalebox{0.8}{0.085} & \scalebox{0.8}{0.227} 
    & \scalebox{0.8}{0.085} & \scalebox{0.8}{0.222} 
    & \scalebox{0.8}{0.089} & \scalebox{0.8}{0.240}
    & \scalebox{0.8}{0.160} & \scalebox{0.8}{0.317} 
    & \scalebox{0.8}{0.096} & \scalebox{0.8}{0.249}  
    & \scalebox{0.8}{0.181} & \scalebox{0.8}{0.351}
    & \scalebox{0.8}{0.386} & \scalebox{0.8}{0.501}
    & \scalebox{0.8}{0.263} & \scalebox{0.8}{0.408} \\
    \midrule

    \multirow{5}{*}{\rotatebox{90}{ETTh2}}
    & \scalebox{0.8}{96} 
    & \boldres{\scalebox{0.8}{0.111}} & \boldres{\scalebox{0.8}{0.258}} 
    & \secondres{\scalebox{0.8}{0.112}} & \secondres{\scalebox{0.8}{0.256}} 
    & \scalebox{0.8}{0.258} & \scalebox{0.8}{0.389} 
    & \scalebox{0.8}{0.125} & \scalebox{0.8}{0.272} 
    & \scalebox{0.8}{0.168} & \scalebox{0.8}{0.329}
    & \scalebox{0.8}{0.175} & \scalebox{0.8}{0.339} 
    & \scalebox{0.8}{0.188} & \scalebox{0.8}{0.349} 
    & \scalebox{0.8}{0.245} & \scalebox{0.8}{0.407}
    & \scalebox{0.8}{0.270} & \scalebox{0.8}{0.410} 
    & \scalebox{0.8}{0.250} & \scalebox{0.8}{0.402} \\
    
    & \scalebox{0.8}{192} 
    & \boldres{\scalebox{0.8}{0.136}} & \boldres{\scalebox{0.8}{0.291}} 
    & \secondres{\scalebox{0.8}{0.143}} & \secondres{\scalebox{0.8}{0.297}} 
    & \scalebox{0.8}{0.309} & \scalebox{0.8}{0.429} 
    & \scalebox{0.8}{0.165} & \scalebox{0.8}{0.321} 
    & \scalebox{0.8}{0.186} & \scalebox{0.8}{0.348} 
    & \scalebox{0.8}{0.214} & \scalebox{0.8}{0.381} 
    & \scalebox{0.8}{0.184} & \scalebox{0.8}{0.346} 
    & \scalebox{0.8}{0.176} & \scalebox{0.8}{0.349}
    & \scalebox{0.8}{0.348} & \scalebox{0.8}{0.481} 
    & \scalebox{0.8}{0.317} & \scalebox{0.8}{0.453} \\
    
    & \scalebox{0.8}{336} 
    & \boldres{\scalebox{0.8}{0.149}} & \boldres{\scalebox{0.8}{0.311}} 
    & \secondres{\scalebox{0.8}{0.157}} & \secondres{\scalebox{0.8}{0.317}} 
    & \scalebox{0.8}{0.353} & \scalebox{0.8}{0.462} 
    & \scalebox{0.8}{0.199} & \scalebox{0.8}{0.359} 
    & \scalebox{0.8}{0.192} & \scalebox{0.8}{0.355} 
    & \scalebox{0.8}{0.304} & \scalebox{0.8}{0.455} 
    & \scalebox{0.8}{0.190} & \scalebox{0.8}{0.348} 
    & \scalebox{0.8}{0.206} & \scalebox{0.8}{0.383}
    & \scalebox{0.8}{0.383} & \scalebox{0.8}{0.509} 
    & \scalebox{0.8}{0.323} & \scalebox{0.8}{0.460} \\
    
    & \scalebox{0.8}{720} 
    & \boldres{\scalebox{0.8}{0.169}} & \boldres{\scalebox{0.8}{0.335}} 
    & \scalebox{0.8}{0.213} & \scalebox{0.8}{0.372} 
    & \scalebox{0.8}{0.538} & \scalebox{0.8}{0.585} 
    & \scalebox{0.8}{0.298} & \scalebox{0.8}{0.447} 
    & \scalebox{0.8}{0.231} & \scalebox{0.8}{0.386} 
    & \scalebox{0.8}{0.536} & \scalebox{0.8}{0.606} 
    & \scalebox{0.8}{0.203} & \scalebox{0.8}{0.366}
    & \secondres{\scalebox{0.8}{0.190}} & \secondres{\scalebox{0.8}{0.347}}
    & \scalebox{0.8}{0.578} & \scalebox{0.8}{0.607} 
    & \scalebox{0.8}{0.388} & \scalebox{0.8}{0.500} \\
    \cmidrule(lr){2-22}
    
    & \scalebox{0.8}{Avg} 
    & \boldres{\scalebox{0.8}{0.141}} & \boldres{\scalebox{0.8}{0.299}} 
    & \secondres{\scalebox{0.8}{0.156}} & \secondres{\scalebox{0.8}{0.311}} 
    & \scalebox{0.8}{0.365} & \scalebox{0.8}{0.466} 
    & \scalebox{0.8}{0.197} & \scalebox{0.8}{0.350} 
    & \scalebox{0.8}{0.194} & \scalebox{0.8}{0.355} 
    & \scalebox{0.8}{0.307} & \scalebox{0.8}{0.445} 
    & \scalebox{0.8}{0.191} & \scalebox{0.8}{0.352} 
    & \scalebox{0.8}{0.181} & \scalebox{0.8}{0.351}
    & \scalebox{0.8}{0.395} & \scalebox{0.8}{0.502} 
    & \scalebox{0.8}{0.320} & \scalebox{0.8}{0.454} \\
    \midrule 
    
    \multirow{5}{*}{\rotatebox{90}{ETTm1}}
    & \scalebox{0.8}{96} 
    & \boldres{\scalebox{0.8}{0.026}} & \boldres{\scalebox{0.8}{0.122}} 
    & \secondres{\scalebox{0.8}{0.027}} & \secondres{\scalebox{0.8}{0.123}} 
    & \scalebox{0.8}{0.028} & \scalebox{0.8}{0.124} 
    & \scalebox{0.8}{0.030} & \scalebox{0.8}{0.128} 
    & \scalebox{0.8}{0.038} & \scalebox{0.8}{0.147} 
    & \scalebox{0.8}{0.036} & \scalebox{0.8}{0.148} 
    & \scalebox{0.8}{0.031} & \scalebox{0.8}{0.134}
    & \scalebox{0.8}{0.066} & \scalebox{0.8}{0.199}
    & \scalebox{0.8}{0.038} & \scalebox{0.8}{0.154} 
    & \scalebox{0.8}{0.030} & \scalebox{0.8}{0.131} \\
    
    & \scalebox{0.8}{192} 
    & \boldres{\scalebox{0.8}{0.039}} & \boldres{\scalebox{0.8}{0.149}} 
    & \secondres{\scalebox{0.8}{0.041}} & \secondres{\scalebox{0.8}{0.156}} 
    & \scalebox{0.8}{0.044} & \scalebox{0.8}{0.157} 
    & \scalebox{0.8}{0.045} & \scalebox{0.8}{0.158} 
    & \scalebox{0.8}{0.062} & \scalebox{0.8}{0.194} 
    & \scalebox{0.8}{0.053} & \scalebox{0.8}{0.178} 
    & \scalebox{0.8}{0.049} & \scalebox{0.8}{0.172} 
    & \scalebox{0.8}{0.076} & \scalebox{0.8}{0.226}
    & \scalebox{0.8}{0.055} & \scalebox{0.8}{0.185} 
    & \scalebox{0.8}{0.052} & \scalebox{0.8}{0.176} \\
    
    & \scalebox{0.8}{336} 
    & \boldres{\scalebox{0.8}{0.048}} & \boldres{\scalebox{0.8}{0.165}} 
    & \secondres{\scalebox{0.8}{0.054}} & \scalebox{0.8}{0.181} 
    & \scalebox{0.8}{0.057} & \scalebox{0.8}{0.181} 
    & \scalebox{0.8}{0.056} & \secondres{\scalebox{0.8}{0.177}} 
    & \scalebox{0.8}{0.069} & \scalebox{0.8}{0.203} 
    & \scalebox{0.8}{0.065} & \scalebox{0.8}{0.197} 
    & \scalebox{0.8}{0.061} & \scalebox{0.8}{0.195}
    & \scalebox{0.8}{0.203} & \scalebox{0.8}{0.381}
    & \scalebox{0.8}{0.077} & \scalebox{0.8}{0.219} 
    & \scalebox{0.8}{0.069} & \scalebox{0.8}{0.201} \\
    
    & \scalebox{0.8}{720} 
    & \boldres{\scalebox{0.8}{0.058}} & \boldres{\scalebox{0.8}{0.182}} 
    & \secondres{\scalebox{0.8}{0.063}} & \secondres{\scalebox{0.8}{0.198}} 
    & \scalebox{0.8}{0.067} & \scalebox{0.8}{0.199} 
    & \scalebox{0.8}{0.068} & \scalebox{0.8}{0.202} 
    & \scalebox{0.8}{0.080} & \scalebox{0.8}{0.225} 
    & \scalebox{0.8}{0.080} & \scalebox{0.8}{0.221}
    & \scalebox{0.8}{0.077} & \scalebox{0.8}{0.224} 
    & \scalebox{0.8}{0.102} & \scalebox{0.8}{0.264}
    & \scalebox{0.8}{0.102} & \scalebox{0.8}{0.271} 
    & \scalebox{0.8}{0.086} & \scalebox{0.8}{0.226} \\
    \cmidrule(lr){2-22}
    
    & \scalebox{0.8}{Avg} 
    & \boldres{\scalebox{0.8}{0.043}} & \boldres{\scalebox{0.8}{0.155}} 
    & \secondres{\scalebox{0.8}{0.046}} & \secondres{\scalebox{0.8}{0.165}} 
    & \scalebox{0.8}{0.049} & \secondres{\scalebox{0.8}{0.165}} 
    & \scalebox{0.8}{0.050} & \scalebox{0.8}{0.166} 
    & \scalebox{0.8}{0.062} & \scalebox{0.8}{0.192} 
    & \scalebox{0.8}{0.059} & \scalebox{0.8}{0.186} 
    & \scalebox{0.8}{0.055} & \scalebox{0.8}{0.181}
    & \scalebox{0.8}{0.112} & \scalebox{0.8}{0.268}
    & \scalebox{0.8}{0.068} & \scalebox{0.8}{0.207} 
    & \scalebox{0.8}{0.059} & \scalebox{0.8}{0.184} \\
    \midrule 
    
    \multirow{5}{*}{\rotatebox{90}{ETTm2}}
    & \scalebox{0.8}{96} 
    & \boldres{\scalebox{0.8}{0.059}} & \boldres{\scalebox{0.8}{0.175}} 
    & \secondres{\scalebox{0.8}{0.059}} & \secondres{\scalebox{0.8}{0.177}} 
    & \scalebox{0.8}{0.063} & \scalebox{0.8}{0.182} 
    & \scalebox{0.8}{0.075} & \scalebox{0.8}{0.199} 
    & \scalebox{0.8}{0.105} & \scalebox{0.8}{0.240} 
    & \scalebox{0.8}{0.097} & \scalebox{0.8}{0.243} 
    & \scalebox{0.8}{0.081} & \scalebox{0.8}{0.213}  
    & \scalebox{0.8}{0.181} & \scalebox{0.8}{0.338}
    & \scalebox{0.8}{0.164} & \scalebox{0.8}{0.327}     
    & \scalebox{0.8}{0.071} & \scalebox{0.8}{0.197} \\
    
    & \scalebox{0.8}{192} 
    & \boldres{\scalebox{0.8}{0.085}} & \boldres{\scalebox{0.8}{0.218}} 
    & \scalebox{0.8}{0.094} & \secondres{\scalebox{0.8}{0.228}} 
    & \secondres{\scalebox{0.8}{0.092}} & \secondres{\scalebox{0.8}{0.228}} 
    & \scalebox{0.8}{0.106} & \scalebox{0.8}{0.246} 
    & \scalebox{0.8}{0.153} & \scalebox{0.8}{0.295} 
    & \scalebox{0.8}{0.135} & \scalebox{0.8}{0.286} 
    & \scalebox{0.8}{0.123} & \scalebox{0.8}{0.272} 
    & \scalebox{0.8}{0.204} & \scalebox{0.8}{0.364}
    & \scalebox{0.8}{0.193} & \scalebox{0.8}{0.344}
    & \scalebox{0.8}{0.107} & \scalebox{0.8}{0.250} \\
        
    & \scalebox{0.8}{336} 
    & \boldres{\scalebox{0.8}{0.108}} & \boldres{\scalebox{0.8}{0.251}} 
    & \scalebox{0.8}{0.121} & \scalebox{0.8}{0.265} 
    & \secondres{\scalebox{0.8}{0.117}} & \secondres{\scalebox{0.8}{0.263}} 
    & \scalebox{0.8}{0.133} & \scalebox{0.8}{0.282} 
    & \scalebox{0.8}{0.195} & \scalebox{0.8}{0.340} 
    & \scalebox{0.8}{0.164} & \scalebox{0.8}{0.322} 
    & \scalebox{0.8}{0.146} & \scalebox{0.8}{0.297} 
    & \scalebox{0.8}{0.242} & \scalebox{0.8}{0.408}
    & \scalebox{0.8}{0.194} & \scalebox{0.8}{0.364} 
    & \scalebox{0.8}{0.135} & \scalebox{0.8}{0.285} \\
    
    & \scalebox{0.8}{720} 
    & \boldres{\scalebox{0.8}{0.146}} & \boldres{\scalebox{0.8}{0.302}} 
    & \scalebox{0.8}{0.155} & \secondres{\scalebox{0.8}{0.310}} 
    & \secondres{\scalebox{0.8}{0.153}} & \secondres{\scalebox{0.8}{0.310}} 
    & \scalebox{0.8}{0.173} & \scalebox{0.8}{0.332} 
    & \scalebox{0.8}{0.191} & \scalebox{0.8}{0.340} 
    & \scalebox{0.8}{0.198} & \scalebox{0.8}{0.363} 
    & \scalebox{0.8}{0.173} & \scalebox{0.8}{0.328}  
    & \scalebox{0.8}{0.259} & \scalebox{0.8}{0.424}
    & \scalebox{0.8}{0.281} & \scalebox{0.8}{0.429} 
    & \scalebox{0.8}{0.177} & \scalebox{0.8}{0.332} \\
    \cmidrule(lr){2-22}
    
    & \scalebox{0.8}{Avg} 
    & \boldres{\scalebox{0.8}{0.100}} & \boldres{\scalebox{0.8}{0.237}} 
    & \scalebox{0.8}{0.107} & \secondres{\scalebox{0.8}{0.245}} 
    & \secondres{\scalebox{0.8}{0.106}} & \scalebox{0.8}{0.246} 
    & \scalebox{0.8}{0.122} & \scalebox{0.8}{0.265} 
    & \scalebox{0.8}{0.161} & \scalebox{0.8}{0.304} 
    & \scalebox{0.8}{0.149} & \scalebox{0.8}{0.304} 
    & \scalebox{0.8}{0.131} & \scalebox{0.8}{0.278} 
    & \scalebox{0.8}{0.222} & \scalebox{0.8}{0.384}
    & \scalebox{0.8}{0.208} & \scalebox{0.8}{0.366} 
    & \scalebox{0.8}{0.123} & \scalebox{0.8}{0.266} \\
    \midrule 
    
    \multirow{5}{*}{\rotatebox{90}{Weather}}
    & \scalebox{0.8}{96} 
    & \boldres{\scalebox{0.8}{0.001}} & \boldres{\scalebox{0.8}{0.020}} 
    & \boldres{\scalebox{0.8}{0.001}} & \secondres{\scalebox{0.8}{0.021}} 
    & \boldres{\scalebox{0.8}{0.001}} & \scalebox{0.8}{0.028} 
    & \boldres{\scalebox{0.8}{0.001}} & \scalebox{0.8}{0.027} 
    & \secondres{\scalebox{0.8}{0.002}} & \scalebox{0.8}{0.031} 
    & \secondres{\scalebox{0.8}{0.002}} & \scalebox{0.8}{0.033} 
    & \secondres{\scalebox{0.8}{0.002}} & \scalebox{0.8}{0.034}  
    & \scalebox{0.8}{0.008} & \scalebox{0.8}{0.076}
    & \scalebox{0.8}{0.003} & \scalebox{0.8}{0.040}     
    & \scalebox{0.8}{0.007} & \scalebox{0.8}{0.072} \\
    
    & \scalebox{0.8}{192} 
    & \boldres{\scalebox{0.8}{0.001}} & \boldres{\scalebox{0.8}{0.025}} 
    & \boldres{\scalebox{0.8}{0.001}} & \boldres{\scalebox{0.8}{0.025}} 
    & \secondres{\scalebox{0.8}{0.002}} & \secondres{\scalebox{0.8}{0.031}} 
    & \secondres{\scalebox{0.8}{0.002}} & \secondres{\scalebox{0.8}{0.031}} 
    & \secondres{\scalebox{0.8}{0.002}} & \scalebox{0.8}{0.032} 
    & \secondres{\scalebox{0.8}{0.002}} & \scalebox{0.8}{0.033} 
    & \secondres{\scalebox{0.8}{0.002}} & \scalebox{0.8}{0.035}  
    & \scalebox{0.8}{0.105} & \scalebox{0.8}{0.092}
    & \scalebox{0.8}{0.003} & \scalebox{0.8}{0.042} 
    & \scalebox{0.8}{0.008} & \scalebox{0.8}{0.076} \\
    
    & \scalebox{0.8}{336} 
    & \boldres{\scalebox{0.8}{0.002}} & \boldres{\scalebox{0.8}{0.028}} 
    & \boldres{\scalebox{0.8}{0.002}} & \boldres{\scalebox{0.8}{0.028}} 
    & \boldres{\scalebox{0.8}{0.002}} & \scalebox{0.8}{0.034} 
    & \boldres{\scalebox{0.8}{0.002}} & \scalebox{0.8}{0.034} 
    & \boldres{\scalebox{0.8}{0.002}} & \secondres{\scalebox{0.8}{0.033}} 
    & \boldres{\scalebox{0.8}{0.002}} & \scalebox{0.8}{0.034} 
    & \boldres{\scalebox{0.8}{0.002}} & \scalebox{0.8}{0.034} 
    & \scalebox{0.8}{0.009} & \scalebox{0.8}{0.085}
    & \secondres{\scalebox{0.8}{0.004}} & \scalebox{0.8}{0.051} 
    & \scalebox{0.8}{0.008} & \scalebox{0.8}{0.079} \\
    
    & \scalebox{0.8}{720}  
    & \boldres{\scalebox{0.8}{0.002}} & \boldres{\scalebox{0.8}{0.032}} 
    & \boldres{\scalebox{0.8}{0.002}} & \boldres{\scalebox{0.8}{0.032}} 
    & \boldres{\scalebox{0.8}{0.002}} & \scalebox{0.8}{0.037} 
    & \boldres{\scalebox{0.8}{0.002}} & \scalebox{0.8}{0.039} 
    & \boldres{\scalebox{0.8}{0.002}} & \secondres{\scalebox{0.8}{0.034}} 
    & \boldres{\scalebox{0.8}{0.002}} & \scalebox{0.8}{0.036} 
    & \secondres{\scalebox{0.8}{0.003}} & \scalebox{0.8}{0.041}  
    & \scalebox{0.8}{0.010} & \scalebox{0.8}{0.090}
    & \scalebox{0.8}{0.005} & \scalebox{0.8}{0.056}     
    & \scalebox{0.8}{0.008} & \scalebox{0.8}{0.078} \\
    \cmidrule(lr){2-22}
    
    & \scalebox{0.8}{Avg} 
    & \boldres{\scalebox{0.8}{0.001}} & \boldres{\scalebox{0.8}{0.026}} 
    & \secondres{\scalebox{0.8}{0.002}} & \secondres{\scalebox{0.8}{0.027}}  
    & \secondres{\scalebox{0.8}{0.002}} & \scalebox{0.8}{0.033} 
    & \secondres{\scalebox{0.8}{0.002}} & \scalebox{0.8}{0.033} 
    & \secondres{\scalebox{0.8}{0.002}} & \scalebox{0.8}{0.033}
    & \secondres{\scalebox{0.8}{0.002}} & \scalebox{0.8}{0.034} 
    & \secondres{\scalebox{0.8}{0.002}} & \scalebox{0.8}{0.036} 
    & \scalebox{0.8}{0.033} & \scalebox{0.8}{0.086}
    & \scalebox{0.8}{0.004} & \scalebox{0.8}{0.047} 
    & \scalebox{0.8}{0.008} & \scalebox{0.8}{0.076} \\
    \midrule 
    
    \multirow{5}{*}{\rotatebox{90}{ECL}}
    & \scalebox{0.8}{96} 
    & \boldres{\scalebox{0.8}{0.159}} & \boldres{\scalebox{0.8}{0.279}} 
    & \secondres{\scalebox{0.8}{0.173}} & \scalebox{0.8}{0.295} 
    & \scalebox{0.8}{0.176} & \secondres{\scalebox{0.8}{0.289}} 
    & \scalebox{0.8}{0.175} & \scalebox{0.8}{0.296} 
    & \scalebox{0.8}{0.235} & \scalebox{0.8}{0.339}
    & \scalebox{0.8}{0.248} & \scalebox{0.8}{0.368} 
    & \scalebox{0.8}{0.248} & \scalebox{0.8}{0.370} 
    & \scalebox{0.8}{0.306} & \scalebox{0.8}{0.415}
    & \scalebox{0.8}{0.285} & \scalebox{0.8}{0.388} 
    & \scalebox{0.8}{0.222} & \scalebox{0.8}{0.341} \\
    
    & \scalebox{0.8}{192} 
    & \boldres{\scalebox{0.8}{0.187}} & \boldres{\scalebox{0.8}{0.305}} 
    & \secondres{\scalebox{0.8}{0.199}} & \secondres{\scalebox{0.8}{0.315}} 
    & \scalebox{0.8}{0.202} & \secondres{\scalebox{0.8}{0.315}} 
    & \scalebox{0.8}{0.207} & \scalebox{0.8}{0.321} 
    & \scalebox{0.8}{0.271} & \scalebox{0.8}{0.364}
    & \scalebox{0.8}{0.291} & \scalebox{0.8}{0.403} 
    & \scalebox{0.8}{0.332} & \scalebox{0.8}{0.437} 
    & \scalebox{0.8}{0.338} & \scalebox{0.8}{0.439}
    & \scalebox{0.8}{0.382} & \scalebox{0.8}{0.475} 
    & \scalebox{0.8}{0.255} & \scalebox{0.8}{0.364} \\
    
    & \scalebox{0.8}{336} 
    & \boldres{\scalebox{0.8}{0.208}} & \boldres{\scalebox{0.8}{0.325}} 
    & \scalebox{0.8}{0.226} & \scalebox{0.8}{0.340} 
    & \secondres{\scalebox{0.8}{0.215}} & \secondres{\scalebox{0.8}{0.331}} 
    & \scalebox{0.8}{0.240} & \scalebox{0.8}{0.352} 
    & \scalebox{0.8}{0.309} & \scalebox{0.8}{0.398} 
    & \scalebox{0.8}{0.316} & \scalebox{0.8}{0.425} 
    & \scalebox{0.8}{0.365} & \scalebox{0.8}{0.458}
    & \scalebox{0.8}{0.373} & \scalebox{0.8}{0.463}
    & \scalebox{0.8}{0.362} & \scalebox{0.8}{0.447} 
    & \scalebox{0.8}{0.287} & \scalebox{0.8}{0.393} \\
    
    & \scalebox{0.8}{720} 
    & \boldres{\scalebox{0.8}{0.223}} & \boldres{\scalebox{0.8}{0.345}} 
    & \scalebox{0.8}{0.249} & \scalebox{0.8}{0.367} 
    & \secondres{\scalebox{0.8}{0.232}} & \secondres{\scalebox{0.8}{0.355}} 
    & \scalebox{0.8}{0.298} & \scalebox{0.8}{0.419} 
    & \scalebox{0.8}{0.354} & \scalebox{0.8}{0.446} 
    & \scalebox{0.8}{0.318} & \scalebox{0.8}{0.426} 
    & \scalebox{0.8}{0.362} & \scalebox{0.8}{0.460}
    & \scalebox{0.8}{0.390} & \scalebox{0.8}{0.478}
    & \scalebox{0.8}{0.377} & \scalebox{0.8}{0.474} 
    & \scalebox{0.8}{0.292} & \scalebox{0.8}{0.404} \\
    
    \cmidrule(lr){2-22}
    & \scalebox{0.8}{Avg}  
    & \boldres{\scalebox{0.8}{0.194}} & \boldres{\scalebox{0.8}{0.314}} 
    & \scalebox{0.8}{0.212} & \scalebox{0.8}{0.329} 
    & \secondres{\scalebox{0.8}{0.206}} & \secondres{\scalebox{0.8}{0.323}} 
    & \scalebox{0.8}{0.230} & \scalebox{0.8}{0.347} 
    & \scalebox{0.8}{0.292} & \scalebox{0.8}{0.387} 
    & \scalebox{0.8}{0.293} & \scalebox{0.8}{0.406} 
    & \scalebox{0.8}{0.327} & \scalebox{0.8}{0.431} 
    & \scalebox{0.8}{0.352} & \scalebox{0.8}{0.449}
    & \scalebox{0.8}{0.352} & \scalebox{0.8}{0.446} 
    & \scalebox{0.8}{0.264} & \scalebox{0.8}{0.376} \\
    \midrule
    
    \multirow{5}{*}{\rotatebox{90}{Traffic}}
    & \scalebox{0.8}{96} 
    & \boldres{\scalebox{0.8}{0.101}} & \boldres{\scalebox{0.8}{0.169}} 
    & \scalebox{0.8}{-} & \scalebox{0.8}{-} 
    & \scalebox{0.8}{-} & \scalebox{0.8}{-} 
    & \secondres{\scalebox{0.8}{0.109}} & \secondres{\scalebox{0.8}{0.185}} 
    & \scalebox{0.8}{0.149} & \scalebox{0.8}{0.250} 
    & \scalebox{0.8}{0.124} & \scalebox{0.8}{0.210} 
    & \scalebox{0.8}{0.146} & \scalebox{0.8}{0.245} 
    & \scalebox{0.8}{0.187} & \scalebox{0.8}{0.291}
    & \scalebox{0.8}{0.164} & \scalebox{0.8}{0.259}     
    & \scalebox{0.8}{0.164} & \scalebox{0.8}{0.270} \\
    
    & \scalebox{0.8}{192} 
    & \boldres{\scalebox{0.8}{0.109}} & \boldres{\scalebox{0.8}{0.179}} 
    & \scalebox{0.8}{-} & \scalebox{0.8}{-} 
    & \scalebox{0.8}{-} & \scalebox{0.8}{-} 
    & \secondres{\scalebox{0.8}{0.118}} & \secondres{\scalebox{0.8}{0.197}} 
    & \scalebox{0.8}{0.156} & \scalebox{0.8}{0.258} 
    & \scalebox{0.8}{0.131} & \scalebox{0.8}{0.221} 
    & \scalebox{0.8}{0.152} & \scalebox{0.8}{0.253} 
    & \scalebox{0.8}{0.210} & \scalebox{0.8}{0.316}
    & \scalebox{0.8}{0.225} & \scalebox{0.8}{0.317} 
    & \scalebox{0.8}{0.179} & \scalebox{0.8}{0.290} \\
    
    & \scalebox{0.8}{336} 
    & \boldres{\scalebox{0.8}{0.111}} & \boldres{\scalebox{0.8}{0.187}} 
    & \scalebox{0.8}{-} & \scalebox{0.8}{-} 
    & \scalebox{0.8}{-} & \scalebox{0.8}{-} 
    & \secondres{\scalebox{0.8}{0.121}} & \secondres{\scalebox{0.8}{0.204}} 
    & \scalebox{0.8}{0.154} & \scalebox{0.8}{0.258} 
    & \scalebox{0.8}{0.136} & \scalebox{0.8}{0.232} 
    & \scalebox{0.8}{0.152} & \scalebox{0.8}{0.255}
    & \scalebox{0.8}{0.224} & \scalebox{0.8}{0.333}
    & \scalebox{0.8}{0.297} & \scalebox{0.8}{0.375}     
    & \scalebox{0.8}{0.190} & \scalebox{0.8}{0.308} \\
    
    & \scalebox{0.8}{720}  
    & \boldres{\scalebox{0.8}{0.128}} & \boldres{\scalebox{0.8}{0.208}} 
    & \scalebox{0.8}{-} & \scalebox{0.8}{-} 
    & \scalebox{0.8}{-} & \scalebox{0.8}{-} 
    & \secondres{\scalebox{0.8}{0.141}} & \secondres{\scalebox{0.8}{0.226}} 
    & \scalebox{0.8}{0.168} & \scalebox{0.8}{0.271} 
    & \scalebox{0.8}{0.163} & \scalebox{0.8}{0.265} 
    & \scalebox{0.8}{0.165} & \scalebox{0.8}{0.267}  
    & \scalebox{0.8}{0.267} & \scalebox{0.8}{0.372}
    & \scalebox{0.8}{0.411} & \scalebox{0.8}{0.378} 
    & \scalebox{0.8}{0.280} & \scalebox{0.8}{0.400} \\
    \cmidrule(lr){2-22}
    
    & \scalebox{0.8}{Avg} 
    & \boldres{\scalebox{0.8}{0.112}} & \boldres{\scalebox{0.8}{0.186}} 
    & \scalebox{0.8}{-} & \scalebox{0.8}{-} 
    & \scalebox{0.8}{-} & \scalebox{0.8}{-} 
    & \secondres{\scalebox{0.8}{0.122}} & \secondres{\scalebox{0.8}{0.203}} 
    & \scalebox{0.8}{0.157} & \scalebox{0.8}{0.259} 
    & \scalebox{0.8}{0.139} & \scalebox{0.8}{0.232} 
    & \scalebox{0.8}{0.154} & \scalebox{0.8}{0.255}  
    & \scalebox{0.8}{0.222} & \scalebox{0.8}{0.328}
    & \scalebox{0.8}{0.274} & \scalebox{0.8}{0.332} 
    & \scalebox{0.8}{0.203} & \scalebox{0.8}{0.317} \\
    \bottomrule
  \end{tabular}
    \end{small}
  \end{threeparttable}
}
\end{table}

\clearpage
\subsection{Full Results of Multi-Modal Covariate-Aware Forecasting}

Table~\ref{tab:full_result_timemmd} reports the full results on the Time-MMD benchmark. We employ the Qwen3-Embedding~\citep{qwen3embedding} as the backbone in CoRA to derive text embeddings. Compared to Sundial~\citep{liu2025sundial} in the zero-shot setting and UniCA~\citep{han2025unica}, CoRA consistently achieves superior performance across both deterministic metrics (MSE, MAE) and probabilistic metrics (CRPS). This demonstrates that CoRA successfully captures meaningful interactions between temporal dynamics and textual covariates. These results further highlight the strength of CoRA as a general and powerful strategy for integrating multi-modal information into TSFMs.

\begin{table}[htbp]
  \caption{Full results of multi-modal covariate-aware forecasting task on TimeMMD dataset.}
  \label{tab:full_result_timemmd}
  \centering
  \resizebox{\columnwidth}{!}{
  \small 

  \begin{tabular}{@{}l|l| *{12}{c}@{}} 
    \toprule
    & \multirow{2}{*}{\textbf{Models}} & 
    \multicolumn{1}{c}{\textbf{CoRA}} &
    \multicolumn{1}{c}{\textbf{UniCA}} &
    \multicolumn{1}{c}{\textbf{Sundial}} &
    \multicolumn{1}{c}{\textbf{NBEATS}} &
    \multicolumn{1}{c}{\textbf{PatchTST}} &
    \multicolumn{1}{c}{\textbf{DeepAR}} &
    \multicolumn{1}{c}{\textbf{TFT}} &
    \multicolumn{1}{c}{\textbf{TiDE}} &
    \multicolumn{1}{c}{\textbf{Time-LLM}} &
    \multicolumn{1}{c}{\textbf{TTM}} &
    \multicolumn{1}{c}{\textbf{Moirai}} &
    \multicolumn{1}{c}{\textbf{TabPFN-TS}}  \\
    & & 
    \textbf{(Ours)} &
    \citeyearpar{han2025unica} &
    \citeyearpar{liu2025sundial} &
    \citeyearpar{olivares2023neural} &
    \citeyearpar{nie2022time} &
    \citeyearpar{salinas2020deepar} &
    \citeyearpar{lim2021temporal} &
    \citeyearpar{das2023long} &
    \citeyearpar{jin2023time} &
    \citeyearpar{ekambaram2024tiny} &
    \citeyearpar{woo2024unified} &
    \citeyearpar{hoo2025tables} \\

    \midrule 
    
    \multirow{4}{*}{\rotatebox{90}{\textbf{\scalebox{0.8}{Average}}}}
    & \textbf{Average} 
    & \textbf{0.641} & 0.661 & 0.662 & 0.882 
    & 0.933 & 1.361 & 0.947 & 0.927 
    & 0.835 & 0.820 & 0.751 & 0.795 \\
    & \textbf{MSE} 
    & \textbf{0.580} & 0.591 & 0.591 & 0.782 
    & 0.793 & 1.605 & 0.992 & 0.869 
    & 0.723 & 0.685 & 0.696 & 0.787 \\
    & \textbf{MAE} 
    & \textbf{0.690} & 0.716 & 0.716 & 0.884 
    & 1.009 & 1.219 & 0.958 & 0.976 
    & 0.847 & 0.866 & 0.821 & 0.837 \\
    & \textbf{CRPS} 
    & \textbf{0.653} & 0.677 & 0.678 & 0.980 
    & 0.996 & 1.260 & 0.891 & 0.937 
    & 0.935 & 0.909 & 0.735 & 0.762 \\
    \midrule
    \multirow{4}{*}{\rotatebox{90}{\textbf{\scalebox{0.8}{Climate}}}}
    & \textbf{Average} 
    & 0.536 & 0.567 & 0.567 & 0.668 
    & 0.724 & 0.737 & 0.695 & 0.575 
    & 0.634 & 0.526 & 0.596 & \textbf{0.525} \\
    & \textbf{MSE} 
    & 0.440 & 0.487 & 0.487 & 0.519
    & 0.640 & 0.623 & 0.599 & 0.465 
    & 0.468 & 0.408 & 0.488 & \textbf{0.407} \\
    & \textbf{MAE} 
    & \textbf{0.562} & 0.595 & 0.595 & 0.712 
    & 0.788 & 0.779 & 0.768 & 0.685 
    & 0.687 & 0.635 & 0.706 & 0.638 \\
    & \textbf{CRPS} 
    & 0.607 & 0.620 & 0.620 & 0.773 
    & 0.743 & 0.809 & 0.719 & 0.574 
    & 0.746 & 0.535 & 0.593 & \textbf{0.529} \\
    \midrule
    \multirow{4}{*}{\rotatebox{90}{\textbf{\scalebox{0.8}{Energy}}}}
    & \textbf{Average} 
    & \textbf{0.888} & 0.892 & 0.892 & 1.611 
    & 1.274 & 3.768 & 1.018 & 1.303 
    & 1.253 & 1.216 & 1.011 & 1.233 \\
    & \textbf{MSE} 
    & \textbf{0.838} & 0.846 & 0.846 & 1.706 
    & 1.305 & 6.328 & 1.047 & 1.391 
    & 1.217 & 1.019 & 1.024 & 1.370 \\
    & \textbf{MAE} 
    & \textbf{0.928} & 0.930 & 0.930 & 1.429 
    & 1.252 & 2.368 & 1.004 & 1.138 
    & 1.161 & 1.042 & 1.035 & 1.163 \\
    & \textbf{CRPS} 
    & \textbf{0.897} & 0.900 & 0.900 & 1.699
    & 1.266 & 2.607 & 1.004 & 1.379
    & 1.380 & 1.587 & 0.975 & 1.167 \\
    \midrule
    \multirow{4}{*}{\rotatebox{90}{\textbf{\scalebox{0.8}{Environment}}}}
    & \textbf{Average}
    & \textbf{0.604} & 0.608  & 0.608 & 0.725
    & 0.644 & 0.689  & 0.638 & 0.638
    & 0.699 & 0.644  & 0.641 & 0.644 \\
    & \textbf{MSE} 
    & 0.527 & \textbf{0.519} & \textbf{0.519} & 0.628
    & 0.589 & 0.648  & 0.601 & 0.572 
    & 0.617 & 0.546  & 0.623 & 0.611 \\
    & \textbf{MAE} 
    & \textbf{0.730} & 0.742  & 0.742 & 0.809 
    & 0.785 & 0.822  & 0.763 & 0.778
    & 0.774 & 0.777  & 0.756 & 0.772 \\
    & \textbf{CRPS} 
    & 0.554 & 0.564  & 0.564 & 0.739
    & 0.558 & 0.596  & 0.550 & 0.564
    & 0.707 & 0.609  & \textbf{0.543} & 0.550 \\
    
    \midrule
    \multirow{4}{*}{\rotatebox{90}{\textbf{\scalebox{0.8}{Health}}}}
    & \textbf{Average}
    & \textbf{0.609} & 0.637  & 0.637 & 0.873 
    & 0.930 & 1.131  & 1.014 & 0.973
    & 0.862 & 0.966  & 0.776 & 0.969 \\
    & \textbf{MSE} 
    & \textbf{0.487} & 0.514  & 0.513 & 0.739
    & 0.874 & 1.023  & 1.059 & 0.916
    & 0.735 & 0.906  & 0.722 & 0.964 \\
    & \textbf{MAE} 
    & \textbf{0.687} & 0.706  & 0.706 & 0.860
    & 0.928 & 1.118  & 1.004 & 0.992
    & 0.846 & 0.989  & 0.821 & 1.008 \\
    & \textbf{CRPS} 
    & \textbf{0.653} & 0.692  & 0.692 & 1.020
    & 0.989 & 1.251  & 0.979 & 1.010
    & 1.004 & 1.002  & 0.786 & 0.936 \\
    \midrule
    \multirow{4}{*}{\rotatebox{90}{\textbf{\scalebox{0.8}{Security}}}}
    & \textbf{Average}
    & \textbf{0.657} & 0.688  & 0.689 & 0.847
    & 1.170 & 1.419  & 1.399 & 1.521
    & 0.862 & 0.763  & 0.746 & 0.678 \\
    & \textbf{MSE}
    & \textbf{0.595} & 0.620  & 0.620 & 0.692
    & 0.882 & 1.078  & 1.614 & 1.260
    & 0.690 & 0.676  & 0.669 & 0.612 \\
    & \textbf{MAE} 
    & \textbf{0.736} & 0.763  & 0.764 & 0.927
    & 1.332 & 1.607  & 1.409 & 1.767 
    & 0.951 & 0.880  & 0.856 & 0.764 \\
    & \textbf{CRPS} 
    & \textbf{0.641} & 0.682  & 0.683 & 0.922
    & 1.295 & 1.571  & 1.175 & 1.535
    & 0.946 & 0.732  & 0.714 & 0.657 \\
    \midrule
    \multirow{4}{*}{\rotatebox{90}{\textbf{\scalebox{0.8}{SocialGood}}}}
    & \textbf{Average} 
    & \textbf{0.745} & 0.778 & 0.778 & 0.863 
    & 1.219 & 1.386 & 1.264 & 0.952 
    & 1.052 & 0.980 & 0.781 & 0.903 \\
    & \textbf{MSE} 
    & 0.784 & 0.762 & 0.762 & 0.780
    & 0.877 & 1.231 & 1.469 & 0.973
    & 0.932 & 0.816 & \textbf{0.735} & 0.917 \\
    & \textbf{MAE} 
    & \textbf{0.719} & 0.788 & 0.788 & 0.843
    & 1.347 & 1.403 & 1.172 & 0.943
    & 1.036 & 1.062 & 0.803 & 0.912 \\
    & \textbf{CRPS}
    & \textbf{0.733} & 0.784 & 0.785 & 0.967
    & 1.434 & 1.523 & 1.150 & 0.941
    & 1.188 & 1.061 & 0.804 & 0.881 \\
    \midrule
    \multirow{4}{*}{\rotatebox{90}{\textbf{\scalebox{0.8}{Traffic}}}}
    & \textbf{Average}
    & 0.448 & 0.458 & 0.458 & 0.584
    & 0.569 & \textbf{0.401} & 0.599 & 0.529
    & 0.484 & 0.647 & 0.704 & 0.616 \\
    & \textbf{MSE} 
    & 0.390 & 0.387 & 0.387 & 0.408
    & 0.385 & \textbf{0.305} & 0.552 & 0.506
    & 0.401 & 0.428 & 0.610 & 0.631 \\
    & \textbf{MAE} 
    & 0.470 & 0.488 & 0.488 & 0.608
    & 0.632 & \textbf{0.435} & 0.589 & 0.528
    & 0.475 & 0.679 & 0.772 & 0.605 \\
    & \textbf{CRPS} 
    & 0.484 & 0.498 & 0.498 & 0.737
    & 0.689 & \textbf{0.462} & 0.657 & 0.553
    & 0.576 & 0.834 & 0.731 & 0.611 \\
    \bottomrule
  \end{tabular}
  }
\end{table}

\subsection{Full Results of Multivariate Forecasting}
Table~\ref{tab:tslib_multivar_full_result} summarizes results of multivariate forecasting across seven widely used datasets. On this benchmark, CoRA achieves state-of-the-art performance across all datasets, substantially improving upon recent deep forecasters. These results demonstrate that CoRA can jointly predict multiple target variables in a unified manner, highlighting its effectiveness as a general adaptation strategy.

\begin{table}[htbp]
  \caption{Full results of the multivariate forecasting task. For all baselines, the look-back length $L$ is fixed at 2880, and $Avg$ means the average results from all four prediction lengths.}
  \vspace{-3pt}
  \renewcommand{\arraystretch}{0.85} 
  \centering
  \resizebox{1\columnwidth}{!}{
  \begin{threeparttable}
  \begin{small}
  \renewcommand{\multirowsetup}{\centering}
  \setlength{\tabcolsep}{1.45pt}
  \label{tab:tslib_multivar_full_result}
  \begin{tabular}{c|c|cc|cc|cc|cc|cc|cc|cc|cc|cc|cc}
    \toprule
    \multicolumn{2}{c|}{\multirow{2}{*}{Models}} & 
    \multicolumn{2}{c}{\rotatebox{0}{\scalebox{0.75}{\textbf{CoRA}}}} &
    \multicolumn{2}{c}{\rotatebox{0}{\scalebox{0.8}{Timer-XL}}} &
    \multicolumn{2}{c}{\rotatebox{0}{\scalebox{0.8}{TimeXer}}} &
    \multicolumn{2}{c}{\rotatebox{0}{\scalebox{0.8}{iTransformer}}} &
    \multicolumn{2}{c}{\rotatebox{0}{\scalebox{0.8}{{PatchTST}}}} &
    \multicolumn{2}{c}{\rotatebox{0}{\scalebox{0.8}{Crossformer}}} &
    \multicolumn{2}{c}{\rotatebox{0}{\scalebox{0.8}{TiDE}}} &
    \multicolumn{2}{c}{\rotatebox{0}{\scalebox{0.8}{DLinear}}} &
    \multicolumn{2}{c}{\rotatebox{0}{\scalebox{0.8}{SCINet}}} &
    \multicolumn{2}{c}{\rotatebox{0}{\scalebox{0.8}{Autoformer}}}  \\
    \multicolumn{2}{c|}{} &
    \multicolumn{2}{c}{\scalebox{0.8}{\textbf{(Ours)}}} &
    \multicolumn{2}{c}{\scalebox{0.8}{\citeyearpar{liu2024timer}}} & 
    \multicolumn{2}{c}{\scalebox{0.8}{\citeyearpar{wang2024timexer}}} & 
    \multicolumn{2}{c}{\scalebox{0.8}{\citeyearpar{liu2023itransformer}}} & 
    \multicolumn{2}{c}{\scalebox{0.8}{\citeyearpar{nie2022time}}} & 
    \multicolumn{2}{c}{\scalebox{0.8}{\citeyearpar{zhang2023crossformer}}} & 
    \multicolumn{2}{c}{\scalebox{0.8}{\citeyearpar{das2023long}}} &
    \multicolumn{2}{c}{\scalebox{0.8}{\citeyearpar{zeng2023transformers}}} &
    \multicolumn{2}{c}{\scalebox{0.8}{\citeyearpar{liu2022scinet}}} &
    \multicolumn{2}{c}{\scalebox{0.8}{\citeyearpar{wu2021autoformer}}}  \\
    \cmidrule(lr){3-4}\cmidrule(lr){5-6} \cmidrule(lr){7-8}\cmidrule(lr){9-10}\cmidrule(lr){11-12} \cmidrule(lr){13-14} \cmidrule(lr){15-16} \cmidrule(lr){17-18} \cmidrule(lr){19-20} \cmidrule(lr){21-22} 
    \multicolumn{2}{c|}{Metric}  
    & \scalebox{0.8}{MSE} & \scalebox{0.8}{MAE}  
    & \scalebox{0.8}{MSE} & \scalebox{0.8}{MAE}  
    & \scalebox{0.8}{MSE} & \scalebox{0.8}{MAE}  
    & \scalebox{0.8}{MSE} & \scalebox{0.8}{MAE}  
    & \scalebox{0.8}{MSE} & \scalebox{0.8}{MAE}  
    & \scalebox{0.8}{MSE} & \scalebox{0.8}{MAE} 
    & \scalebox{0.8}{MSE} & \scalebox{0.8}{MAE} 
    & \scalebox{0.8}{MSE} & \scalebox{0.8}{MAE} 
    & \scalebox{0.8}{MSE} & \scalebox{0.8}{MAE} 
    & \scalebox{0.8}{MSE} & \scalebox{0.8}{MAE} \\
    \toprule 
    
    \multirow{5}{*}{\rotatebox{90}{ETTh1}}
    & \scalebox{0.8}{96}
    & \boldres{\scalebox{0.8}{0.344}} & \boldres{\scalebox{0.8}{0.381}} 
    & \scalebox{0.8}{0.483} & \scalebox{0.8}{0.485} 
    & \secondres{\scalebox{0.8}{0.411}} & \secondres{\scalebox{0.8}{0.438}} 
    & \scalebox{0.8}{0.436} & \scalebox{0.8}{0.466} 
    & \scalebox{0.8}{0.428} & \scalebox{0.8}{0.450} 
    & \scalebox{0.8}{0.479} & \scalebox{0.8}{0.494} 
    & \scalebox{0.8}{0.565} & \scalebox{0.8}{0.536}
    & \scalebox{0.8}{0.433} & \scalebox{0.8}{0.451} 
    & \scalebox{0.8}{0.713} & \scalebox{0.8}{0.625}
    & \scalebox{0.8}{0.630} & \scalebox{0.8}{0.524} \\ 
    
    & \scalebox{0.8}{192}
    & \boldres{\scalebox{0.8}{0.387}} & \boldres{\scalebox{0.8}{0.408}} 
    & \scalebox{0.8}{0.520} & \scalebox{0.8}{0.506} 
    & \secondres{\scalebox{0.8}{0.442}} & \secondres{\scalebox{0.8}{0.459}} 
    & \scalebox{0.8}{0.469} & \scalebox{0.8}{0.487} 
    & \scalebox{0.8}{0.476} & \scalebox{0.8}{0.477} 
    & \scalebox{0.8}{0.587} & \scalebox{0.8}{0.550} 
    & \scalebox{0.8}{0.634} & \scalebox{0.8}{0.572}
    & \scalebox{0.8}{0.479} & \scalebox{0.8}{0.482} 
    & \scalebox{0.8}{0.736} & \scalebox{0.8}{0.638}
    & \scalebox{0.8}{0.762} & \scalebox{0.8}{0.519} \\ 
    
    & \scalebox{0.8}{336}
    & \boldres{\scalebox{0.8}{0.412}} & \boldres{\scalebox{0.8}{0.425}} 
    & \scalebox{0.8}{0.540} & \scalebox{0.8}{0.564} 
    & \secondres{\scalebox{0.8}{0.477}} & \secondres{\scalebox{0.8}{0.484}} 
    & \scalebox{0.8}{0.510} & \scalebox{0.8}{0.515} 
    & \scalebox{0.8}{0.519} & \scalebox{0.8}{0.504} 
    & \scalebox{0.8}{0.641} & \scalebox{0.8}{0.600} 
    & \scalebox{0.8}{0.672} & \scalebox{0.8}{0.593}
    & \scalebox{0.8}{0.533} & \scalebox{0.8}{0.519} 
    & \scalebox{0.8}{0.773} & \scalebox{0.8}{0.658}
    & \scalebox{0.8}{0.886} & \scalebox{0.8}{0.766} \\ 
    
    & \scalebox{0.8}{720}
    & \boldres{\scalebox{0.8}{0.471}} & \boldres{\scalebox{0.8}{0.473}} 
    & \scalebox{0.8}{0.647} & \scalebox{0.8}{0.633} 
    & \scalebox{0.8}{0.639} & \secondres{\scalebox{0.8}{0.572} }
    & \secondres{\scalebox{0.8}{0.619}} & \scalebox{0.8}{0.591} 
    & \scalebox{0.8}{0.639} & \scalebox{0.8}{0.586} 
    & \scalebox{0.8}{0.867} & \scalebox{0.8}{0.732} 
    & \scalebox{0.8}{0.751} & \scalebox{0.8}{0.646}
    & \scalebox{0.8}{0.633} & \scalebox{0.8}{0.596} 
    & \scalebox{0.8}{0.897} & \scalebox{0.8}{0.717}
    & \scalebox{0.8}{0.971} & \scalebox{0.8}{0.836} \\ 
    \cmidrule(lr){2-22}
    
    & \scalebox{0.8}{Avg}
    & \boldres{\scalebox{0.8}{0.404}} & \boldres{\scalebox{0.8}{0.422}} 
    & \scalebox{0.8}{0.548} & \scalebox{0.8}{0.547} 
    & \secondres{\scalebox{0.8}{0.492}} & \secondres{\scalebox{0.8}{0.488}} 
    & \scalebox{0.8}{0.508} & \scalebox{0.8}{0.515} 
    & \scalebox{0.8}{0.516} & \scalebox{0.8}{0.504} 
    & \scalebox{0.8}{0.643} & \scalebox{0.8}{0.594} 
    & \scalebox{0.8}{0.656} & \scalebox{0.8}{0.587}
    & \scalebox{0.8}{0.519} & \scalebox{0.8}{0.512} 
    & \scalebox{0.8}{0.780} & \scalebox{0.8}{0.660}
    & \scalebox{0.8}{0.812} & \scalebox{0.8}{0.661} \\ 
    \midrule 

    \multirow{5}{*}{\rotatebox{90}{ETTh2}}
    & \scalebox{0.8}{96}
    & \boldres{\scalebox{0.8}{0.271}} & \boldres{\scalebox{0.8}{0.329}} 
    & \secondres{\scalebox{0.8}{0.314}} & \secondres{\scalebox{0.8}{0.378}} 
    & \scalebox{0.8}{0.350} & \scalebox{0.8}{0.409} 
    & \scalebox{0.8}{0.344} & \scalebox{0.8}{0.414} 
    & \scalebox{0.8}{0.369} & \scalebox{0.8}{0.426} 
    & \scalebox{0.8}{0.725} & \scalebox{0.8}{0.622} 
    & \scalebox{0.8}{0.442} & \scalebox{0.8}{0.468}
    & \scalebox{0.8}{0.458} & \scalebox{0.8}{0.474} 
    & \scalebox{0.8}{0.544} & \scalebox{0.8}{0.535}
    & \scalebox{0.8}{0.731} & \scalebox{0.8}{0.635} \\ 
    
    & \scalebox{0.8}{192}
    & \boldres{\scalebox{0.8}{0.328}} & \boldres{\scalebox{0.8}{0.373}} 
    & \secondres{\scalebox{0.8}{0.387}} & \secondres{\scalebox{0.8}{0.428}} 
    & \scalebox{0.8}{0.414} & \scalebox{0.8}{0.452} 
    & \scalebox{0.8}{0.408} & \scalebox{0.8}{0.454} 
    & \scalebox{0.8}{0.469} & \scalebox{0.8}{0.491} 
    & \scalebox{0.8}{0.771} & \scalebox{0.8}{0.684} 
    & \scalebox{0.8}{0.509} & \scalebox{0.8}{0.504}
    & \scalebox{0.8}{0.547} & \scalebox{0.8}{0.521} 
    & \scalebox{0.8}{0.614} & \scalebox{0.8}{0.570}
    & \scalebox{0.8}{0.789} & \scalebox{0.8}{0.677} \\ 
    
    & \scalebox{0.8}{336}
    & \boldres{\scalebox{0.8}{0.353}} & \boldres{\scalebox{0.8}{0.397}} 
    & \secondres{\scalebox{0.8}{0.445}} & \secondres{\scalebox{0.8}{0.473}} 
    & \scalebox{0.8}{0.455} & \scalebox{0.8}{0.482} 
    & \scalebox{0.8}{0.473} & \scalebox{0.8}{0.502} 
    & \scalebox{0.8}{0.563} & \scalebox{0.8}{0.548} 
    & \scalebox{0.8}{0.852} & \scalebox{0.8}{0.701} 
    & \scalebox{0.8}{0.582} & \scalebox{0.8}{0.549}
    & \scalebox{0.8}{0.667} & \scalebox{0.8}{0.619} 
    & \scalebox{0.8}{0.669} & \scalebox{0.8}{0.596}
    & \scalebox{0.8}{0.898} & \scalebox{0.8}{0.738} \\ 
    
    & \scalebox{0.8}{720}
    & \boldres{\scalebox{0.8}{0.372}} & \boldres{\scalebox{0.8}{0.424}} 
    & \scalebox{0.8}{0.541} & \scalebox{0.8}{0.538} 
    & \scalebox{0.8}{0.595} & \scalebox{0.8}{0.561} 
    & \secondres{\scalebox{0.8}{0.533}} & \secondres{\scalebox{0.8}{0.533}} 
    & \scalebox{0.8}{0.558} & \scalebox{0.8}{0.547} 
    & \scalebox{0.8}{0.893} & \scalebox{0.8}{0.755} 
    & \scalebox{0.8}{0.688} & \scalebox{0.8}{0.606}
    & \scalebox{0.8}{0.808} & \scalebox{0.8}{0.743} 
    & \scalebox{0.8}{0.841} & \scalebox{0.8}{0.667}
    & \scalebox{0.8}{0.941} & \scalebox{0.8}{0.778} \\ 
    \cmidrule(lr){2-22}
    
    & \scalebox{0.8}{Avg}
    & \boldres{\scalebox{0.8}{0.331}} & \boldres{\scalebox{0.8}{0.381}} 
    & \secondres{\scalebox{0.8}{0.422}} & \secondres{\scalebox{0.8}{0.454}} 
    & \scalebox{0.8}{0.454} & \scalebox{0.8}{0.476} 
    & \scalebox{0.8}{0.440} & \scalebox{0.8}{0.476} 
    & \scalebox{0.8}{0.490} & \scalebox{0.8}{0.503} 
    & \scalebox{0.8}{0.810} & \scalebox{0.8}{0.691} 
    & \scalebox{0.8}{0.555} & \scalebox{0.8}{0.532}
    & \scalebox{0.8}{0.620} & \scalebox{0.8}{0.589} 
    & \scalebox{0.8}{0.667} & \scalebox{0.8}{0.592}
    & \scalebox{0.8}{0.840} & \scalebox{0.8}{0.707} \\ 
    \midrule 
    
    \multirow{5}{*}{\rotatebox{90}{ETTm1}}
    & \scalebox{0.8}{96}
    & \boldres{\scalebox{0.8}{0.294}} & \boldres{\scalebox{0.8}{0.336}} 
    & \scalebox{0.8}{0.313} & \scalebox{0.8}{0.374} 
    & \scalebox{0.8}{0.356} & \scalebox{0.8}{0.397} 
    & \scalebox{0.8}{0.342} & \scalebox{0.8}{0.388} 
    & \scalebox{0.8}{0.333} & \scalebox{0.8}{0.382} 
    & \scalebox{0.8}{0.330} & \scalebox{0.8}{0.384} 
    & \scalebox{0.8}{0.317} & \scalebox{0.8}{0.364}
    & \secondres{\scalebox{0.8}{0.311}} & \secondres{\scalebox{0.8}{0.358}} 
    & \scalebox{0.8}{0.391} & \scalebox{0.8}{0.427}
    & \scalebox{0.8}{0.740} & \scalebox{0.8}{0.627} \\ 
    
    & \scalebox{0.8}{192}
    & \boldres{\scalebox{0.8}{0.325}} & \boldres{\scalebox{0.8}{0.361}} 
    & \scalebox{0.8}{0.358} & \scalebox{0.8}{0.403} 
    & \scalebox{0.8}{0.388} & \scalebox{0.8}{0.416} 
    & \scalebox{0.8}{0.363} & \scalebox{0.8}{0.402} 
    & \scalebox{0.8}{0.375} & \scalebox{0.8}{0.408} 
    & \scalebox{0.8}{0.363} & \scalebox{0.8}{0.403} 
    & \scalebox{0.8}{0.352} & \scalebox{0.8}{0.387}
    & \secondres{\scalebox{0.8}{0.341}} & \secondres{\scalebox{0.8}{0.377}} 
    & \scalebox{0.8}{0.410} & \scalebox{0.8}{0.438}
    & \scalebox{0.8}{0.858} & \scalebox{0.8}{0.696} \\ 
    
    & \scalebox{0.8}{336}
    & \boldres{\scalebox{0.8}{0.347}} & \boldres{\scalebox{0.8}{0.380}} 
    & \scalebox{0.8}{0.397} & \scalebox{0.8}{0.430} 
    & \scalebox{0.8}{0.403} & \scalebox{0.8}{0.429} 
    & \scalebox{0.8}{0.386} & \scalebox{0.8}{0.419} 
    & \scalebox{0.8}{0.442} & \scalebox{0.8}{0.453} 
    & \scalebox{0.8}{0.413} & \scalebox{0.8}{0.445} 
    & \scalebox{0.8}{0.371} & \scalebox{0.8}{0.397}
    & \secondres{\scalebox{0.8}{0.366}} & \secondres{\scalebox{0.8}{0.392}} 
    & \scalebox{0.8}{0.431} & \scalebox{0.8}{0.450}
    & \scalebox{0.8}{0.895} & \scalebox{0.8}{0.705} \\ 
    
    & \scalebox{0.8}{720}
    & \boldres{\scalebox{0.8}{0.381}} & \boldres{\scalebox{0.8}{0.407}} 
    & \scalebox{0.8}{0.456} & \scalebox{0.8}{0.469} 
    & \scalebox{0.8}{0.444} & \scalebox{0.8}{0.455} 
    & \scalebox{0.8}{0.423} & \scalebox{0.8}{0.444} 
    & \scalebox{0.8}{0.449} & \scalebox{0.8}{0.453} 
    & \scalebox{0.8}{0.639} & \scalebox{0.8}{0.596} 
    & \scalebox{0.8}{0.413} & \secondres{\scalebox{0.8}{0.422}}
    & \secondres{\scalebox{0.8}{0.410}} & \secondres{\scalebox{0.8}{0.422}} 
    & \scalebox{0.8}{0.468} & \scalebox{0.8}{0.472}
    & \scalebox{0.8}{0.934} & \scalebox{0.8}{0.700} \\ 
    \cmidrule(lr){2-22}
    
    & \scalebox{0.8}{Avg}
    & \boldres{\scalebox{0.8}{0.337}} & \boldres{\scalebox{0.8}{0.371}} 
    & \scalebox{0.8}{0.381} & \scalebox{0.8}{0.419} 
    & \scalebox{0.8}{0.398} & \scalebox{0.8}{0.424} 
    & \scalebox{0.8}{0.379} & \scalebox{0.8}{0.413} 
    & \scalebox{0.8}{0.400} & \scalebox{0.8}{0.424} 
    & \scalebox{0.8}{0.436} & \scalebox{0.8}{0.457} 
    & \scalebox{0.8}{0.363} & \scalebox{0.8}{0.393}
    & \secondres{\scalebox{0.8}{0.357}} & \secondres{\scalebox{0.8}{0.387} }
    & \scalebox{0.8}{0.425} & \scalebox{0.8}{0.447}
    & \scalebox{0.8}{0.857} & \scalebox{0.8}{0.682} \\ 
    \midrule 
    
    \multirow{5}{*}{\rotatebox{90}{ETTm2}}
    & \scalebox{0.8}{96}
    & \secondres{\scalebox{0.8}{0.167}} & \boldres{\scalebox{0.8}{0.252}} 
    & \scalebox{0.8}{0.225} & \scalebox{0.8}{0.319} 
    & \scalebox{0.8}{0.189} & \scalebox{0.8}{0.287} 
    & \scalebox{0.8}{0.189} & \scalebox{0.8}{0.285}
    & \scalebox{0.8}{0.182} & \scalebox{0.8}{0.280} 
    & \scalebox{0.8}{0.391} & \scalebox{0.8}{0.469} 
    & \scalebox{0.8}{0.198} & \scalebox{0.8}{0.294}
    & \boldres{\scalebox{0.8}{0.165}} & \secondres{\scalebox{0.8}{0.262}} 
    & \scalebox{0.8}{0.242} & \scalebox{0.8}{0.337}
    & \scalebox{0.8}{0.381} & \scalebox{0.8}{0.453} \\ 
    
    & \scalebox{0.8}{192}
    & \secondres{\scalebox{0.8}{0.224}} & \boldres{\scalebox{0.8}{0.295}} 
    & \scalebox{0.8}{0.291} & \scalebox{0.8}{0.366} 
    & \scalebox{0.8}{0.249} & \scalebox{0.8}{0.330} 
    & \scalebox{0.8}{0.238} & \scalebox{0.8}{0.318} 
    & \scalebox{0.8}{0.238} & \scalebox{0.8}{0.317} 
    & \scalebox{0.8}{0.475} & \scalebox{0.8}{0.514} 
    & \scalebox{0.8}{0.323} & \scalebox{0.8}{0.387}
    & \boldres{\scalebox{0.8}{0.220}} & \secondres{\scalebox{0.8}{0.304}} 
    & \scalebox{0.8}{0.282} & \scalebox{0.8}{0.361}
    & \scalebox{0.8}{0.449} & \scalebox{0.8}{0.493} \\ 
    
    & \scalebox{0.8}{336}
    & \secondres{\scalebox{0.8}{0.278}} & \boldres{\scalebox{0.8}{0.334}} 
    & \scalebox{0.8}{0.344} & \scalebox{0.8}{0.402} 
    & \scalebox{0.8}{0.291} & \scalebox{0.8}{0.352} 
    & \scalebox{0.8}{0.298} & \scalebox{0.8}{0.356}
    & \scalebox{0.8}{0.311} & \scalebox{0.8}{0.368} 
    & \scalebox{0.8}{0.663} & \scalebox{0.8}{0.674} 
    & \scalebox{0.8}{0.332} & \scalebox{0.8}{0.390}
    & \boldres{\scalebox{0.8}{0.268}} & \secondres{\scalebox{0.8}{0.338}} 
    & \scalebox{0.8}{0.322} & \scalebox{0.8}{0.387}
    & \scalebox{0.8}{0.503} & \scalebox{0.8}{0.521} \\ 
    
    & \scalebox{0.8}{720}
    & \boldres{\scalebox{0.8}{0.354}} & \boldres{\scalebox{0.8}{0.388}} 
    & \scalebox{0.8}{0.412} & \scalebox{0.8}{0.445} 
    & \secondres{\scalebox{0.8}{0.368}} & \secondres{\scalebox{0.8}{0.402}} 
    & \scalebox{0.8}{0.377} & \scalebox{0.8}{0.407} 
    & \scalebox{0.8}{0.437} & \scalebox{0.8}{0.454} 
    & \scalebox{0.8}{0.745} & \scalebox{0.8}{0.716} 
    & \scalebox{0.8}{0.372} & \scalebox{0.8}{0.409}
    & \scalebox{0.8}{0.410} & \scalebox{0.8}{0.435} 
    & \scalebox{0.8}{0.385} & \scalebox{0.8}{0.426}
    & \scalebox{0.8}{0.494} & \scalebox{0.8}{0.514} \\ 
    \cmidrule(lr){2-22}
    
    & \scalebox{0.8}{Avg}
    & \boldres{\scalebox{0.8}{0.256}} & \boldres{\scalebox{0.8}{0.317}}
    & \scalebox{0.8}{0.318} & \scalebox{0.8}{0.383} 
    & \scalebox{0.8}{0.274} & \scalebox{0.8}{0.343} 
    & \scalebox{0.8}{0.276} & \scalebox{0.8}{0.342} 
    & \scalebox{0.8}{0.292} & \scalebox{0.8}{0.355} 
    & \scalebox{0.8}{0.569} & \scalebox{0.8}{0.593} 
    & \scalebox{0.8}{0.306} & \scalebox{0.8}{0.370}
    & \secondres{\scalebox{0.8}{0.266}} & \secondres{\scalebox{0.8}{0.335}} 
    & \scalebox{0.8}{0.308} & \scalebox{0.8}{0.378}
    & \scalebox{0.8}{0.457} & \scalebox{0.8}{0.495} \\ 
    \midrule 
    
    \multirow{5}{*}{\rotatebox{90}{Weather}}
    & \scalebox{0.8}{96}
    & \boldres{\scalebox{0.8}{0.158}} & \boldres{\scalebox{0.8}{0.206}} 
    & \scalebox{0.8}{0.255} & \scalebox{0.8}{0.299}     
    & \scalebox{0.8}{0.186} & \scalebox{0.8}{0.246} 
    & \scalebox{0.8}{0.187} & \scalebox{0.8}{0.252} 
    & \scalebox{0.8}{0.160} & \scalebox{0.8}{0.219} 
    & \secondres{\scalebox{0.8}{0.159}} & \secondres{\scalebox{0.8}{0.212}} 
    & \scalebox{0.8}{0.171} & \scalebox{0.8}{0.231}
    & \scalebox{0.8}{0.169} & \scalebox{0.8}{0.230} 
    & \scalebox{0.8}{0.168} & \scalebox{0.8}{0.231}
    & \scalebox{0.8}{0.400} & \scalebox{0.8}{0.433} \\ 
    
    & \scalebox{0.8}{192}
    & \secondres{\scalebox{0.8}{0.201}} & \boldres{\scalebox{0.8}{0.248}} 
    & \scalebox{0.8}{0.315} & \scalebox{0.8}{0.344} 
    & \scalebox{0.8}{0.233} & \scalebox{0.8}{0.286} 
    & \scalebox{0.8}{0.231} & \scalebox{0.8}{0.291} 
    & \scalebox{0.8}{0.210} & \scalebox{0.8}{0.265}
    & \boldres{\scalebox{0.8}{0.198}} & \secondres{\scalebox{0.8}{0.263}} 
    & \scalebox{0.8}{0.211} & \scalebox{0.8}{0.265}
    & \scalebox{0.8}{0.210} & \scalebox{0.8}{0.267} 
    & \scalebox{0.8}{0.216} & \scalebox{0.8}{0.274}
    & \scalebox{0.8}{0.447} & \scalebox{0.8}{0.448} \\ 
    
    & \scalebox{0.8}{336}
    & \secondres{\scalebox{0.8}{0.249}} & \boldres{\scalebox{0.8}{0.288}} 
    & \scalebox{0.8}{0.331} & \scalebox{0.8}{0.366}     
    & \scalebox{0.8}{0.281} & \scalebox{0.8}{0.318} 
    & \scalebox{0.8}{0.273} & \scalebox{0.8}{0.325} 
    & \scalebox{0.8}{0.273} & \scalebox{0.8}{0.309} 
    & \boldres{\scalebox{0.8}{0.246}} & \scalebox{0.8}{0.298}
    & \scalebox{0.8}{0.253} & \secondres{\scalebox{0.8}{0.296}}
    & \scalebox{0.8}{0.257} & \scalebox{0.8}{0.310} 
    & \scalebox{0.8}{0.299} & \scalebox{0.8}{0.333}
    & \scalebox{0.8}{0.462} & \scalebox{0.8}{0.452} \\ 
    
    & \scalebox{0.8}{720}
    & \secondres{\scalebox{0.8}{0.311}} & \secondres{\scalebox{0.8}{0.333}} 
    & \scalebox{0.8}{0.361} & \scalebox{0.8}{0.384} 
    & \scalebox{0.8}{0.347} & \scalebox{0.8}{0.361} 
    & \scalebox{0.8}{0.314} & \scalebox{0.8}{0.352} 
    & \scalebox{0.8}{0.359} & \scalebox{0.8}{0.368} 
    & \scalebox{0.8}{0.335} & \scalebox{0.8}{0.369} 
    & \boldres{\scalebox{0.8}{0.300}} & \boldres{\scalebox{0.8}{0.332}}
    & \scalebox{0.8}{0.314} & \scalebox{0.8}{0.357} 
    & \scalebox{0.8}{0.314} & \scalebox{0.8}{0.344}
    & \scalebox{0.8}{0.693} & \scalebox{0.8}{0.616} \\ 
    \cmidrule(lr){2-22}
    
    & \scalebox{0.8}{Avg}
    & \boldres{\scalebox{0.8}{0.230}} & \boldres{\scalebox{0.8}{0.269}} 
    & \scalebox{0.8}{0.316} & \scalebox{0.8}{0.348} 
    & \scalebox{0.8}{0.262} & \scalebox{0.8}{0.303} 
    & \scalebox{0.8}{0.251} & \scalebox{0.8}{0.305} 
    & \scalebox{0.8}{0.251} & \scalebox{0.8}{0.290} 
    & \scalebox{0.8}{0.235} & \scalebox{0.8}{0.285} 
    & \secondres{\scalebox{0.8}{0.234}} & \secondres{\scalebox{0.8}{0.281}}
    & \scalebox{0.8}{0.237} & \scalebox{0.8}{0.291} 
    & \scalebox{0.8}{0.249} & \scalebox{0.8}{0.296}
    & \scalebox{0.8}{0.500} & \scalebox{0.8}{0.487} \\ 
    \midrule 

    \multirow{5}{*}{\rotatebox{90}{ECL}}
    & \scalebox{0.8}{96} 
    & \boldres{\scalebox{0.8}{0.124}} & \boldres{\scalebox{0.8}{0.220}} 
    & \scalebox{0.8}{0.131} & \scalebox{0.8}{0.229} 
    & \scalebox{0.8}{0.137} & \scalebox{0.8}{0.241} 
    & \scalebox{0.8}{0.167} & \scalebox{0.8}{0.275} 
    & \scalebox{0.8}{0.136} & \scalebox{0.8}{0.240} 
    & \scalebox{0.8}{0.133} & \scalebox{0.8}{0.232} 
    & \scalebox{0.8}{0.130} & \secondres{\scalebox{0.8}{0.226}}
    & \secondres{\scalebox{0.8}{0.129}} & \scalebox{0.8}{0.227} 
    & \scalebox{0.8}{0.144} & \scalebox{0.8}{0.252}
    & \scalebox{0.8}{0.256} & \scalebox{0.8}{0.362} \\ 
    
    & \scalebox{0.8}{192}
    & \boldres{\scalebox{0.8}{0.142}} & \boldres{\scalebox{0.8}{0.238}} 
    & \scalebox{0.8}{0.147} & \scalebox{0.8}{0.244} 
    & \scalebox{0.8}{0.154} & \scalebox{0.8}{0.256} 
    & \scalebox{0.8}{0.177} & \scalebox{0.8}{0.283} 
    & \scalebox{0.8}{0.151} & \scalebox{0.8}{0.254} 
    & \scalebox{0.8}{0.162} & \scalebox{0.8}{0.266} 
    & \scalebox{0.8}{0.146} & \scalebox{0.8}{0.242}
    & \secondres{\scalebox{0.8}{0.144}} & \secondres{\scalebox{0.8}{0.242}} 
    & \scalebox{0.8}{0.163} & \scalebox{0.8}{0.271}
    & \scalebox{0.8}{0.267} & \scalebox{0.8}{0.371} \\ 
    
    & \scalebox{0.8}{336}
    & \boldres{\scalebox{0.8}{0.159}} & \boldres{\scalebox{0.8}{0.256}} 
    & \boldres{\scalebox{0.8}{0.159}} & \secondres{\scalebox{0.8}{0.257}} 
    & \scalebox{0.8}{0.189} & \scalebox{0.8}{0.291} 
    & \scalebox{0.8}{0.196} & \scalebox{0.8}{0.302} 
    & \secondres{\scalebox{0.8}{0.167}} & \scalebox{0.8}{0.269} 
    & \scalebox{0.8}{0.191} & \scalebox{0.8}{0.286} 
    & \scalebox{0.8}{0.163} & \scalebox{0.8}{0.259}
    & \boldres{\scalebox{0.8}{0.159}} & \scalebox{0.8}{0.260} 
    & \scalebox{0.8}{0.178} & \scalebox{0.8}{0.286}
    & \scalebox{0.8}{0.278} & \scalebox{0.8}{0.376} \\ 
    
    & \scalebox{0.8}{720}
    & \scalebox{0.8}{0.194} & \secondres{\scalebox{0.8}{0.287}} 
    & \boldres{\scalebox{0.8}{0.183}} & \boldres{\scalebox{0.8}{0.279}} 
    & \scalebox{0.8}{0.210} & \scalebox{0.8}{0.311} 
    & \scalebox{0.8}{0.234} & \scalebox{0.8}{0.335} 
    & \scalebox{0.8}{0.199} & \scalebox{0.8}{0.297} 
    & \scalebox{0.8}{0.249} & \scalebox{0.8}{0.338} 
    & \scalebox{0.8}{0.199} & \scalebox{0.8}{0.290}
    & \secondres{\scalebox{0.8}{0.192}} & \scalebox{0.8}{0.292} 
    & \scalebox{0.8}{0.239} & \scalebox{0.8}{0.331}
    & \scalebox{0.8}{0.367} & \scalebox{0.8}{0.451} \\ 
    \cmidrule(lr){2-22}
    
    & \scalebox{0.8}{Avg}
    & \boldres{\scalebox{0.8}{0.155}} & \boldres{\scalebox{0.8}{0.250}} 
    & \boldres{\scalebox{0.8}{0.155}} & \secondres{\scalebox{0.8}{0.252}} 
    & \scalebox{0.8}{0.172} & \scalebox{0.8}{0.275} 
    & \scalebox{0.8}{0.194} & \scalebox{0.8}{0.299} 
    & \scalebox{0.8}{0.163} & \scalebox{0.8}{0.265} 
    & \scalebox{0.8}{0.184} & \scalebox{0.8}{0.281} 
    & \scalebox{0.8}{0.160} & \scalebox{0.8}{0.254}
    & \secondres{\scalebox{0.8}{0.156}} & \scalebox{0.8}{0.255} 
    & \scalebox{0.8}{0.181} & \scalebox{0.8}{0.285}
    & \scalebox{0.8}{0.292} & \scalebox{0.8}{0.390} \\ 
    \midrule
    
    \multirow{5}{*}{\rotatebox{90}{Traffic}}
    & \scalebox{0.8}{96}
    & \boldres{\scalebox{0.8}{0.350}} & \boldres{\scalebox{0.8}{0.245}} 
    & \scalebox{0.8}{0.569} & \scalebox{0.8}{0.428} 
    & \scalebox{0.8}{0.377} & \scalebox{0.8}{0.269} 
    & \secondres{\scalebox{0.8}{0.375}} & \scalebox{0.8}{0.275} 
    & \scalebox{0.8}{0.397} & \scalebox{0.8}{0.286} 
    & \scalebox{0.8}{0.481} & \secondres{\scalebox{0.8}{0.256}} 
    & \scalebox{0.8}{0.377} & \scalebox{0.8}{0.264}
    & \scalebox{0.8}{0.379} & \scalebox{0.8}{0.270} 
    & \scalebox{0.8}{0.455} & \scalebox{0.8}{0.342}
    & \scalebox{0.8}{0.538} & \scalebox{0.8}{0.405} \\ 
    
    & \scalebox{0.8}{192}
    & \boldres{\scalebox{0.8}{0.372}} & \boldres{\scalebox{0.8}{0.257}} 
    & \scalebox{0.8}{0.570} & \scalebox{0.8}{0.513} 
    & \secondres{\scalebox{0.8}{0.387}} & \scalebox{0.8}{0.274} 
    & \scalebox{0.8}{0.395} & \scalebox{0.8}{0.284} 
    & \scalebox{0.8}{0.410} & \scalebox{0.8}{0.293} 
    & \scalebox{0.8}{0.492} & \scalebox{0.8}{0.270} 
    & \scalebox{0.8}{0.390} & \secondres{\scalebox{0.8}{0.269}}
    & \scalebox{0.8}{0.392} & \scalebox{0.8}{0.276} 
    & \scalebox{0.8}{0.462} & \scalebox{0.8}{0.346}
    & \scalebox{0.8}{0.776} & \scalebox{0.8}{0.468} \\ 
    
    & \scalebox{0.8}{336}
    & \boldres{\scalebox{0.8}{0.389}} & \boldres{\scalebox{0.8}{0.267}} 
    & \scalebox{0.8}{0.589} & \scalebox{0.8}{0.521} 
    & \secondres{\scalebox{0.8}{0.400}} & \scalebox{0.8}{0.281} 
    & \scalebox{0.8}{0.410} & \scalebox{0.8}{0.292} 
    & \scalebox{0.8}{0.423} & \scalebox{0.8}{0.299} 
    & \scalebox{0.8}{0.514} & \scalebox{0.8}{0.277} 
    & \scalebox{0.8}{0.403} & \secondres{\scalebox{0.8}{0.275}}
    & \scalebox{0.8}{0.407} & \scalebox{0.8}{0.284}
    & \scalebox{0.8}{0.481} & \scalebox{0.8}{0.356}
    & \scalebox{0.8}{0.769} & \scalebox{0.8}{0.460} \\ 
    
    & \scalebox{0.8}{720}
    & \boldres{\scalebox{0.8}{0.426}} & \boldres{\scalebox{0.8}{0.288}} 
    & \scalebox{0.8}{0.658} & \scalebox{0.8}{0.577} 
    & \scalebox{0.8}{0.440} & \scalebox{0.8}{0.298} 
    & \scalebox{0.8}{0.447} & \scalebox{0.8}{0.312} 
    & \scalebox{0.8}{0.457} & \scalebox{0.8}{0.313} 
    & \scalebox{0.8}{0.601} & \scalebox{0.8}{0.337} 
    & \secondres{\scalebox{0.8}{0.438}} & \secondres{\scalebox{0.8}{0.294}}
    & \scalebox{0.8}{0.447} & \scalebox{0.8}{0.307} 
    & \scalebox{0.8}{0.512} & \scalebox{0.8}{0.363}
    & \scalebox{0.8}{0.885} & \scalebox{0.8}{0.524} \\ 
    \cmidrule(lr){2-22}
    
    & \scalebox{0.8}{Avg}
    & \boldres{\scalebox{0.8}{0.384}} & \boldres{\scalebox{0.8}{0.265}} 
    & \scalebox{0.8}{0.597} & \scalebox{0.8}{0.510} 
    & \secondres{\scalebox{0.8}{0.401}} & \scalebox{0.8}{0.281}
    & \scalebox{0.8}{0.407} & \scalebox{0.8}{0.291} 
    & \scalebox{0.8}{0.422} & \scalebox{0.8}{0.298} 
    & \scalebox{0.8}{0.522} & \scalebox{0.8}{0.285} 
    & \scalebox{0.8}{0.402} & \secondres{\scalebox{0.8}{0.276}}
    & \scalebox{0.8}{0.406} & \scalebox{0.8}{0.284} 
    & \scalebox{0.8}{0.478} & \scalebox{0.8}{0.352}
    & \scalebox{0.8}{0.742} & \scalebox{0.8}{0.464} \\ 
    \bottomrule
  \end{tabular}
    \end{small}
  \end{threeparttable}}
\end{table}

\subsection{Full Results of Generality}
We conduct extensive experiments on the EPF dataset using several representative TSFMs. As shown in Table~\ref{tab:generality}, CoRA consistently improves the performance of all TSFMs across both MSE and MAE metrics. Compared with their zero-shot baselines, the improvements are significant, demonstrating the generality and effectiveness of CoRA as a universal covariate adaptation method. We report results under the same training configuration and additionally provide the relative improvement ratio in MSE as a more intuitive assessment of the benefits brought by CoRA.
\begin{table}[htbp]
  \caption{Full results of CoRA generalize to other Time Series Foundation Models. We report the MSE/MAE and the relative MSE reduction ratios (Promotion) achieved by CoRA.}
  \label{tab:generality}
  \centering
  \begin{threeparttable}
  \begin{small}
  \renewcommand{\multirowsetup}{\centering}
  \setlength{\tabcolsep}{4.2pt}
  \resizebox{1\columnwidth}{!}{
  \begin{tabular}{c|cc|cc|cc|cc|cc|cc}
    \toprule
    {Datasets}
    & \multicolumn{2}{c}{NP} & \multicolumn{2}{c}{PJM} & \multicolumn{2}{c}{BE}  
    & \multicolumn{2}{c}{FR} & \multicolumn{2}{c}{DE} & \multicolumn{2}{c}{Avg} \\
    \cmidrule(lr){2-3} \cmidrule(lr){4-5} \cmidrule(lr){6-7} 
    \cmidrule(lr){8-9}\cmidrule(lr){10-11} \cmidrule(lr){12-13} 
    {Models}
    & MSE & MAE & MSE & MAE & MSE & MAE 
    & MSE & MAE & MSE & MAE & MSE & MAE \\
    \toprule
    Sundial 
    & 0.263 & 0.288 & 0.089 & 0.186 
    & 0.364 & 0.271 & 0.361 & 0.217 
    & 0.543 & 0.462 & 0.324 & 0.285 \\
    \textbf{ + CoRA } 
    & \textbf{0.222} & \textbf{0.246} & \textbf{0.073} 
    & \textbf{0.165} & \textbf{0.339} & \textbf{0.236} 
    & \textbf{0.357} & \textbf{0.206} & \textbf{0.401} 
    & \textbf{0.388} & \textbf{0.278} & \textbf{0.248} \\
    \cmidrule(lr){1-13}
    Promotion 
    & \multicolumn{2}{c|}{15.59\%} & \multicolumn{2}{c|}{17.98\%}
    & \multicolumn{2}{c|}{6.87\%}  & \multicolumn{2}{c|}{1.11\%}
    & \multicolumn{2}{c|}{26.15\%} & \multicolumn{2}{c}{14.20\%} \\
    \midrule
    TimesFM  
    & 0.255 & \textbf{0.271} & 0.085 & \textbf{0.182}
    & 0.383 & 0.252 & 0.398 & 0.206
    & 0.526 & 0.456 & 0.329 & 0.273 \\
    \textbf{ + CoRA }
    & \textbf{0.246} & \textbf{0.271} & \textbf{0.083}
    & \textbf{0.182} & \textbf{0.380} & \textbf{0.251}
    & \textbf{0.394} & \textbf{0.205} & \textbf{0.487}
    & \textbf{0.433} & \textbf{0.318} & \textbf{0.268} \\
    \cmidrule(lr){1-13}
    Promotion 
    & \multicolumn{2}{c|}{3.53\%} & \multicolumn{2}{c|}{2.35\%}
    & \multicolumn{2}{c|}{0.78\%} & \multicolumn{2}{c|}{1.01\%}
    & \multicolumn{2}{c|}{7.41\%} & \multicolumn{2}{c}{3.34\%} \\
    \midrule
    Chronos-Bolt  
    & 0.246 & 0.265 & 0.082 & 0.178 
    & 0.356 & 0.239 & 0.357 & 0.191 
    & 0.494 & 0.442 & 0.307 & 0.263 \\
    \textbf{ + CoRA }
    & \textbf{0.235} & \textbf{0.255} & \textbf{0.076} 
    & \textbf{0.170} & \textbf{0.353} & \textbf{0.233}
    & \textbf{0.352} & \textbf{0.184} & \textbf{0.445} 
    & \textbf{0.414} & \textbf{0.292} & \textbf{0.251} \\
    \cmidrule(lr){1-13}
    Promotion 
    & \multicolumn{2}{c|}{4.47\%} & \multicolumn{2}{c|}{7.32\%}
    & \multicolumn{2}{c|}{0.84\%} & \multicolumn{2}{c|}{1.40\%}
    & \multicolumn{2}{c|}{9.92\%} & \multicolumn{2}{c}{4.89\%} \\
    \midrule
    FlowState  
    & 0.229 & 0.256 & 0.081 & 0.\textbf{177}
    & 0.362 & 0.252 & 0.365 & 0.203
    & 0.497 & 0.446 & 0.307 & 0.267 \\
    \textbf{ + CoRA }
    & \textbf{0.225} & \textbf{0.253} & \textbf{0.078} 
    & \textbf{0.177} & \textbf{0.355} & \textbf{0.243} 
    & \textbf{0.364} & \textbf{0.199} & \textbf{0.464} 
    & \textbf{0.424} & \textbf{0.297} & \textbf{0.259} \\
    \cmidrule(lr){1-13}
    Promotion 
    & \multicolumn{2}{c|}{1.75\%} & \multicolumn{2}{c|}{3.70\%}
    & \multicolumn{2}{c|}{1.93\%} & \multicolumn{2}{c|}{0.27\%}
    & \multicolumn{2}{c|}{6.64\%} & \multicolumn{2}{c}{3.26\%}\\
    \bottomrule
  \end{tabular}}
  \end{small}
  \end{threeparttable}
\end{table}

\clearpage

\section{Showcases}
\begin{figure*}[htbp]
\begin{center}
	\centerline{\includegraphics[width=\columnwidth]{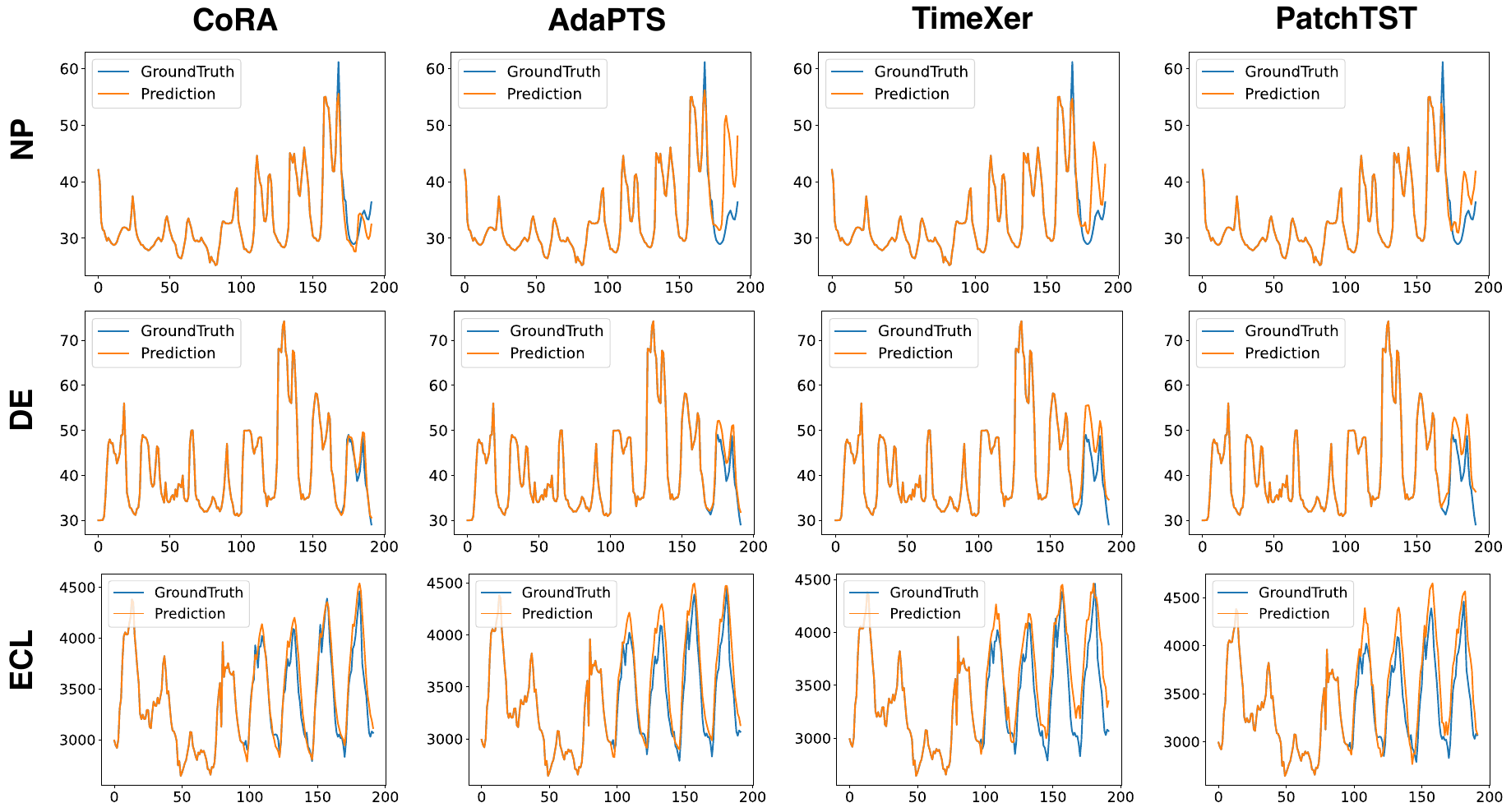}}
        \vspace{-7pt}
	\caption{Visualization of uni-modal covariate-aware results on NP, DE and ECL dataset.}
	\label{fig:unimodal_showcase}
\end{center}
\vspace{-20pt}
\end{figure*}

\begin{figure*}[htbp]
\begin{center}
	\centerline{\includegraphics[width=0.8\columnwidth]{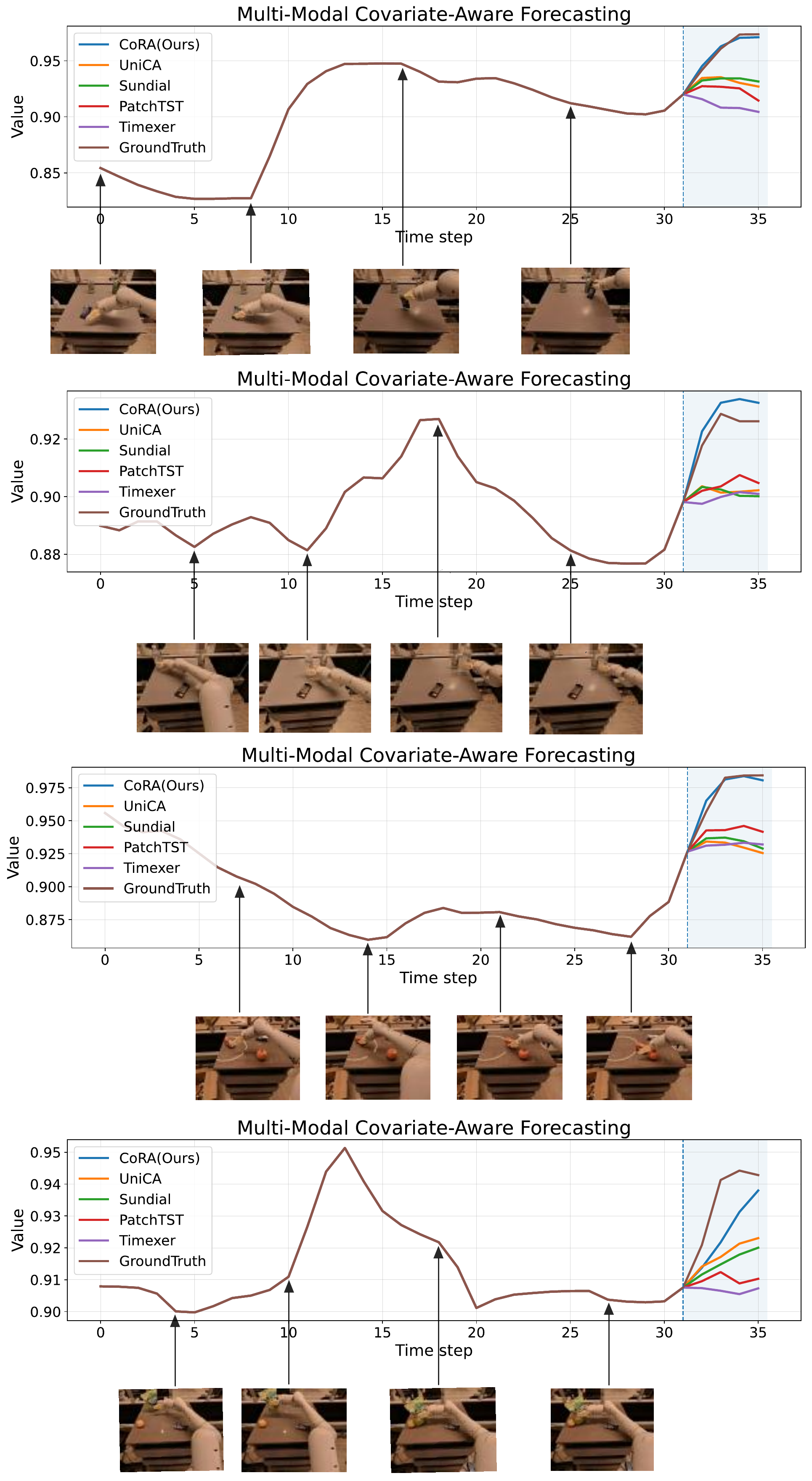}}
        \vspace{-7pt}
	\caption{Visualization of multi-modal covariate-aware results on RT-1 dataset.}
	\label{fig:multimodal_showcase}
\end{center}
\vspace{-20pt}
\end{figure*}

To facilitate a clear comparison among various models, we present additional prediction showcases for uni-modal covariate-aware forecasting in Figure~\ref{fig:unimodal_showcase}. These examples are provided by the following methods: AdaPTS~\citep{benechehab2025adapts}, TimeXer~\citep{wang2024timexer}, and PatchTST~\citep{nie2022time}. Of all the models, CoRA delivers the most accurate future series predictions.
Additionally, we provide the showcases of multi-modal covariate-aware forecasting in Figure~\ref{fig:multimodal_showcase}.

\section{Limitations}
A notable limitation of CoRA lies in its treatment of temporally aligned auxiliary modalities such as language and image sequences. At present, CoRA applies a simple mean aggregation along the temporal dimension, which inevitably discards fine-grained temporal dynamics and leads to underutilization of the rich and potentially complementary information contained in these modalities. Future work could investigate more sophisticated fusion strategies that explicitly preserve temporal dependencies, thereby enabling CoRA to more effectively leverage auxiliary modalities and further improve its adaptability across diverse forecasting scenarios.

\end{document}